\documentclass[10pt,journal,compsoc]{IEEEtran}

\usepackage[nocompress]{cite}
\usepackage[pdftex]{graphicx}
\usepackage{amsmath}
\usepackage{algorithm}
\usepackage{algpseudocode}

\usepackage[caption=false,font=footnotesize,labelfont=sf,textfont=sf]{subfig}
\usepackage{dblfloatfix} %\usepackage{fixltx2e}+\usepackage{stfloats}
\usepackage{url}
\usepackage{multirow}
\usepackage{booktabs}
\usepackage{bm}
\usepackage{amssymb}
\usepackage{ntheorem}
\theoremheaderfont{\normalfont\bfseries}
\theorembodyfont{\itshape}
\theoremseparator{}
\newtheorem{proposition}{Proposition}
\newtheorem{observation}{Observation}
\usepackage[pagebackref=false,breaklinks=true,bookmarks=false,hidelinks]{hyperref}

\usepackage{threeparttable}
\usepackage{xcolor}

\begin{document}
\title{\huge{AP-Loss for Accurate One-Stage Object Detection}}

\author{Kean Chen,~Weiyao Lin,~Jianguo Li,~John See,~Ji Wang,~and Junni Zou%
\IEEEcompsocitemizethanks{
%\IEEEcompsocthanksitem The basic idea of this paper appeared in our initial conference version~\cite{chen2019towards}. In this version, we extend our approach by introducing an acceleration strategy, further robustness analysis and more performance results.
\IEEEcompsocthanksitem The paper is supported in part by the following grants: China Major Project for New Generation of AI Grant (No. 2018AAA0100400), National Natural Science Foundation of China (No. 61971277), CREST Malaysia Grant T03C1-17. We gratefully acknowledge the support from Tencent YouTu Lab and Suzhou Institute of Artificial Intelligence, SJTU.
\IEEEcompsocthanksitem K. Chen, W. Lin and J. Zou are with the School of Electronic Information and Electrical Engineering, Shanghai Jiao Tong University, Shanghai, China. E-mail: \{ckadashuaige, wylin, zou-jn\}@sjtu.edu.cn.
\IEEEcompsocthanksitem J. Li is with the Intel Labs, Beijing, China. E-mail: jianguo.li@intel.com.
\IEEEcompsocthanksitem J. See is with the Faculty of Computing and Informatics, Multimedia University, Malaysia. E-mail: johnsee@mmu.edu.my.
\IEEEcompsocthanksitem J. Wang is with the Tencent YouTu Lab, Shanghai, China. E-mail: jensenjwang@tencent.com.
}
\thanks{(Corresponding author: Weiyao Lin.)}
}

\IEEEtitleabstractindextext{%
\begin{abstract}
One-stage object detectors are trained by optimizing classification-loss and localization-loss simultaneously, with the former suffering much from extreme foreground-background class imbalance issue due to the large number of anchors. This paper alleviates this issue by proposing a novel framework to replace the classification task in one-stage detectors with a ranking task, and adopting the Average-Precision loss (AP-loss) for the ranking problem. Due to its non-differentiability and non-convexity, the AP-loss cannot be optimized directly. For this purpose, we develop a novel optimization algorithm, which seamlessly combines the error-driven update scheme in perceptron learning and backpropagation algorithm in deep networks. We provide in-depth analyses on the good convergence property and computational complexity of the proposed algorithm, both theoretically and empirically. Experimental results demonstrate notable improvement in addressing the imbalance issue in object detection over existing AP-based optimization algorithms. An improved state-of-the-art performance is achieved in one-stage detectors based on AP-loss over detectors using classification-losses on various standard benchmarks. The proposed framework is also highly versatile in accommodating different network architectures. Code is available at \url{https://github.com/cccorn/AP-loss}.
\end{abstract}
}
\maketitle
\IEEEdisplaynontitleabstractindextext
\IEEEpeerreviewmaketitle

\IEEEraisesectionheading{\section{Introduction}}
\IEEEPARstart{O}bject detection is one of the fundamental problems in computer vision, which involves simultaneous localization and recognition of objects from images. It has made great progress owing to the rapid advances of deep learning methods in the last 5 years. A lot of deep learning based solutions have been proposed~\cite{ren2015faster,liu2016ssd,redmon2016you,he2017mask,cai2018cascade}, which typically adopt a multi-task architecture to handle the classification and localization tasks with separate loss functions~\cite{lin2018focal,girshick2015fast,liu2016ssd,redmon2016you}. The classification task aims to recognize the object in a given box, while the localization task aims to predict the precise bounding box of the object.

\begin{figure}[t]
\small
\centering
\subfloat[Acc \(=0.88\)\label{fig-classification}]{
  \includegraphics[width=0.98\linewidth]{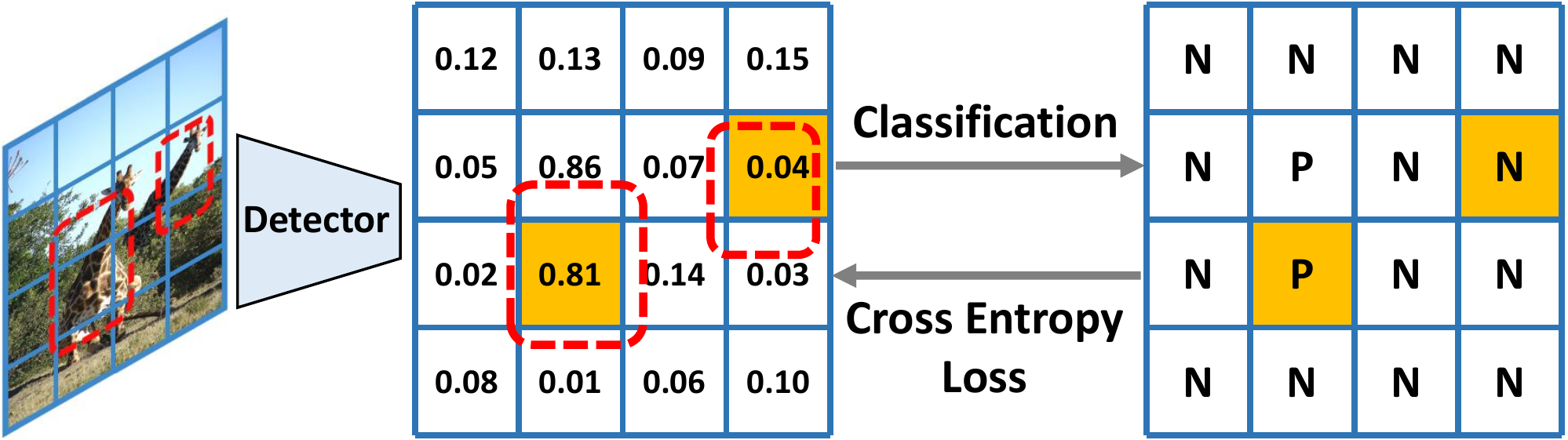}
  }
\par
\vspace{-2mm}
\subfloat[AP \(=0.33\)\label{fig-ranking}]{
  \includegraphics[width=0.98\linewidth]{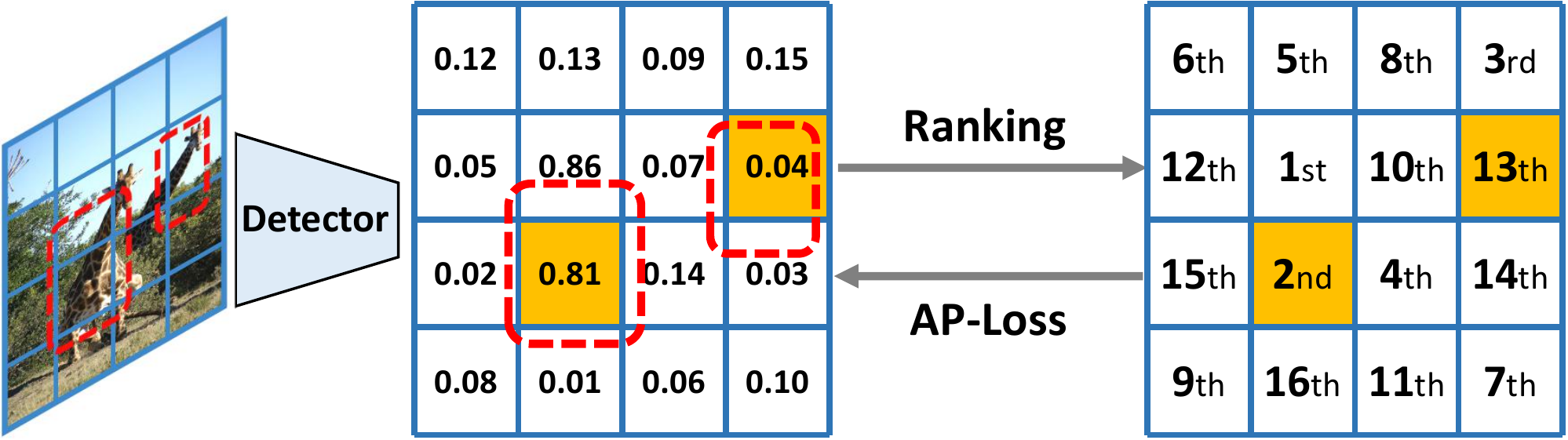}
  }
\vspace{-1mm}
  \caption{Dashed red boxes are the ground truth object boxes. The orange filled boxes and other blank boxes are anchors with positive and negative ground truth labels, repectively. (a) shows that the detection performance is poor but the classification accuracy is still high due to large number of true negatives. (b) shows the ranking metric AP can better reflect the actual condition as it does not suffer from the large number of true negatives, and is more intrinsically consistent with the detection task.}
\vspace{-3.0ex}
\label{fig-classification-ranking}
\end{figure}

Depending on whether detectors require additional modules to produce candidate object boxes, the general solutions can be divided into two flavours: one-stage detectors and two-stage detectors. Two-stage detectors first generate a limited number of object box proposals using category-independent proposal methods, such as selective search~\cite{uijlings2013selective}, CPMC~\cite{carreira2012cpmc}, multi-scale combinatorial grouping~\cite{arbelaez2014multiscale} and RPN~\cite{ren2015faster}. This is followed by adopting classification and localization tasks on those proposals. Typical works of this category include R-CNN~\cite{girshick2014rich} and its Fast~\cite{girshick2015fast}, and Faster~\cite{ren2015faster} variants, R-FCN~\cite{dai2016r}, FPN~\cite{lin2017feature}, Mask R-CNN~\cite{he2017mask}, and Cascade R-CNN~\cite{cai2018cascade}. On the contrary, one-stage detectors predict the object class directly from the densely pre-designed candidate boxes, also commonly known as anchors. Known works in literature include the popular YOLO series~\cite{redmon2016you,redmon2017yolo9000,redmon2018yolov3}, SSD~\cite{liu2016ssd}, DSSD~\cite{fu2017dssd}, DSOD~\cite{shen2017dsod}, RetinaNet~\cite{lin2018focal}, RefineDet~\cite{zhang2018single}, CornerNet~\cite{law2018cornernet} and M2Det~\cite{zhao2018m2det}. One-stage detectors generally compute much faster than two-stage detectors but consequently, they suffer from noticeable gaps in accuracy. Some studies~\cite{lin2018focal,shrivastava2016ohem,li2018gradient} pointed out that one possible reason lies in the extreme imbalance between foreground and background regions, which causes class bias during optimization of the classification task, which in turn impacts detection performance. The classification metric could be very high for a trivial solution which predicts negative label (i.e. background) for almost all candidate boxes (a likely situation in large images with few objects), while the detection performance remains poor. \autoref{fig-classification} illustrates one such example.
In this case, the accuracy metric is high due to the large number of true negatives, while detection performance is obviously poor. For instance, the detector can completely miss an object, predicting a false positive but still garnering high accuracy score due to the sheer number of true negatives.

To tackle this issue in one-stage object detectors, some works introduce new classification losses such as balanced loss~\cite{redmon2016you,redmon2017yolo9000}, Focal-Loss~\cite{lin2018focal}, as well as tailored training methods such as Online Hard Example Mining (OHEM)~\cite{liu2016ssd,shrivastava2016ohem}. These losses model each sample (anchor box) independently, and attempt to re-weight the foreground and background samples within the classification loss to cater for the imbalanced condition; this is done without considering the relationship or distribution among different samples. Besides, the designed balance weights are hand-crafted hyper-parameters that do not generalize well across datasets and detectors. In other words, it is hard to determine the degree of importance of samples, or how `easy' the samples should be ignored in a universal sense, and vice versa. We argue that more crucially, the extreme class imbalance enlarges the gap between classification task and detection task. Thus, the methods that offer simplistic modification of classification losses or frameworks to address the imbalanced conditions are far from optimal and limited mostly by their generalization ability. In this paper, instead of modifying the classification loss, we propose to frame this as a ranking task, where the ranking task is notably useful in tackling imbalance issue as shown in \cite{cortes2004auc,cruz2016tackling,cruz2017combining,natole2019stochastic}, in which the associated ranking loss explicitly models sample relationships, and is less sensitive to the ratio of positive and negative samples. We further choose the ranking metric AP (Average Precision)~\cite{everingham2015pascal,lin2014microsoft,salton1986introduction} as our target loss due to the guiding intuition that AP is also the evaluation metric for object detection, as shown in \autoref{fig-ranking}. In object detection, the AP metric evaluates detection results by considering both precision and recall at different thresholds. Compared to the accuracy metric (depicted in \autoref{fig-classification}), the AP metric does not suffer from a substantially large number of true negatives, which reflects the actual situation of the task much better. In addition to AP, another ranking metric AUC~\cite{cortes2004auc} can also handle the imbalance issue to some extent and is studied in our experiments.

\begin{figure*}[ht]
\centering
\small
  \includegraphics[width=0.88\linewidth]{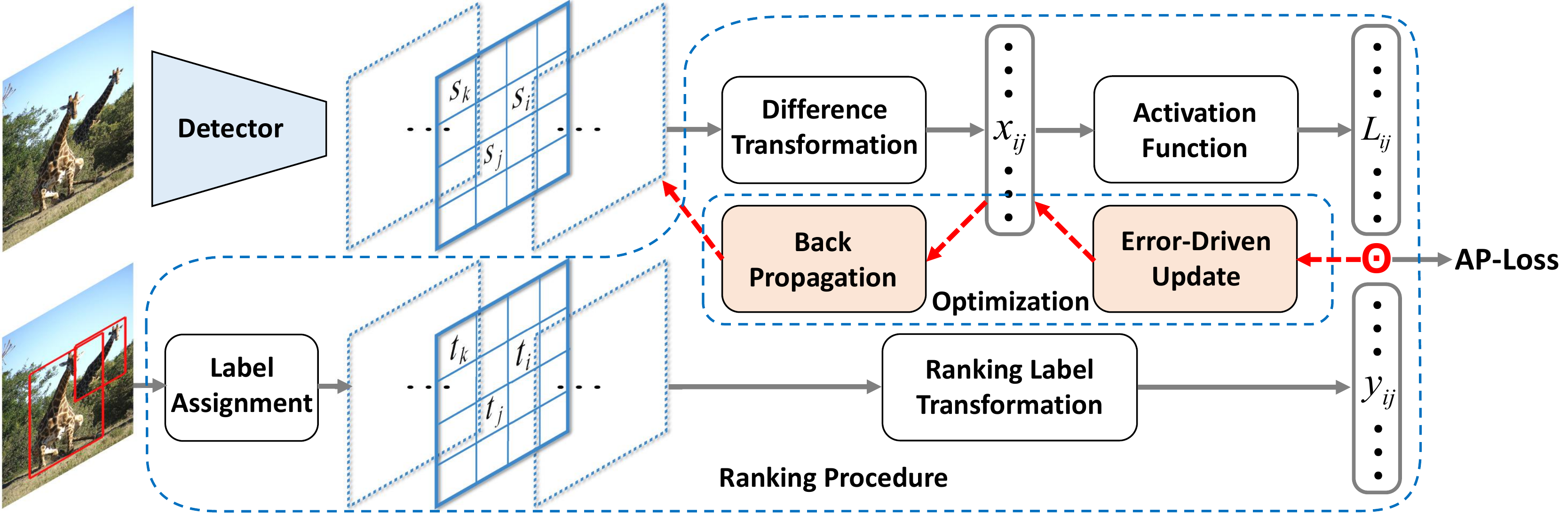}
\vspace{-0mm}
  \caption{Overall framework of the proposed approach. We replace the classification-task in one-stage detectors with a ranking task, where the ranking procedure produces the primary terms of AP-loss and the corresponding label vector. The optimization algorithm is based on an error-driven learning scheme combined with backpropagation. The localization-task branch is not shown here due to no modification.}
\vspace{-2.0ex}
\label{fig-overall}
\end{figure*}

Optimization of the AP-loss is considered a non-trivial problem. Due to its non-differentiability, standard gradient descent methods are not amenable for AP-loss. Worser still, due to its non-decomposability~\cite{song2016training}, the AP-loss cannot be expressed as the sum over the output units of network which makes it impossible to optimize each unit separately. Besides, one stage object detectors are commonly trained in large scale datasets (\textit{e.g.} PASCAL VOC~\cite{everingham2015pascal} and COCO~\cite{lin2014microsoft}) with very densely designed anchors. This indicates that a large number of training samples are involved in the training phase. Such large-scale training with deep networks requires the losses to be suitable for SGD (or its variants) based optimization. In short, optimizing the AP-loss for object detection requires effective way to handle the non-differentiability and non-decomposability issues as well as good scalability in handling large training sample sizes.

\subsection{Existing handling of AP-loss}
There are three known aspects of how issues concerning the AP-loss have been handled:
\begin{enumerate}
    \item
    AP based loss has been studied within structured SVM models~\cite{yue2007support,mohapatra2014efficient}. The structured SVM model~\cite{tsochantaridis2005large} is proposed to predict general structured output labels. Specifically, the structured model predicts a confidence score for each possible label where the label with highest score is selected as the output. In a ranking task, the predicted label is exactly the \textit{rank} of all input samples, and the score is generated by a hand-picked discriminant function. Note that such methods are restricted to linear SVM model so the performance is limited to data in simple manifolds.
    \item
    A structured hinge loss~\cite{Mohapatra_2018_CVPR} can be used to optimize the AP-loss. The optimization target is similar to that of the previous one, but more precisely, it functions by maximizing the margin between the ground truth label and the most violated label (\textit{i.e.} the label predicted with loss-augmented inference). This method is no longer restricted to linear models and can be implemented on some general but more complex learning models like neural networks using gradient descent. However, this method does not directly optimize AP-loss itself but rather an upper bound of it. It is still unknown whether such hinge relaxation with some specific discriminant function is near-optimal in a ranking task.
    \item
    Approximate gradient methods~\cite{song2016training,henderson2016end} are proposed with the aim of optimizing the AP-loss. \cite{song2016training} adopts an expectation taken over the underlying data distribution to smoothen the objective function, and its gradient can then be approximately estimated. \cite{henderson2016end} adopts an envelope function as the linear estimation of the original loss. This envelope function is similar in shape as the original loss function while having a well-defined gradient that is non-zero almost everywhere. These two methods both approximate the AP-loss with some smooth functions through different techniques, so that gradient descent can be used on it. Nevertheless, these methods are less efficient and easily fall into local optimum even for the case of linear models. The non-convexity and non-quasiconvexity of the AP-loss still presents an open problem that warrants further consideration.
\end{enumerate}

\subsection{Contributions of this work}
This manuscript is an extension of our CVPR 2019 paper~\cite{chen2019towards}, we make the following contributions.
We address the imbalance issue by replacing the classification task in one-stage detectors with a ranking task, so that the class imbalance problem can be handled with Average Precision (AP)-loss, a rank-based loss. To enable this to work, we propose a novel error-driven learning algorithm to effectively optimize the non-differentiable AP-based objective function.
More specifically, some extra transformations are added to the score output of one-stage detector to obtain the AP-loss, which includes a linear transformation that transforms the scores to pairwise differences, and a non-linear and non-differentiable ``activation function'' that transforms the pairwise differences to the primary terms of the AP-loss. This way, the AP-loss can be obtained by the dot product between the primary terms and the label vector. It is worth noting that the difficulty for using gradient method on the AP-loss lies in passing gradients through the non-differentiable activation function.
Inspired by the perceptron learning algorithm~\cite{rosenblatt1957perceptron}, we adopt an error-driven learning scheme to directly pass the update signal through the non-differentiable activation function. Different from gradient method, our learning scheme gives each variable an update signal proportional to the error it makes. Then, we adopt the backpropagation algorithm to transfer the update signal to the weights of neural network. We theoretically and experimentally prove that the proposed optimization algorithm does not suffer from the non-differentiability and non-convexity of the objective function. An acceleration strategy is also proposed to make the algorithm more practical. The experimental results show that AP-loss has better generalization ability to handle different imbalance conditions and stronger robustness against adversarial perturbations.
We also observe that the AP-loss can be used in two-stage detectors with detailed results and discussions shown in the Supplementary. The main contributions of this paper are summarized as follows:
\begin{itemize}
\vspace{-0mm}
\setlength{\topsep}{0pt}
\setlength{\itemsep}{1pt}
\setlength{\parskip}{1pt}
\item We propose a novel framework in one-stage object detectors which adopts the ranking loss to handle the class imbalance issue.
\item We propose an error-driven learning algorithm that can efficiently optimize the non-differentiable and non-convex AP-based objective function with both theoretical and experimental verifications.
\item We show notable performance improvement with the proposed method on state-of-the-art one-stage detectors over different kinds of classification-losses without changing the model architecture.
\item AP-loss demonstrates to be more robust against different kinds of adversarial perturbations and noises than Focal-loss and other balanced-loss solutions.
\end{itemize}

%-------------------------------------------------------------------------
\section{Related Work}

{\bf \noindent One-stage detectors:}
In object detection, the one-stage approaches have relatively simpler architecture and higher efficiency than two-stage approaches. OverFeat~\cite{sermanet2013overfeat} is one of the first CNN-based one-stage detectors. Thereafter, different designs of one-stage detectors are proposed, including SSD~\cite{liu2016ssd}, YOLO~\cite{redmon2016you}, DSSD~\cite{fu2017dssd} and DSOD~\cite{shen2017dsod,li2018tiny}. These methods demonstrate good processing efficiency as one-stage detectors, but generally yield lower accuracy than two-stage detectors.
After that, RetinaNet~\cite{lin2018focal} narrows down the performance gap (especially on the challenging COCO benchmark~\cite{lin2014microsoft}) between one-stage approaches and two-stage approaches with some innovative designs on loss function and network architectures. More recently, several works further improve the one-stage detection framework through different aspects. RefineDet~\cite{zhang2018single} proposes a two-step cascaded regression framework to first generate refined anchors and then detect objects based on them. DES~\cite{Zhang_2018_CVPR} adopts a semantic segmentation branch and a global activation module to enriched the semantics of object detection features. PFPN~\cite{kim2018parallel} presents a parallel feature transformation pipeline to generate features with consistent abstraction levels. RFBNet~\cite{Liu_2018_ECCV} proposes a receptive field block that takes relationship between size and eccentricity to enhance the feature discriminability and robustness. Therefore, with the competitive accuracy performance, higher efficiency and simpler mechanism, one-stage methods stand at an important position in object detection.

{\bf \noindent Imbalance Issue in detection networks:}
As commonly known, the performance of one-stage detectors benefits much from densely designed anchors, which introduce extreme imbalance between foreground and background samples. Several studies are proposed to address this challenge, including OHEM~\cite{liu2016ssd,shrivastava2016ohem}, Focal-Loss~\cite{lin2018focal}, A-Fast-RCNN~\cite{wang2017fast}, Gradient Harmonized Mechanism (GHM)~\cite{li2018gradient}, Libra R-CNN~\cite{pang2019libra}, DR-loss~\cite{qian2019dr}. More complex techniques have been proposed in the past years as surveyed in \cite{oksuz2019imbalance}. OHEM~\cite{liu2016ssd,shrivastava2016ohem} selects the top-\(k\) hardest samples in each training iteration and only enables loss and backpropagation on these hard samples. Focal-Loss~\cite{lin2018focal} modifies the cross-entropy loss using additional modulating factors, which include constant factor for positive-negative samples and polynomial factor for hard-easy samples. A-Fast-RCNN~\cite{wang2017fast} further improves the hard example mining technique, by generating high quality hard positive samples via adversarial learning. GHM~\cite{li2018gradient} proposes a gradient harmonizing mechanism to smooth the gradient contribution based on the their distribution, which works similar to Focal-loss with easy samples, \textit{i.e.} both down-weighting the gradient from easy samples, while the difference is GHM also down-weights the hard samples in some degrees. However, the validity of assumption ``more uniform contribution of gradient is better'' is still unknown in general condition. Libra R-CNN~\cite{pang2019libra} studies imbalance issue in three levels - sample level, feature level, objective level, and proposed solutions for each level separately. Nevertheless, there are two hurdles that are still open to discussion. Firstly, hand-crafted hyper-parameters for weight balance do not generalize well across datasets and detectors. Secondly, the relationship among sample anchors is far from well modeled.

{\bf \noindent AP as a loss for Object Detection:}
Average Precision (AP) is widely used as the evaluation metric in many tasks such as object detection~\cite{everingham2015pascal} and information retrieval~\cite{salton1986introduction}.
However, AP is far from a good and common choice as an optimization goal in object detection due to its non-differentiability and non-convexity.
Some methods have been proposed to optimize the AP-loss in object detection, such as AP-loss in the linear structured SVM model~\cite{yue2007support,mohapatra2014efficient}, structured hinge loss as upper bound of the AP-loss~\cite{Mohapatra_2018_CVPR}, approximate gradient methods~\cite{song2016training,henderson2016end},
reinforcement learning to fine-tune a pre-trained object detector with AP based metric~\cite{rao2018learning}.
Although these methods give valuable results in optimizing the AP-loss, their performances are still limited due to the intrinsic limitations.
In details, the proposed approach differs from them in 4 aspects.
(1) \cite{yue2007support} uses cutting plane algorithm with tolerance~\cite{tsochantaridis2005large} to optimizing the structured SVM model, \cite{mohapatra2014efficient} accelerates it by proposing a efficient algorithm for finding the most violated label. Note that both of them only work for linear SVM model, while our approach can be used for any differentiable linear or non-linear models such as neural networks.
(2) \cite{Mohapatra_2018_CVPR} generalize the structured SVM model to neural networks with structured hinge loss. Our approach directly optimizes the AP-loss while \cite{Mohapatra_2018_CVPR} introduces notable loss gap after relaxation.
(3) \cite{song2016training,henderson2016end} use expectation and envelope function to approximate the original objective function respectively. The approximate function is smooth so that the gradient can be readily estimated. However, such approximations preserve the non-convexity and non-quasiconvexity of AP-loss, which make the gradient descent less efficient. Our approach dose not approximate the gradient and dose not suffer from the non-convexity of objective function as in \cite{song2016training,henderson2016end}.
(4) \cite{rao2018learning} adopts policy gradient to fine tune the pre-trained detector with AP metric. Thus the AP optimization procedure is built upon the classification based training in a stage-by-stage way. Our approach can train the detectors in an end-to-end way, while \cite{rao2018learning} cannot.

{\bf \noindent Perceptron Learning Algorithm:}
The core of our optimization algorithm is the ``error-driven update'' which is generalized from the perceptron learning algorithm~\cite{rosenblatt1957perceptron}, and helps overcome the difficulty of the non-differentiable objective functions. The perceptron is a simple artificial neuron using the Heaviside step function as the activation function.
The learning algorithm was first invented by Frank Rosenblatt~\cite{rosenblatt1957perceptron}. As the Heaviside step function in perceptron is non-differentiable, it is not amenable for gradient method.
Instead of using a surrogate loss like cross-entropy, the perceptron learning algorithm employs an error-driven update scheme directly on the weights of neurons. This algorithm is guaranteed to converge in finite steps if the training data is linearly separable.
Further works like~\cite{krauth1987learning,anlauf1989adatron,wendemuth1995learning} have studied and improved the stability and robustness of the perceptron learning algorithm.

%------------------------------------------------------------------------
\section{Method}
We aim to replace the classification task with AP-loss based ranking task in one-stage detectors such as RetinaNet~\cite{lin2018focal} and SSD~\cite{liu2016ssd}.
\autoref{fig-overall} shows the two key components of our approach, \textit{i.e.}, the ranking procedure and the error-driven optimization algorithm.
Below, we will first present how AP-loss is derived from traditional score output.
Then, we will introduce the error-driven optimization algorithm.
Finally, we also present the theoretical analyses of the proposed optimization algorithm and outline the training details.
Note that all changes are made on the loss part of the classification branch without changing the backbone model and localization branch.

%------------------------------------------------------------------------
\subsection{Ranking Task and AP-Loss}
\subsubsection{Ranking Task}

\begin{figure}[t]
\small
\centering
\subfloat[]{
  \includegraphics[width=0.31\linewidth]{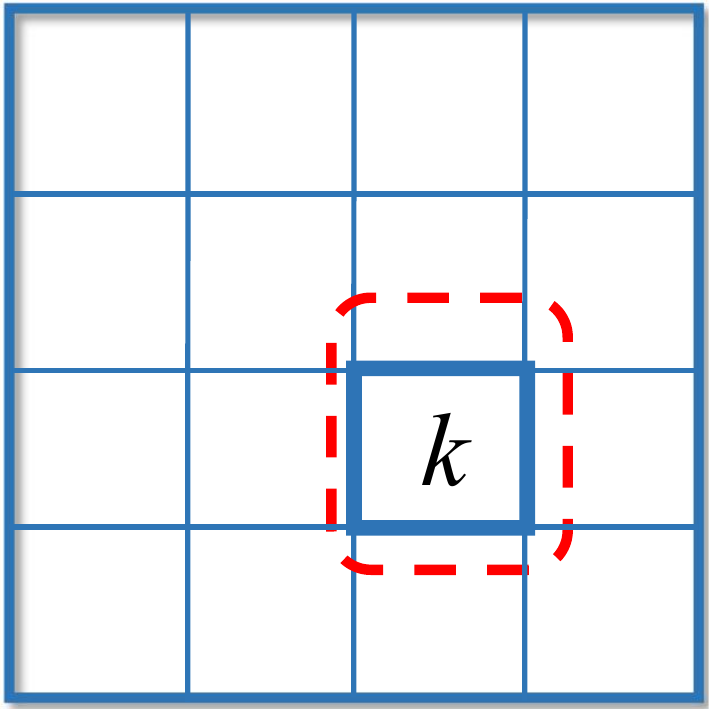}
  }
\quad
\subfloat[]{
  \includegraphics[width=0.51\linewidth]{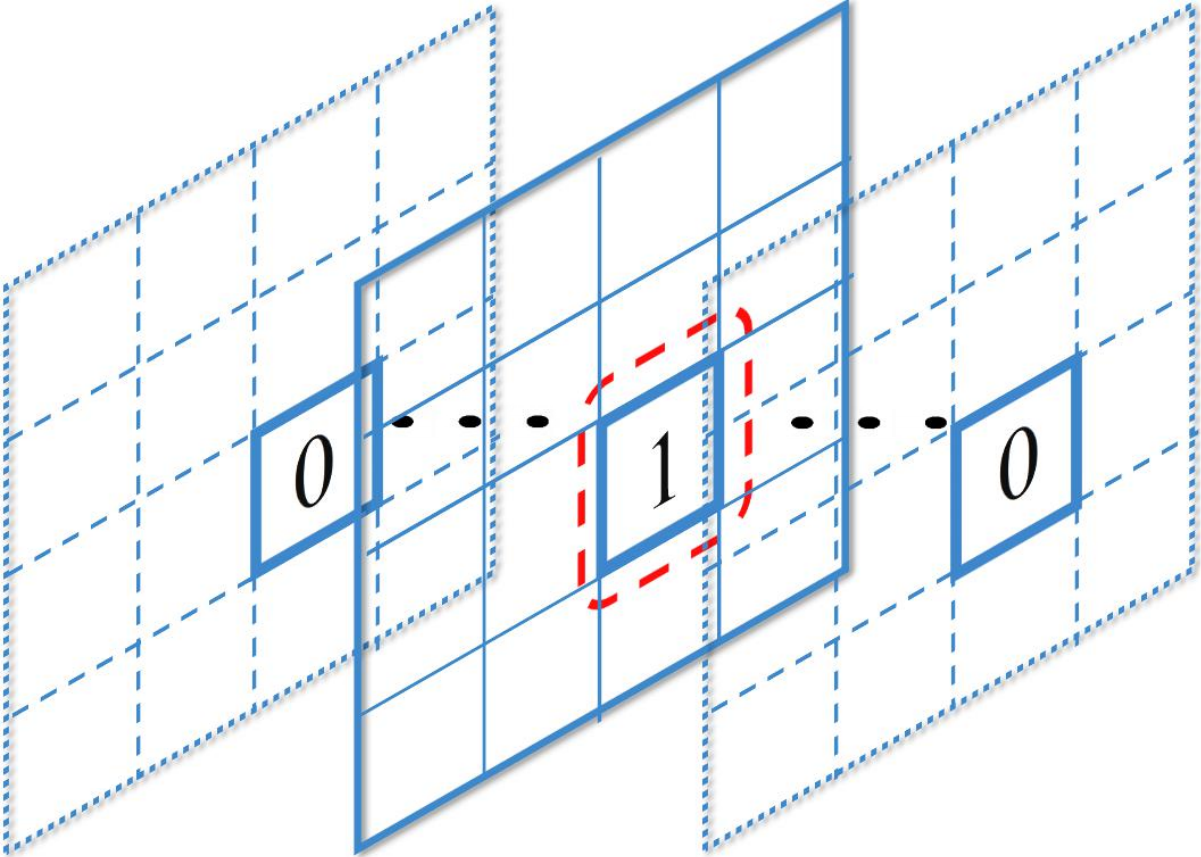}
  }
\vspace{-1ex}
  \caption{Comparison of label assignments. The dashed red box is the ground truth box with class $k$.
  (a) In traditional classification task of one-stage detectors, the anchor is assigned a foreground label $k$.
  (b) In our ranking task framework, the anchor replicates $K$ times, and assign the $k$-th anchor to label $1$, others 0.}
\label{fig-labelassignment}
\vspace{-2ex}
\end{figure}

In traditional one-stage detectors, given input image $I$, suppose the pre-defined boxes (also called anchors) set is $B$, each box $b_i \in B$ will be assigned a label $t_i \in \{-1,0,1,\ldots,K\}$ based on ground truth and the IoU strategy~\cite{girshick2015fast,ren2015faster}, where label $1\sim K$ means the object class ID, label ``0'' means background and label ``$-1$'' means ignored boxes. During training and testing phase, the detector outputs a score-vector $(s_i^0, \cdots, s_i^K)$ for each box $b_i$.

In our framework, instead of one box with $K+1$ dimensional score predictions,
we replicate each box $b_i$ for $K$  times to obtain $b_{ik}$ where $k=1,\cdots,K$, and the $k$-th box is responsible for the $k$-th class.
Each box $b_{ik}$ will be assigned a label $t_{ik} \in \{-1,0,1\}$ through the same IoU strategy (label $-1$ for not counted into the ranking loss).
Therefore, in the training and testing phase, the detector will predict only one scalar score $s_{ik}$ for each box $b_{ik}$.
\autoref{fig-labelassignment} illustrates our label formulation and the difference to traditional case. Note that this label assignment strategy is similar to that in RetinaNet~\cite{lin2018focal} since we both use binary labels in the multi-class object detection, while the difference is we further transform it into ranking labels.

The ranking task dictates that every positive boxes should be ranked higher than all the negative boxes w.r.t their scores.
Note that AP of our ranking result is computed over the scores from all classes.
This is slightly different from the evaluation metric meanAP for object detection systems, which computes AP for each class and obtains the average value.
We compute AP this way because the score distribution should be unified for all classes while ranking each class separately cannot achieve this goal.

\subsubsection{AP-Loss}
For simplicity, we still use $B$ to denote the anchor box set after replication, and $b_i$ to denote the $i$-th anchor box without the replication subscript. Each box $b_i$ thus corresponds to one scalar score $s_i$ and one binary label $t_i$.
Some transformations are required to formulate a ranking loss as illustrated in \autoref{fig-overall}.
First, the \textit{difference transformation} transfers the score $s_i$ to the difference form
\begin{equation}
\small
\forall i,j, \,\,\, x_{ij}=-(s(b_i; \bm{\theta})-s(b_j; \bm{\theta})) = -(s_i - s_j)
\end{equation}
where $s(b_i; \bm{\theta})$ is a CNN based score function with weights $\bm{\theta}$ for box \(b_i\).
The \textit{ranking label transformation} transfers labels $t_i$ to the corresponding pairwise ordering form
\begin{equation}
\small
\forall i,j, \,\,\, y_{ij} = \mathbf{1}_{t_i=1,t_j=0}
\end{equation}
where $\mathbf{1}$ is a indicator function which equals to 1 only if the subscript condition holds (\textit{i.e.}, $t_i=1,t_j=0$), otherwise $0$.
Then, we define an vector-valued \textit{activation function} $\bm{L}(\cdot)$ to produce the primary terms of the AP-loss as
\begin{equation}
\small
L_{ij}(\bm{x})=\frac{H(x_{ij})}{1+\sum_{k\in \mathcal{P}\cup\mathcal{N},k\neq i} H(x_{ik})} = L_{ij}
\label{Lij}
\end{equation}
where \(H(\cdot)\) is the Heaviside step function:
\begin{equation}
\small
H(x)=\left\{
\begin{aligned}
0& & x < 0 \\
1& & x\geq 0
\end{aligned}
\right.
\end{equation}

A ranking is denoted as \textit{proper ranking} when there are no two samples scored equally (\textit{i.e.}, $\forall i\neq j, \,\, s_i \neq s_j$). Without loss of generality, we will treat all rankings as a proper ranking by breaking ties arbitrarily. Now, we can formulate the AP-loss $\mathcal{L}_{AP}$ as
\begin{equation}
\small
\begin{split}
&\mathcal{L}_{AP} = 1- \text{AP} = 1-\frac{1}{|\mathcal{P}|}\sum_{i\in \mathcal{P}}\frac{rank^{+}(i)}{rank(i)} \\
& = 1-\frac{1}{|\mathcal{P}|} \sum_{i\in \mathcal{P}}\frac{1+\sum_{j\in \mathcal{P},j\neq i}H(x_{ij})}{1+\sum_{j\in \mathcal{P},j\neq i} H(x_{ij})+\sum_{j\in \mathcal{N}}H(x_{ij})} \\
& = \frac{1}{|\mathcal{P}|}\sum_{i\in \mathcal{P}}\sum_{j\in \mathcal{N}} L_{ij} = \frac{1}{|\mathcal{P}|} \sum_{i,j} L_{ij} \cdot y_{ij} = \frac{1}{|\mathcal{P}|} \langle \bm{L}(\bm{x}), \bm{y} \rangle
\end{split}
\label{Lap_1}
\end{equation}
where \(rank(i)\) and \(rank^{+}(i)\) denote the ranking position of score \(s_i\) among all valid samples and positive samples respectively, $\mathcal{P}=\{i | t_i=1\}$, $\mathcal{N}=\{i | t_i=0\}$, $|\mathcal{P}|$ is the size of set $\mathcal{P}$,
$\bm{L}$ and $\bm{y}$ are vector form for all $L_{ij}$ and $y_{ij}$ respectively, $\langle, \rangle$ means dot-product of two input vectors.
Note that $\bm{x},\bm{y}, \bm{L} \in \mathbb{R}^d$, where $d = (|\mathcal{P}|+|\mathcal{N}|)^2$.

Finally, the optimization problem can be written as:
\begin{equation}
\small
\min_{\bm{\theta}} \mathcal{L}_{AP}(\bm{\theta}) = 1-\text{AP}(\bm{\theta})=\frac{1}{|\mathcal{P}|} \langle \bm{L}(\bm{x}(\bm{\theta})), \bm{y}\rangle
\end{equation}
where $\bm{\theta}$ denotes the weights of detector model.
As the activation function $\bm{L}(\cdot)$ is non-differentiable, a novel optimization/learning scheme is required instead of the standard gradient descent method.

Besides the AP metric, other ranking based metric can also be used to design the ranking loss for our framework.
One example is the AUC-loss~\cite{li2013learning,cortes2004auc} which measures the area under ROC curve for ranking purpose, and has a slightly different ``activation function'' as
\begin{equation}
\small
L_{ij}'(\bm{x})=\frac{H(x_{ij})}{|\mathcal{N}|}
\end{equation}
As AP is consistent with the evaluation metric of the object detection task, we argue that AP-loss is intuitively more suitable than AUC-loss for this task,
and will provide empirical study in our experiments.

%------------------------------------------------------------------------
\subsection{Optimization Algorithm}
\subsubsection{Error-Driven Update}

Recalling the perceptron learning algorithm, the update for input variable is ``error-driven'', which means the update is directly derived from the difference between desired output and current output. We adopt this idea and further generalize it to accommodate the case of activation function with vector-valued input and output. Suppose $x_{ij}$ is the input and $L_{ij}$ is the current output, the update for $x_{ij}$ is thus
\begin{equation}
\small
\Delta x_{ij}=L^{*}_{ij}-L_{ij}
\label{error-driven}
\end{equation}
where \(L^{*}_{ij}\) is the desired output. Note that the AP-loss achieves its minimum possible value $0$ when each term $L_{ij}\cdot y_{ij} = 0$.
There are two cases. If $y_{ij}=1$, we should set the desired output $L^{*}_{ij}=0$.
If $y_{ij}=0$, we do not care the update and set it to $0$, since it does not contribute to the AP-loss.
Consequently, the update can be simplified as
\begin{equation}
\small
\Delta x_{ij}=-L_{ij} \cdot y_{ij}
\label{eq:update-xij}
\end{equation}

\subsubsection{Backpropagation}
We now have the desired vector-form update $\Delta \bm{x}$, and then will find an update for model weights $\Delta \bm{\theta}$  which will produce most appropriate movement for $\bm{x}$.
We use dot-product to measure the similarity of successive movements, and regularize the change of weights (\textit{i.e.} $\Delta\bm{\theta}$) with $L_2$-norm based penalty term. The optimization problem can be written as:
\begin{equation}
\small
\arg\min_{\Delta \bm{\theta}} \{-\langle \Delta \bm{x}, \bm{x}(\bm{\theta}^{(n)}+\Delta \bm{\theta})-\bm{x}(\bm{\theta}^{(n)}) \rangle +\lambda \| \Delta \bm{\theta} \|_2^2\}
\end{equation}
where $\bm{\theta}^{(n)}$ denotes the model weights at the $n$-th step. With that, the first-order expansion of \(\boldsymbol{x}(\boldsymbol{\theta})\) is given by:
\begin{equation}
\small
\bm{x}(\bm{\theta})=\bm{x}(\bm{\theta}^{(n)})+\frac{\partial \bm{x}(\bm{\theta}^{(n)})}{\partial \bm{\theta}} \cdot (\bm{\theta}-\bm{\theta}^{(n)})
 + o(\|\bm{\theta}-\bm{\theta}^{(n)}\|)
\end{equation}
where $\partial \bm{x}(\bm{\theta}^{(n)})/ \partial \bm{\theta}$ is the Jacobian matrix of vector-valued function $\bm{x}(\bm{\theta})$ at $\bm{\theta}^{(n)}$.
Ignoring the high-order infinitesimal, we obtain the step-wise minimization process:
\begin{equation}
\small
\bm{\theta}^{(n+1)}-\bm{\theta}^{(n)}=\arg\min_{\Delta\bm{\theta}}\{-\langle \Delta \bm{x}, \frac{\partial \bm{x}(\bm{\theta}^{(n)})}{\partial \bm{\theta}} \Delta\bm{\theta} \rangle  + \lambda \| \Delta\bm{\theta}\|_2^2\}
\end{equation}
We can then easily obtain the optimal solution:
\begin{equation}
\small
\bm{\theta}^{(n+1)}=\bm{\theta}^{(n)}+\frac{1}{2\lambda}(\frac{\partial \bm{x}(\bm{\theta}^{(n)})}{\partial \bm{\theta}})^T \Delta \bm{x}
\end{equation}
According to the chain rule of derivative, we can directly implement it by setting the gradient of \(x_{ij}\) to \(-\Delta x_{ij}\) (c.f. \autoref{eq:update-xij}) and proceeding with backpropagation. Hence, the gradient for score \(s_i\) can be easily obtained by backward propagating the gradient through the difference transformation:
\begin{equation}
\small
\begin{split}
g_i= - \sum_{j,k} & \Delta x_{jk} \cdot \frac{\partial x_{jk}}{\partial s_i} = \sum_j \Delta x_{ij} - \sum_j \Delta x_{ji} \\
& = \sum_j L_{ji} \cdot y_{ji} - \sum_j L_{ij} \cdot y_{ij}.
\end{split}
\label{gradient}
\end{equation}

%------------------------------------------------------------------------
\subsection{Analyses}
\subsubsection{Convergence}
To better understand the characteristics of the AP-loss, we first provide a theoretical analysis on the convergence of the optimization algorithm, which is generalized from the convergence property of the original perceptron learning algorithm.
\begin{proposition}
The AP-loss optimizing algorithm is guaranteed to converge in finite steps if below conditions hold: \par
\noindent(1) the learning model is linear; \par
\noindent(2) the training data is linearly separable.
\label{convergence}
\end{proposition}
The proof of this proposition is provided in Appendix-1 of supplementary.
Although convergence is somewhat weak due to the need of strong conditions, it is \textbf{non-trivial} since the AP-loss function is not convex or quasiconvex even for the case of linear model and linearly separable data, so that gradient descent based algorithm may still fail to converge on a smoothed AP-loss function even under such strong conditions. One such example is presented in Appendix-2 of supplementary. It means that, under such conditions, our algorithm still optimizes better than the approximate gradient descent algorithm for AP-loss.
Furthermore, with some mild modifications, even though the training data is not separable, the accumulated AP-loss can also be bounded proportionally by the best performance of the learning model. More details are presented in Appendix-3 of supplementary.

\subsubsection{Consistency}
Besides convergence, we observed that the proposed optimization algorithm is inherently consistent with widely used classification-loss functions.
\begin{observation}
When the activation function \(L(\cdot)\) takes the form of softmax function and loss-augmented step function, our optimization algorithm can be expressed as the gradient descent algorithm on cross-entropy loss and hinge loss respectively.
\label{consistency}
\end{observation}
The detailed analysis of this observation is presented here:

{\bf \noindent Cross Entropy Loss:}
Consider the multi-class classification task. The outputs of neural network are \((x_1,\ldots, x_K)\) where \(K\) is the number of classes, and the ground truth label is \(y \in \{1,\ldots,K\}\). Using softmax as the activation function, we have:
\begin{equation}
\small
(L_1,\ldots,L_K) = softmax(\bm{x}) = (\frac{e^{x_1}}{\sum_i e^{x_i}},\ldots,\frac{e^{x_K}}{\sum_i e^{x_i}})
\end{equation}
The cross entropy loss is:
\begin{equation}
\small
\mathcal{L}_{ce}=-\sum_i \mathbf{1}_{y=i} \log(L_i)
\end{equation}
Hence the gradient of \(x_i\) is
\begin{equation}
\small
g_i=L_i-\mathbf{1}_{y=i}
\end{equation}
Note that \(g_i\) is ``error-driven'' with the desired output \(\mathbf{1}_{y=i}\) and current output \(L_i\). This form is consistent with our error-driven update scheme.

{\bf \noindent Hinge Loss:}
Consider the binary classification task. The output of neural network is \(x\), and the ground truth label is \(y \in \{1,2\}\). Define \((x_1,x_2)=(-x,x)\). Using loss-augmented step function as the activation function, we have:
\begin{equation}
\small
(L_1,L_2)=(H(x_1-1),H(x_2-1))
\end{equation}
where \(H(\cdot)\) is the Heaviside step function. The hinge loss is:
\begin{equation}
\small
\mathcal{L}_{hinge}=\mathbf{1}_{y=1} \max\{1-x_1,0\} + \mathbf{1}_{y=2} \max\{1-x_2,0\}
\end{equation}
Hence the gradient of \(x_i\) is
\begin{equation}
\small
g_i = \mathbf{1}_{y=i} \cdot (L_i-1)
\end{equation}
There are two cases. If \(y=i\), the gradient \(g_i\) is ``error-driven'' with the desired output \(1\) and current output \(L_i\). If \(y \neq i\), the gradient \(g_i\) equals zero, since \(x_i\) does not contribute to the loss. This form is consistent with our error-driven update scheme.

We argue that the observed consistency is on the basis of the ``error-driven'' property.
As is known, the gradients of those widely used loss functions are proportional to their prediction errors, where the prediction here refers to the output of activation function. In other words, their activation functions have a nice property: the vector field of prediction errors is conservative, allowing it being the gradient of some surrogate loss function.
However, our activation function does not have this property, which makes our optimization not able to express as gradient descent with any surrogate loss function.

\subsection{Details of Training Algorithm} \label{training-detail}
Apart from the theoretical analyses, the proposed algorithm still faces some practical challenges within the deep learning framework. In this section, we present several important details and advanced techniques to meet these challenges and thus improve the feasibility of the proposed algorithm with details summarized in Algorithm \ref{IAP}.
\subsubsection{Mini-batch Training} \label{minibatch_training}
The mini-batch training strategy is widely used in deep learning frameworks~\cite{he2016deep,liu2016ssd,lin2018focal} as it accounts for more stability than the case with batch size equal to 1.
The mini-batch training helps our optimization algorithm quite a lot for escaping the so-called ``score-shift'' scenario.
The AP-loss can be computed both from a batch of images and from a single image with multiple anchor boxes.
Consider an extreme case: our detector can predict perfect ranking in both image $I_1$ and image $I_2$, but the lowest score in image $I_1$ is even greater than the highest score in image $I_2$. There are ``score-shift'' between two images so that the detection performance is poor when computing AP-loss per-image.
Aggregating scores over images in a mini-batch can avoid such problem, so that the mini-batch training is crucial for good convergence and good performance.

\subsubsection{Piecewise Step function} \label{section:piecewise-step-function} \label{picewise_step_function}
\begin{figure}[t]
\centering
  \includegraphics[width=0.9\linewidth]{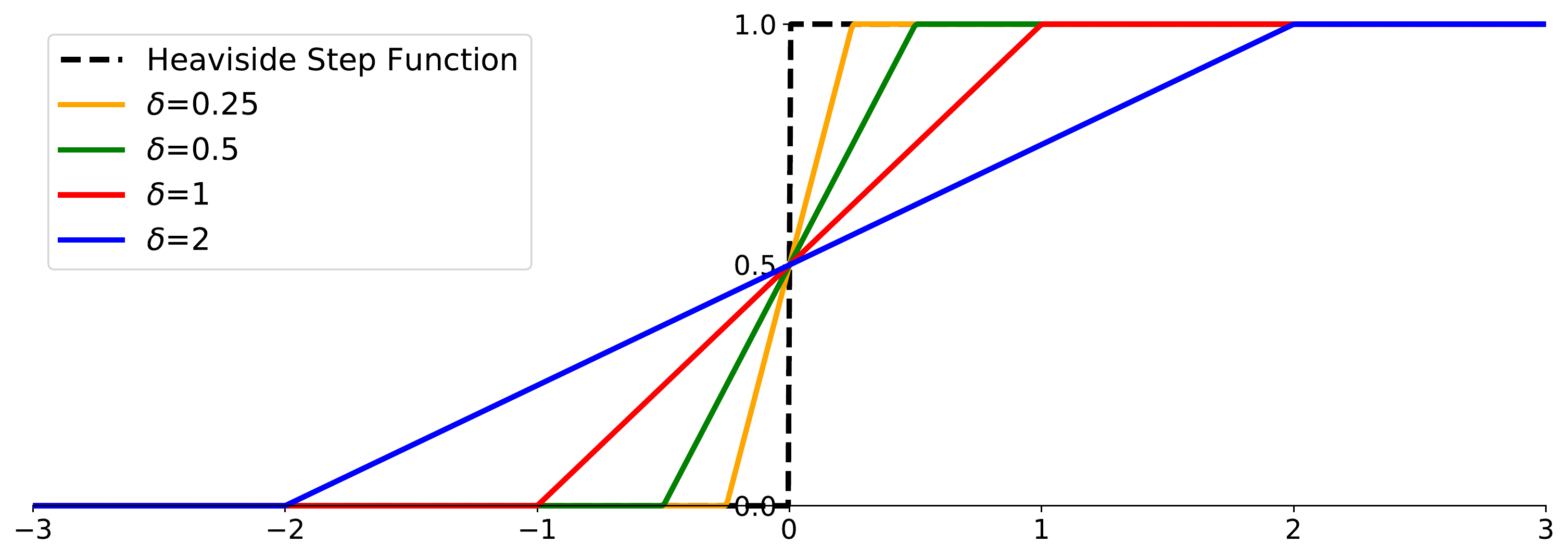}
\vspace{-2mm}
  \caption{Heaviside step function and piecewise step function. (Best viewed in color)}
\vspace{-3mm}
\label{fig-step_function}
\end{figure}

During early stage of training, scores $s_i$ are very close to each other (\textit{i.e.} almost all inputs to Heaviside step function $H(x)$ are near zero), so that a small change of input will cause a big output difference, which destabilizes the updating process. To tackle this issue, we replace $H(x)$ with a piecewise step function:
\begin{equation}
\small
f(x)=\left\{
\begin{aligned}
&0\, , & x < -\delta \\
&\frac{x}{2\delta}+0.5\, , & -\delta \leq x \leq \delta \\
&1\, , & \delta < x
\end{aligned}
\right.
\label{smooth}
\end{equation}
The piecewise step functions with different \(\delta\) are shown in \autoref{fig-step_function}. When \(\delta\) approaches \(+0\), the piecewise step function approaches the original step function. Note that \(f(\cdot )\) is only different from \(H(\cdot )\) near zero point. We argue that the precise form of the piecewise step function is not crucial. Other monotonic and symmetric smooth functions that only differs from \(H(\cdot )\) near zero point could be equally effective. The choice of $\delta$ relates closely to the weight decay hyper-parameter in CNN optimization. Intuitively, parameter $\delta$ controls the width of decision boundary between positive and negative samples. Smaller $\delta$ enforces a narrower decision boundary, which causes the weights to shrink correspondingly (similar effect to that caused by the weight decay). Further details are presented in the experiments.

\subsubsection{Interpolated AP} \label{interpolated_ap}
The interpolated AP~\cite{salton1986introduction} is widely adopted by many object detection benchmarks like PASCAL VOC~\cite{everingham2015pascal} and MS COCO~\cite{lin2014microsoft}.
The common justification for interpolating the precision-recall curve~\cite{everingham2015pascal} is ``to reduce the impact of 'wiggles' in the precision-recall curve, caused by small variations in the ranking of examples''.
Under the same consideration, we adopt the interpolated AP instead of the original version. Specifically, the interpolation is applied on $L_{ij}$ to make the precision at the $k$-th smallest positive sample monotonically increasing with \(k\) where the precision is $(1-\sum_{j\in \mathcal{N}} L_{ij})$ in which $i$ is the index of the $k$-th smallest positive sample. It is worth noting that the interpolated AP is a smooth approximation of the actual AP so that it is a practical choice to help to stabilize the gradient and to reduce the impact of 'wiggles' in the update signals. The details of the interpolated AP based algorithm is summarized in Algorithm \ref{IAP}.

\begin{table*}[t]
\small
\centering
\caption{Ablation experiments on Batch Size, \(\delta\) of Piecewise Step Function and Interpolated AP. Models are tested on VOC2007 {\tt test} set. The metric AP is averaged over multiple IoU thresholds of \(0.50:0.05:0.95\).}
\begin{tabular}{cccc}
\midrule[1pt]
Batch Size & AP & AP\(_{50}\) & AP\(_{75}\) \\
\midrule[1pt]
1 & 52.4 & 80.2 & 56.7 \\
2 & 53.0 & 81.7 & 57.8 \\
4 & 52.8 & 82.2 & 58.0 \\
8 & \textbf{53.1} & \textbf{82.3} & \textbf{58.1} \\
\midrule[1pt]
\label{voc-minibatch-training}
\end{tabular}
\quad\quad\quad\,\,\,\,
\begin{tabular}{cccc}
\midrule[1pt]
\(\delta\) & AP & AP\(_{50}\) & AP\(_{75}\) \\
\midrule[1pt]
0.25 & 50.2 & 80.7 & 53.6 \\
0.5 & 51.3 & 81.6 & 55.4 \\
1 & \textbf{53.1} & 82.3 & \textbf{58.1} \\
2 & 52.8 & \textbf{82.6} & 57.2 \\
4 & 50.5 & 80.2 & 54.7 \\
\midrule[1pt]
\label{voc-smoothing}
\end{tabular}
\quad\quad\quad\,\,
\begin{tabular}{cccc}
\midrule[1pt]
Interpolated  & AP & AP\(_{50}\) & AP\(_{75}\) \\
\midrule[1pt]
No & 52.6 & 82.2 & 57.1 \\
Yes & \textbf{53.1} & \textbf{82.3} & \textbf{58.1} \\
\midrule[1pt]
\label{voc-interpolated-ap}
\end{tabular}
\label{voc-abaltion-study}
\end{table*}

\begin{algorithm}[t]
\small
\caption{Mini-batch training for Interpolated AP}
\begin{algorithmic}[1]
  \Require All scores \(\{s_i\}\) and corresponding labels \(\{t_i\}\) in a mini-batch
  \Ensure Gradient of input \(\{g_i\}\)
  \State $\forall i, \,\,\, g_i \gets 0$
  \State $\text{MaxPrec} \gets 0$
  \State $\mathcal{P} \gets \{i \mid t_i=1\}, \,\,\, \mathcal{N} \gets \{i \mid t_i=0\}$
  \State $s_{\text{min}} \gets \min_{i\in \mathcal{P}}\{s_{i}\}$
  \State $\mathcal{\widehat{N}} \gets \{i\in \mathcal{N} \mid s_{i}> s_{\text{min}}-\delta\}$
  \State $O \gets argsort(\{s_i \mid i\in \mathcal{P}\}) $ \Comment{Indexes of scores sorted in ascending order}
  \For {$i \in O$}
    \State Compute \(x_{ij}=s_j-s_i\) for all \(j\in \mathcal{P}\cup\mathcal{\widehat{N}}\)
    \State Compute \(L_{ij}\) for all \(j\in \mathcal{\widehat{N}}\) \Comment{According to \autoref{Lij} and \autoref{smooth}}
    \State $\text{Prec} \gets 1-\sum_{j \in \mathcal{\widehat{N}}} L_{ij}$
    \If {$\text{Prec} \geq \text{MaxPrec}$}
      \State $ \text{MaxPrec} \gets \text{Prec}$
    \Else \Comment{Interpolation}
      \State $\forall j \in \mathcal{\widehat{N}}, \,\,\, L_{ij} \gets L_{ij} \cdot (1-\text{MaxPrec}) / (1-\text{Prec})$
    \EndIf
    \State $g_i \gets -\sum_{j\in \mathcal{\widehat{N}}} L_{ij}$ \Comment{According to \autoref{gradient}}
    \State $\forall j \in \mathcal{\widehat{N}}, \,\,\, g_j \gets g_j+L_{ij}$ \Comment{According to \autoref{gradient}}
  \EndFor
  \State $\forall i, \,\,\, g_i \gets g_i / |\mathcal{P}|$ \Comment{Normalization}
\end{algorithmic}
\label{IAP}
\end{algorithm}
\vspace{-4mm}

\subsubsection{Complexity and Acceleration Strategies} \label{complexity_and_acceleration}
The computation cost and memory cost for AP-loss are generally higher than other classification losses due to the need of pairwise differences computation. The time and space complexities are both \(O((|\mathcal{P}|+|\mathcal{N}|)^2)\) where \(|\mathcal{P}|+|\mathcal{N}|\) is normally about \(10^5\sim 10^6\) per image with RetinaNet. Such high complexities are intractable in a large-scale training task. Hence two techniques are proposed to tackle these issues.

\noindent\textbf{Loop on Positive Indices:} As shown in \autoref{Lap_1}, the AP-loss is equivalent to a inner product of \(\bm{L}\) and \(\bm{y}\), and we know that \(y_{ij}=1\) only for \(i\in \mathcal{P}, j\in \mathcal{N}\) otherwise \(y_{ij}=0\). Hence we only need compute \(L_{ij}\) with \(i\in \mathcal{P},j\in \mathcal{N}\). We implement it by adopting a loop on the indices of positive samples. In each iteration \(i\), we compute \(L_{ij}\) for all \(j\in \mathcal{N}\). The memory for \(\{x_{ij}, y_{ij}, L_{ij}\}\) can thus be released at the end of each single iteration, and only the gradient \(g_i\) for each score \(s_i\) is required for storing. Hence the time and space complexities can be reduced to \(O(|\mathcal{P}|\cdot(|\mathcal{P}|+|\mathcal{N}|))\) and \(O(|\mathcal{P}|+|\mathcal{N}|)\) respectively. In addition, note that \(|\mathcal{N}|\gg |\mathcal{P}|\) always holds in one-stage detectors. This implies that the time and memory complexities tends towards \(O(|\mathcal{P}|\cdot|\mathcal{N}|)\) and \(O(|\mathcal{N}|)\) respectively.

\noindent\textbf{Non-Trivial Negative Samples:} Note that if a negative sample \(j\in \mathcal{N}\) satisfies \(x_{ij}=s_j-s_i\leq -\delta\) for all \(i\in \mathcal{P}\), then we have \(\widetilde{H}(x_{ij})=0\) for all \(i\in \mathcal{P}\). This implies that we can ignore such trivial negative samples in the AP-loss computation and only focus on the non-trivial ones. A simple process can be used to find all non-trivial negative samples before the AP-loss computation; which involves two steps: \textbf{(1)} Find the minimal score \(s_{\text{min}}\) among all positive samples, \textbf{(2)} Find all negative samples that have larger scores than \(s_{\text{min}}-\delta\). This process only has a time complexity of \(O(|\mathcal{P}|+|\mathcal{N}|)\). Hence, the total time complexity is \(O(|\mathcal{P}|\cdot|\mathcal{\mathcal{\widehat{N}}}|+|\mathcal{N}|)\) where \(\mathcal{\widehat{N}}\) denotes the set of non-trivial negative indices. If we have \(|\mathcal{\widehat{N}}|\ll|\mathcal{N}|\), the AP-loss computation can be significantly accelerated due to the benefit of reduced size of negative samples. Note that it may not improve the speed in the initial stage of training, since almost all negative samples are non-trivial. However, the algorithm will progressively become faster along with the increasing performance of detector. We verify this hypothesis in our experiments.

%-------------------------------------------------------------------------
\section{Experiments}
\subsection{Experimental Settings}
We evaluate the proposed method on the state-of-the-art one-stage detectors RetinaNet~\cite{lin2018focal} and SSD~\cite{liu2016ssd}. The implementation details are the same as in~\cite{lin2018focal} and~\cite{liu2016ssd} respectively unless explicitly stated. Our experiments are performed on two benchmark datasets: PASCAL VOC~\cite{everingham2015pascal} and MS COCO~\cite{lin2014microsoft}. The PASCAL VOC dataset has 20 classes, with VOC2007 containing 9,963 images for train/val/test and VOC2012 containing 11,530 for train/val. The MS COCO dataset has 80 classes, containing 123,287 images for train/val. We conduct the experiments on a workstation with two NVidia TitanX GPUs.

\subsubsection{Baseline Model}
{\bf \noindent RetinaNet:}
We use ResNet~\cite{he2016deep} as the backbone model which is pre-trained on the ImageNet-1k classification dataset~\cite{deng2009imagenet}. At each level of FPN~\cite{lin2017feature}, the anchors have 2 sub-octave scales (\(2^{k/2}\), for \(k\leq 1\)) and 3 aspect ratios [0.5, 1, 2]. We set \(\delta=1\) in \autoref{smooth}. We fix the batch normalization layers to be frozen in training phase. We adopt the mini-batch training on 2 GPUs with 8 images per GPU.  We adopt the same data augmentation strategies as~\cite{liu2016ssd}, while do not use any data augmentation during testing phase. For all the ablation experiments, the input image is fixed to 512\(\times\)512 in training phase, while we maintain the original aspect ratio and resize the image to ensure the shorter side with 600 pixels in testing phase. For fairer comparison, in Section \ref{section-benchmark-results} Benchmark Results, we instead use 512\(\times\)512/800\(\times\)800 in training, while resize the shorter side to 500/800 in testing (for the detector AP-loss500/800 respectively). We apply the non-maximum suppression with IoU of 0.5 for each class. Weight decay of 0.0001 and momentum of 0.9 are used.

{\bf \noindent SSD:}
We use VGG-16~\cite{simonyan2014very} as the backbone model which is pre-trained on the ImageNet-1k classification dataset~\cite{deng2009imagenet}. We use {\tt conv4\_3}, {\tt conv7}, {\tt conv8\_2}, {\tt conv9\_2}, {\tt conv10\_2}, {\tt conv11\_2}, {\tt conv12\_2} to predict both location and their corresponding confidences. An additional convolution layer is added after {\tt conv4\_3} to scale the feature. The associated anchors are the same as that designed in~\cite{liu2016ssd}. In testing phase, the input image is fixed to 512\(\times\)512. For Focal-loss with SSD, we observe that the hyper-parameters \(\gamma=1,\alpha=0.25\) lead to a much better performance than the original settings in~\cite{lin2018focal} which are \(\gamma=2,\alpha=0.25\). Hence we evaluate the Focal-loss with new \(\gamma\) and \(\alpha\) in our experiments on SSD. Other details are similar to that for RetinaNet.

\subsubsection{Dataset}
{\bf \noindent PASCAL VOC:} When evaluated on the VOC2007 {\tt test} set, models are trained on the VOC2007 and VOC2012 {\tt trainval} sets. When evaluated on the VOC2012 {\tt test} set, models are trained on the VOC2007 and VOC2012 {\tt trainval} sets puls the VOC2007 {\tt test} set. Similar to the evaluation metrics used in the MS COCO benchmark, we also report the AP averaged over multiple IoU thresholds of \(0.50:0.05:0.95\). All evaluated models are trained for 160 epochs with an initial learning rate of 0.001 which is then divided by 10 at 110 epochs and again at 140 epochs.

{\bf \noindent MS COCO:} All models are trained on the widely used {\tt trainval35k} set (80k train images and 35k subset of val images), and tested on {\tt minival} set (5k subset of val images) or {\tt test-dev} set. We train the networks for 100 epochs with an initial learning rate of 0.001 which is then divided by 10 at 60 epochs and again at 80 epochs.

\subsection{Ablation Study}
We first investigate the impact of our design settings of the proposed framework. We fix the ResNet-50 and VGG-16 as backbone for RetinaNet and SSD respectively. We conduct several controlled experiments on PASCAL VOC2007 {\tt test} set (and COCO {\tt minival} if stated) for the ablation study.

\begin{table*}[t]
\small
\centering
\caption{Comparison through different training losses. Models are tested on VOC2007 {\tt test} and COCO {\tt minival} sets. The metric AP is averaged over multiple IoU thresholds of \(0.50:0.05:0.95\).}
\setlength{\tabcolsep}{4mm}{
\begin{tabular}{cccccccccc}
\midrule[1pt]
\multirow{2}*{Training Loss} & \multicolumn{3}{c}{RetinaNet + PASCAL VOC} & \multicolumn{3}{c}{RetinaNet + COCO} & \multicolumn{3}{c}{SSD + PASCAL VOC} \\
\cmidrule{2-4}\cmidrule{5-7}\cmidrule{8-10}
 & AP & AP\(_{50}\) & AP\(_{75}\) & AP & AP\(_{50}\) & AP\(_{75}\) & AP & AP\(_{50}\) & AP\(_{75}\)\\
\midrule[1pt]
CE-Loss + OHEM & 49.1 & 81.5 & 51.5 & 30.8 & 50.9 & 32.6 & 43.6 & 76.0 & 44.7\\
Focal Loss & 51.3 & 80.9 & 55.3 & 33.9 & 55.0 & 35.7 & 39.3 & 69.9 & 38.0 \\
\midrule[1pt]
AUC-Loss & 49.3 & 79.7 & 51.8 & 25.5 & 44.9 & 26.0 & 33.8 & 63.7 & 31.5\\
AP-Loss & \textbf{53.1} & \textbf{82.3} & \textbf{58.1} & \textbf{35.0} & \textbf{57.2} & \textbf{36.6} & \textbf{45.2} & \textbf{77.3} & \textbf{47.3}\\
\midrule[1pt]
\end{tabular}}
\label{base-loss}
\end{table*}

\begin{figure*}[t!]
\small
\centering
  \includegraphics[width=0.186\linewidth]{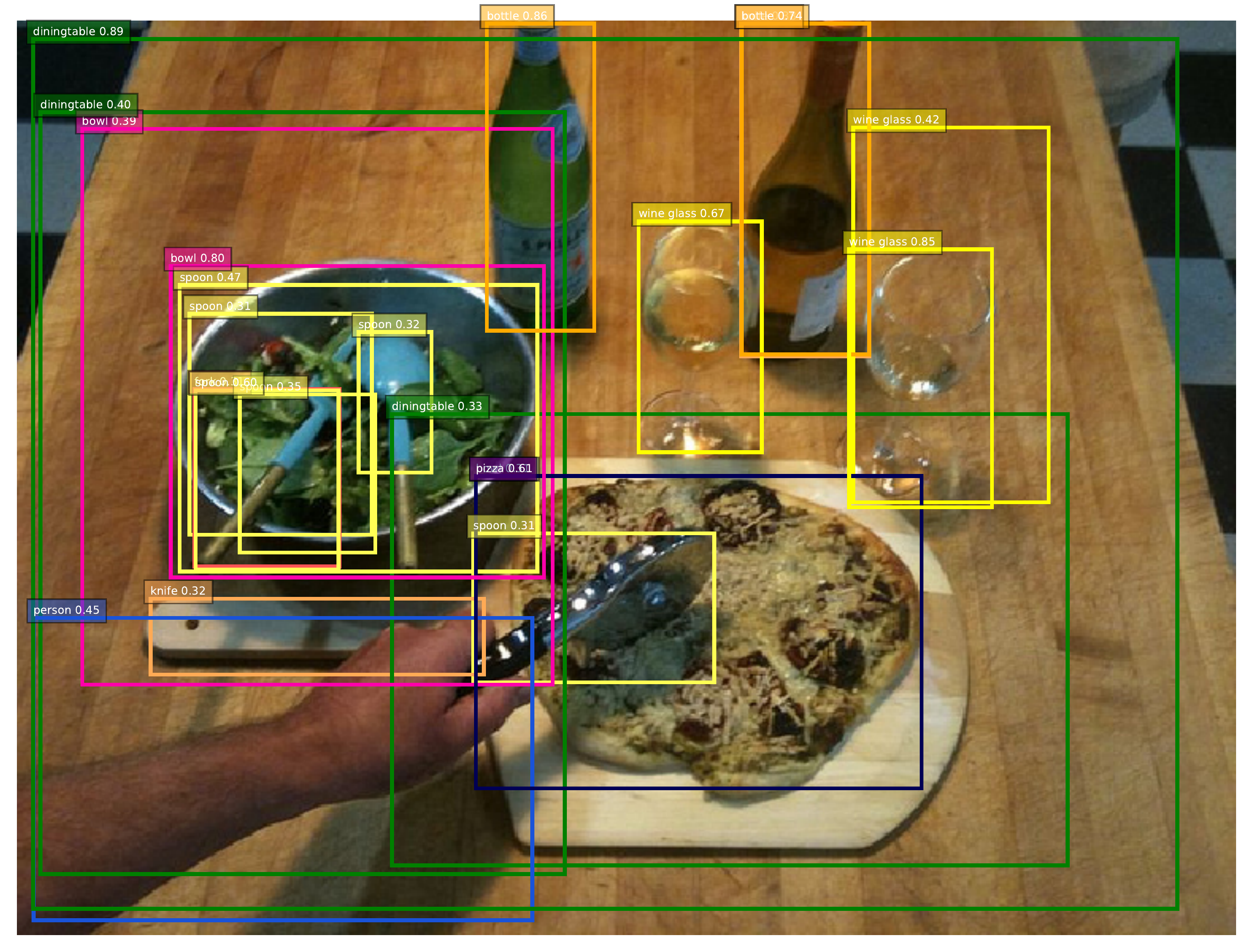}
  \includegraphics[width=0.21\linewidth]{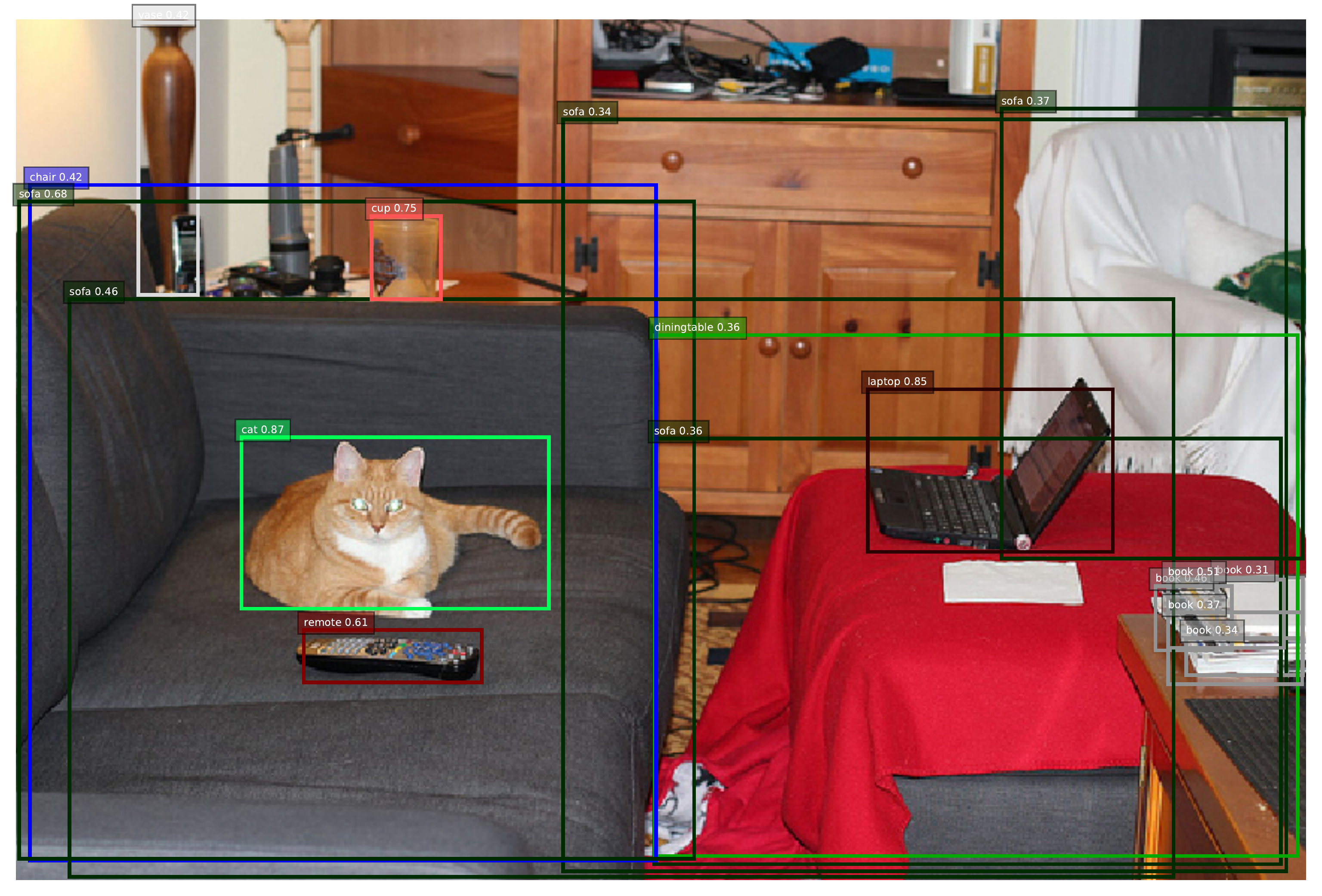}
  \includegraphics[width=0.21\linewidth]{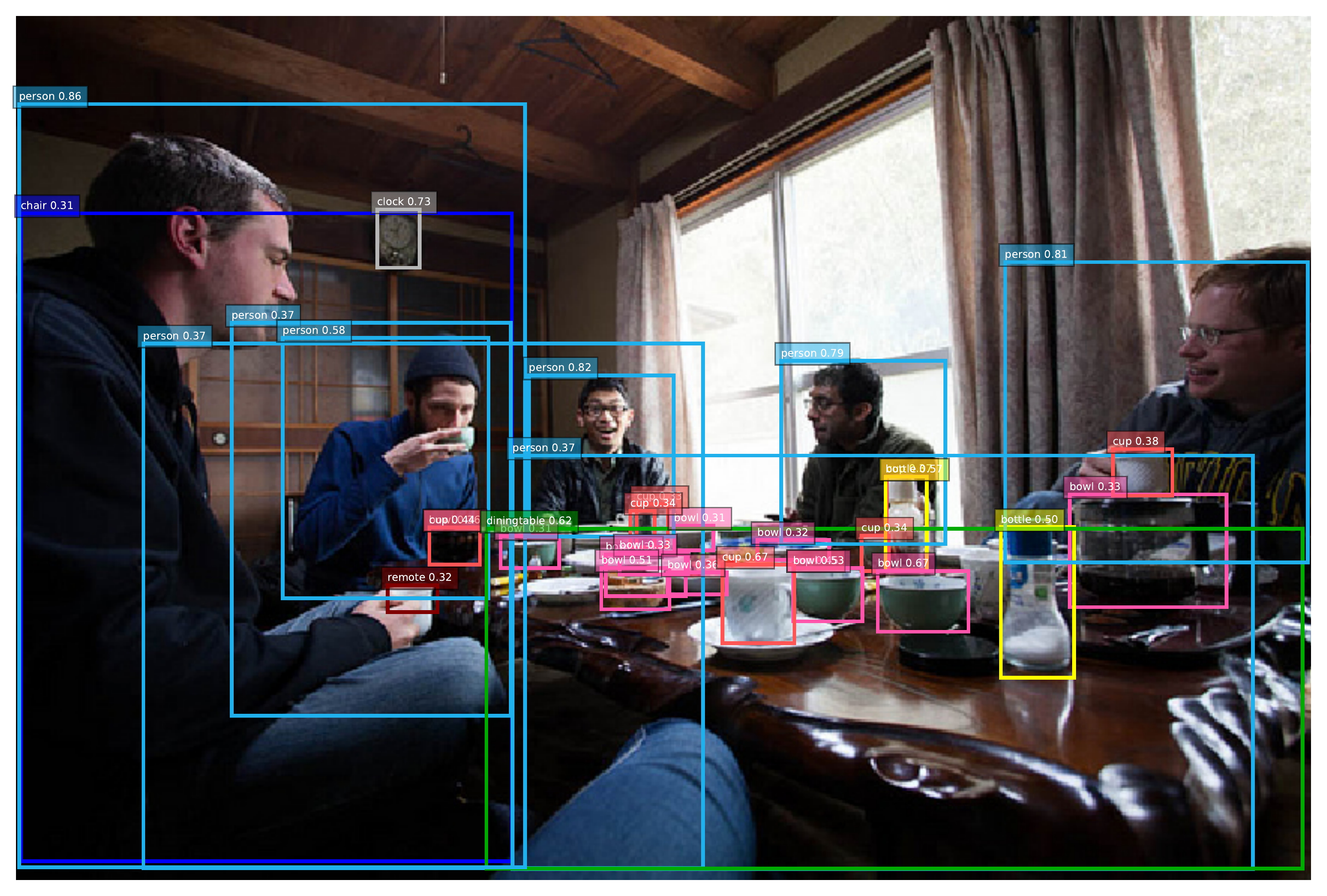}
  \includegraphics[width=0.186\linewidth]{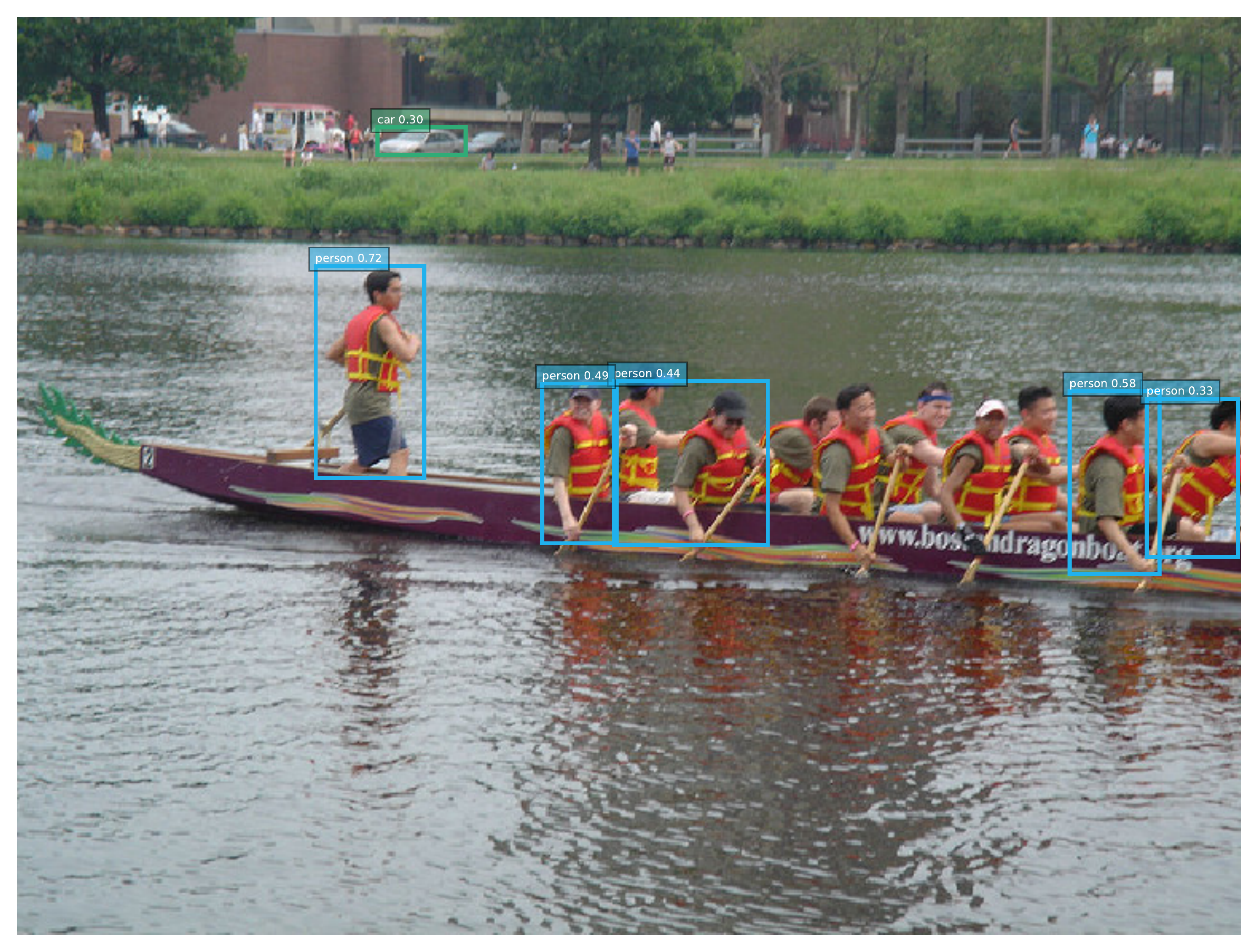}
  \includegraphics[width=0.186\linewidth]{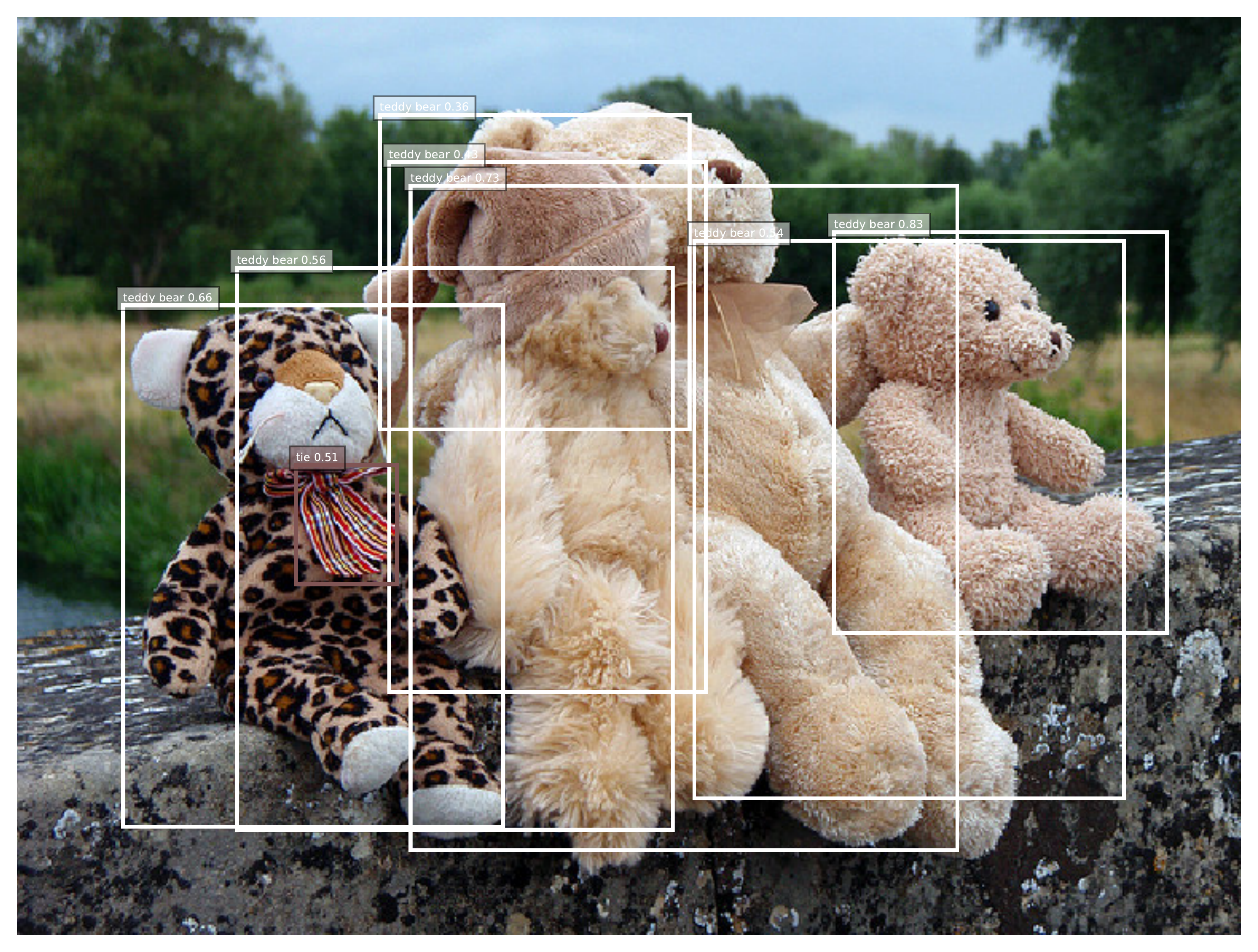}
\par
  \includegraphics[width=0.186\linewidth]{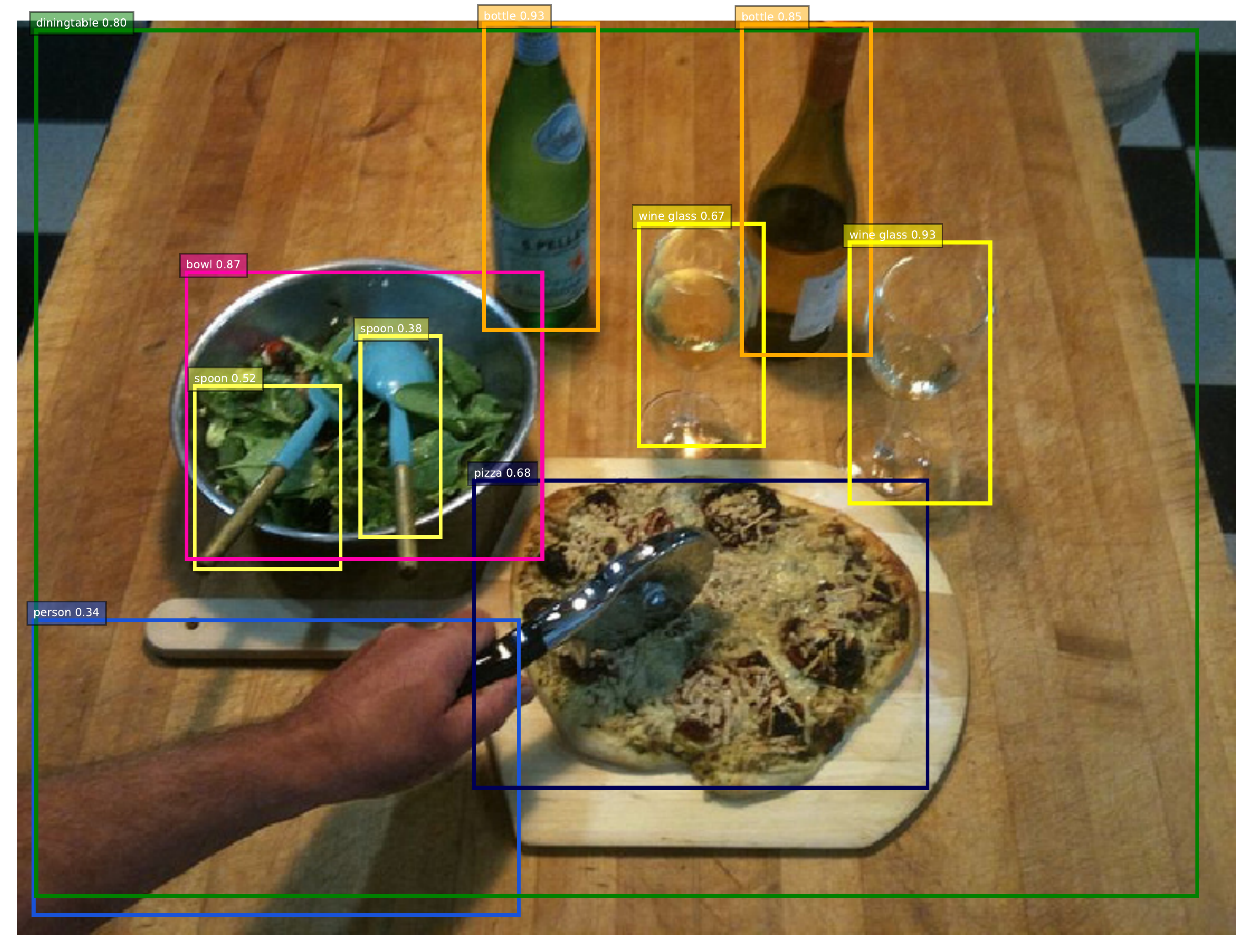}
  \includegraphics[width=0.21\linewidth]{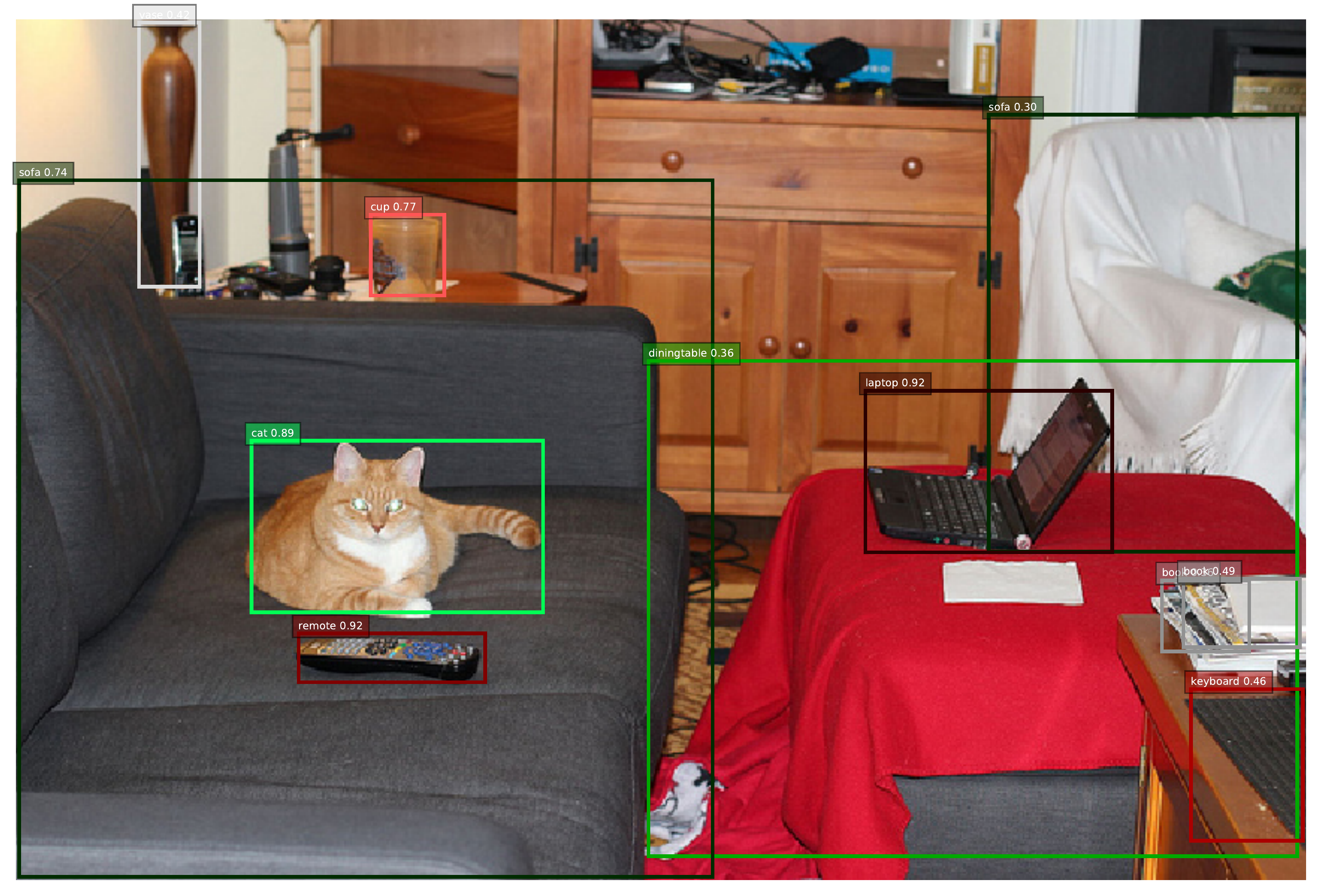}
  \includegraphics[width=0.21\linewidth]{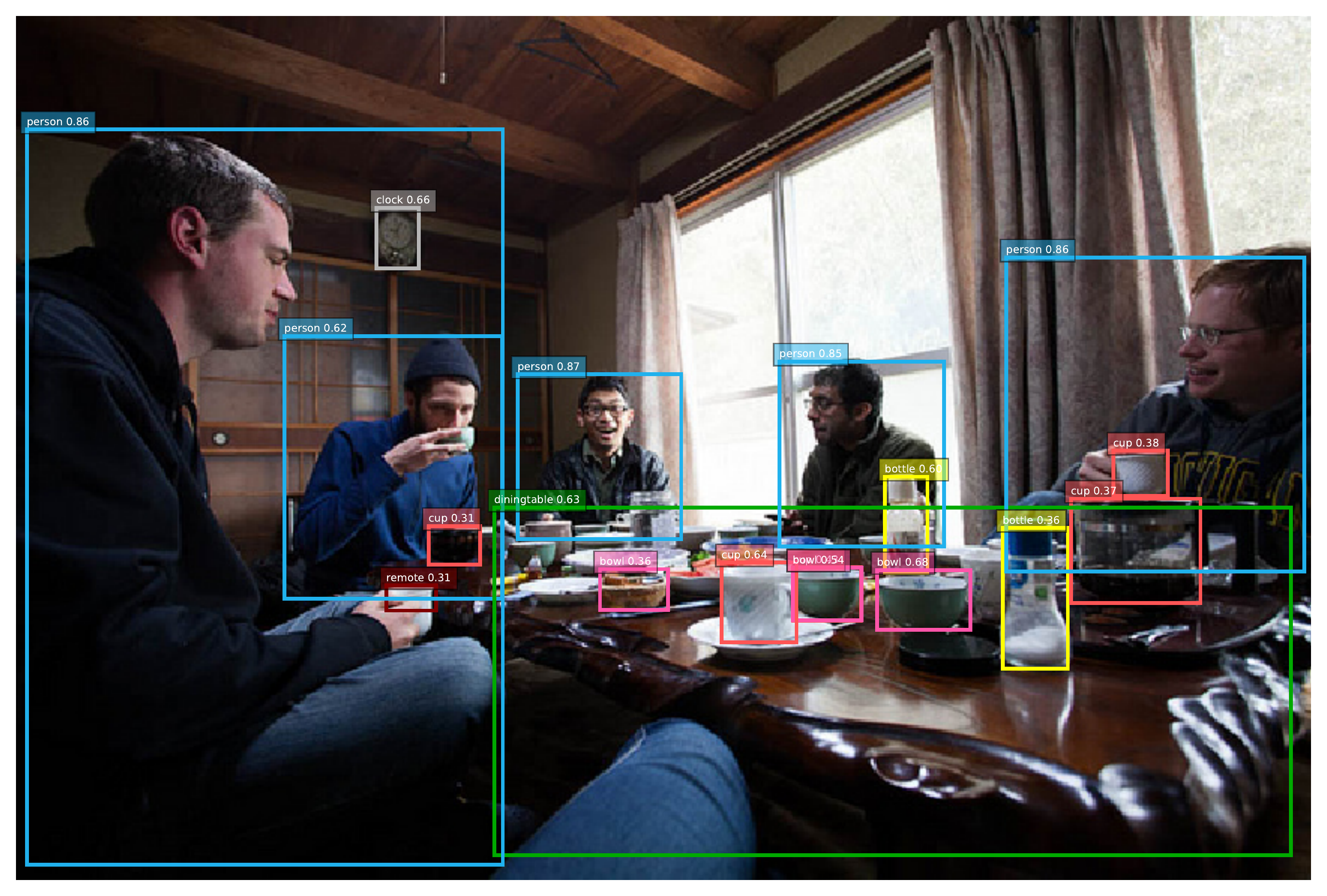}
  \includegraphics[width=0.186\linewidth]{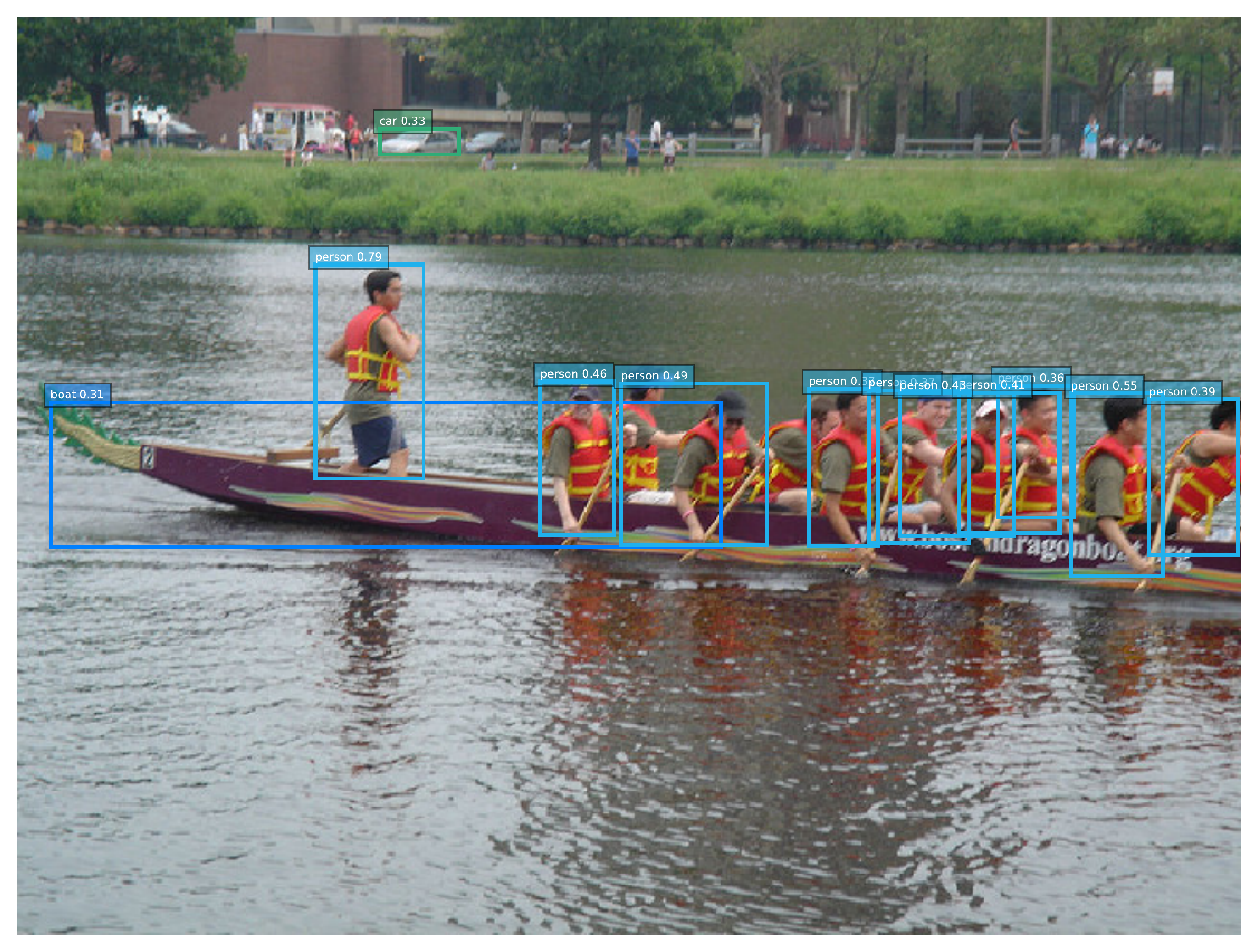}
  \includegraphics[width=0.186\linewidth]{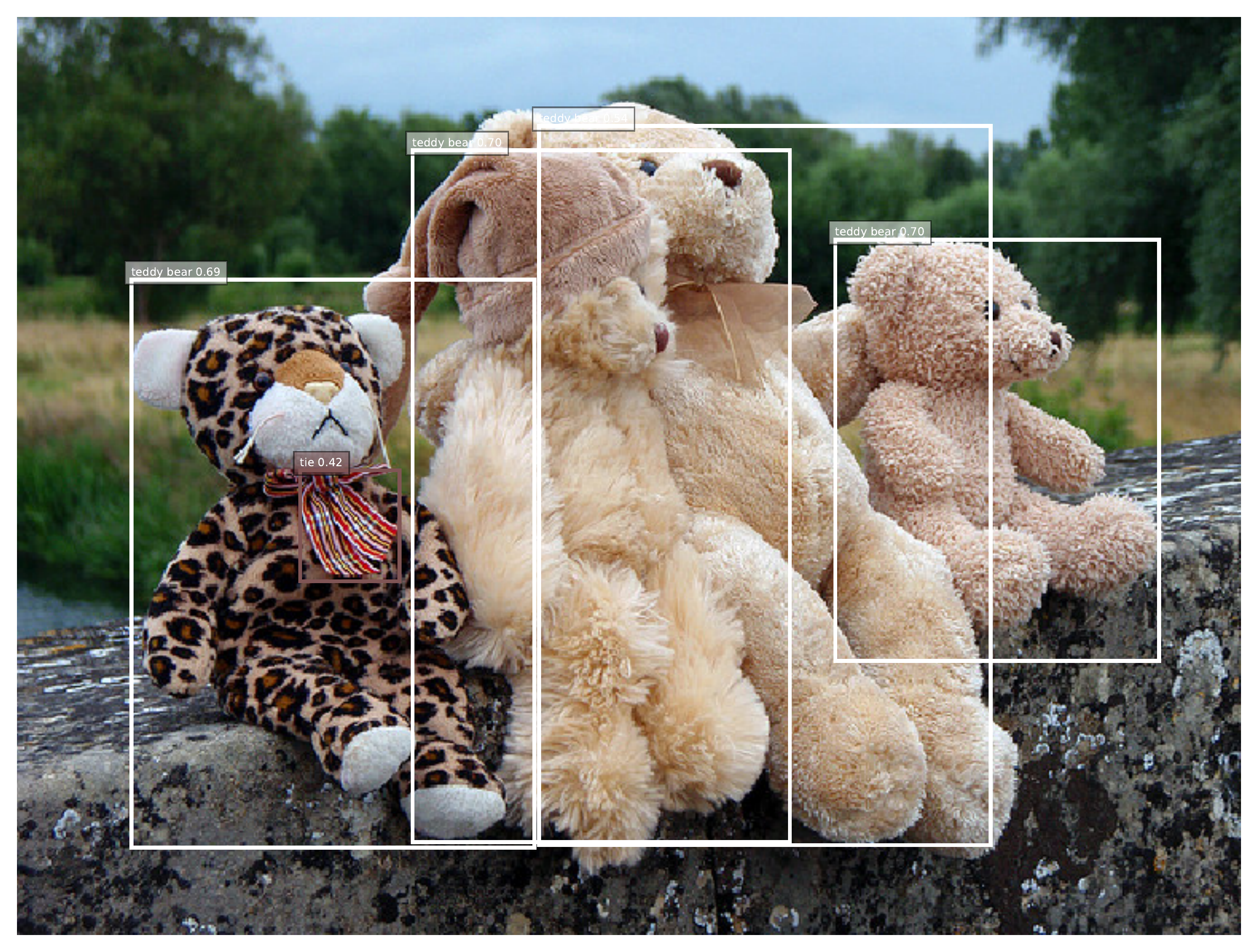}
\vspace{-4mm}
  \caption{Detection results on COCO dataset. Top: Baseline results by RetinaNet with Focal-loss. Bottom: Our results by RetinaNet with AP-loss.}
\vspace{-1mm}
\label{fig-result}
\end{figure*}

\subsubsection{Comparison on Different Parameter Settings}
Here we study the impact of the practical modifications introduced in Section \ref{minibatch_training}, \ref{picewise_step_function}, \ref{interpolated_ap}. All results are based on RetinaNet and shown in \autoref{voc-abaltion-study}.

{\bf \noindent Mini-batch Training:}
First, we study the mini-batch training, and report detector results at different batch-size. Note that the learning rate is changed linearly according to the batch-size, as suggested in \cite{goyal2017accurate}. Besides, the number of training epochs is fixed across different batch-sizes. We note that larger batch-size (\textit{i.e.} 8) outperforms all the other smaller batch-size. This verifies our previous hypothesis that large mini-batch training helps to eliminate the ``score-shift'' from different images, and thus stabilizes the AP-loss through robust gradient calculation (more detailed results are shown in Supplementary). Hence, a batch-size of 8 is used in our further studies.
We also observe that the ``score-shift'' is not a very serious problem in practice (results with batch-size 1 are still acceptable). This is likely because a lot of objects and background areas have very similar appearances in different images, thus the score distributions are similar among images that have similar objects and background areas. Besides, we conjecture that the neural network model itself may have regularization and generalization ability to some degree, which indicates that model training with only in-image ranking objective may also learn some general patterns that can properly rank objects and backgrounds across different images.

{\bf \noindent Piecewise Step Function:}
Second, we study the piecewise step function, and report detector performance on the piecewise step function with different \(\delta\). As mentioned before, we argue that the choice of \(\delta\) is trivial and is dependent on other network hyper-parameters such as weight decay. Smaller \(\delta\) makes the function sharper, which yields unstable training at initial phase. Larger \(\delta\) makes the function deviate from the properties of the original AP-loss, which also worsens the performance. \(\delta=1\) is a good choice we used in our further studies. Besides, as an alternative choice, the sigmoid function can also be tuned on the scale parameter to achieve a competitive result of 53.5\% AP (more detailed results are shown in Supplementary). This verifies our previous statement that the precise form of the piecewise step function is not crucial.

{\bf \noindent Interpolated AP:}
Third, we study the impact of interpolated AP in our optimization algorithm and report the results. Marginal benefits are observed for interpolated AP over standard AP, so we use interpolated AP in all the following studies.

\begin{figure}[t]
\small
\centering
\subfloat[\label{fig-performance_curve}]{
  \includegraphics[width=0.49\linewidth]{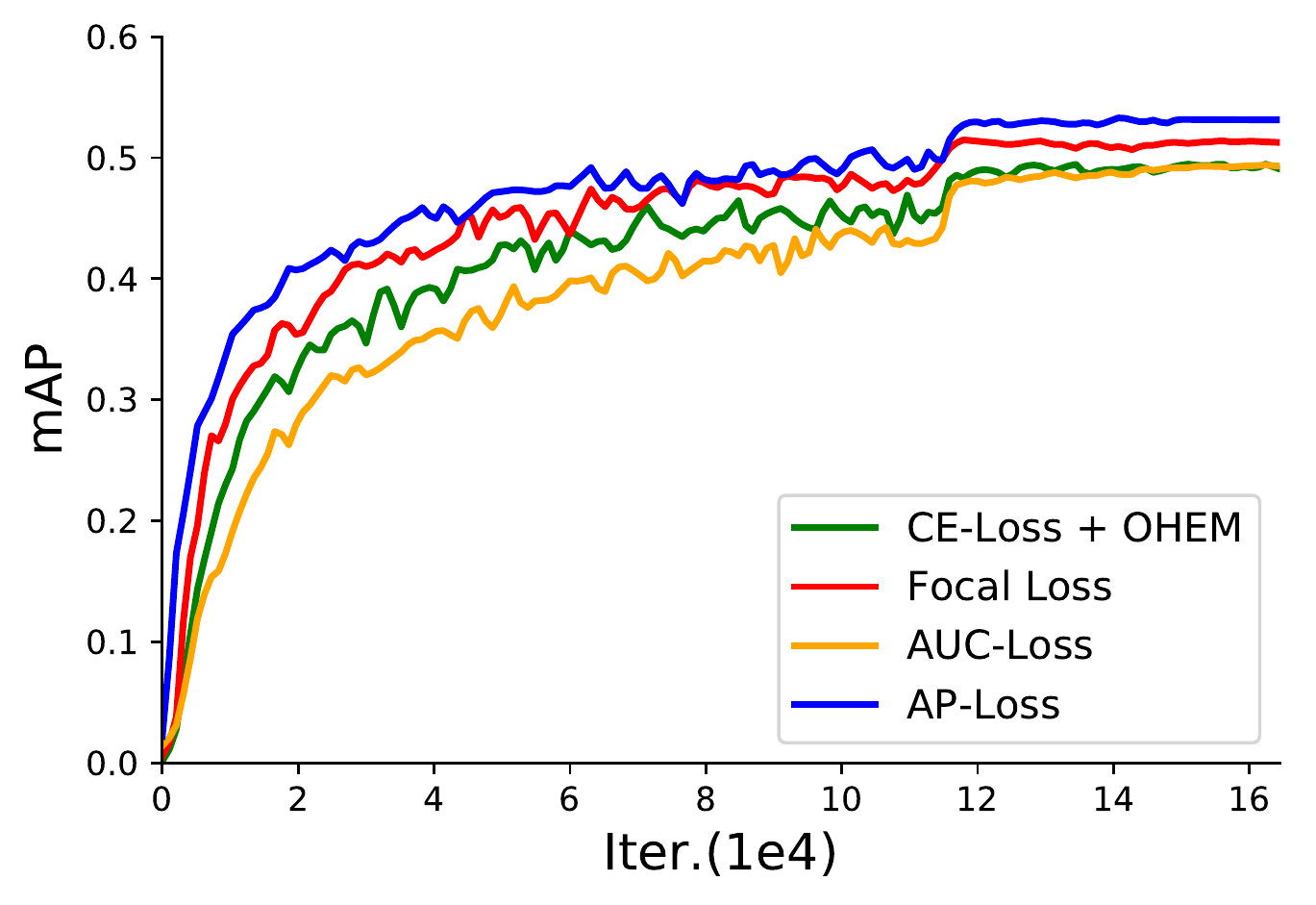}
  }
\subfloat[\label{fig-ssd_performance_curve}]{
  \includegraphics[width=0.49\linewidth]{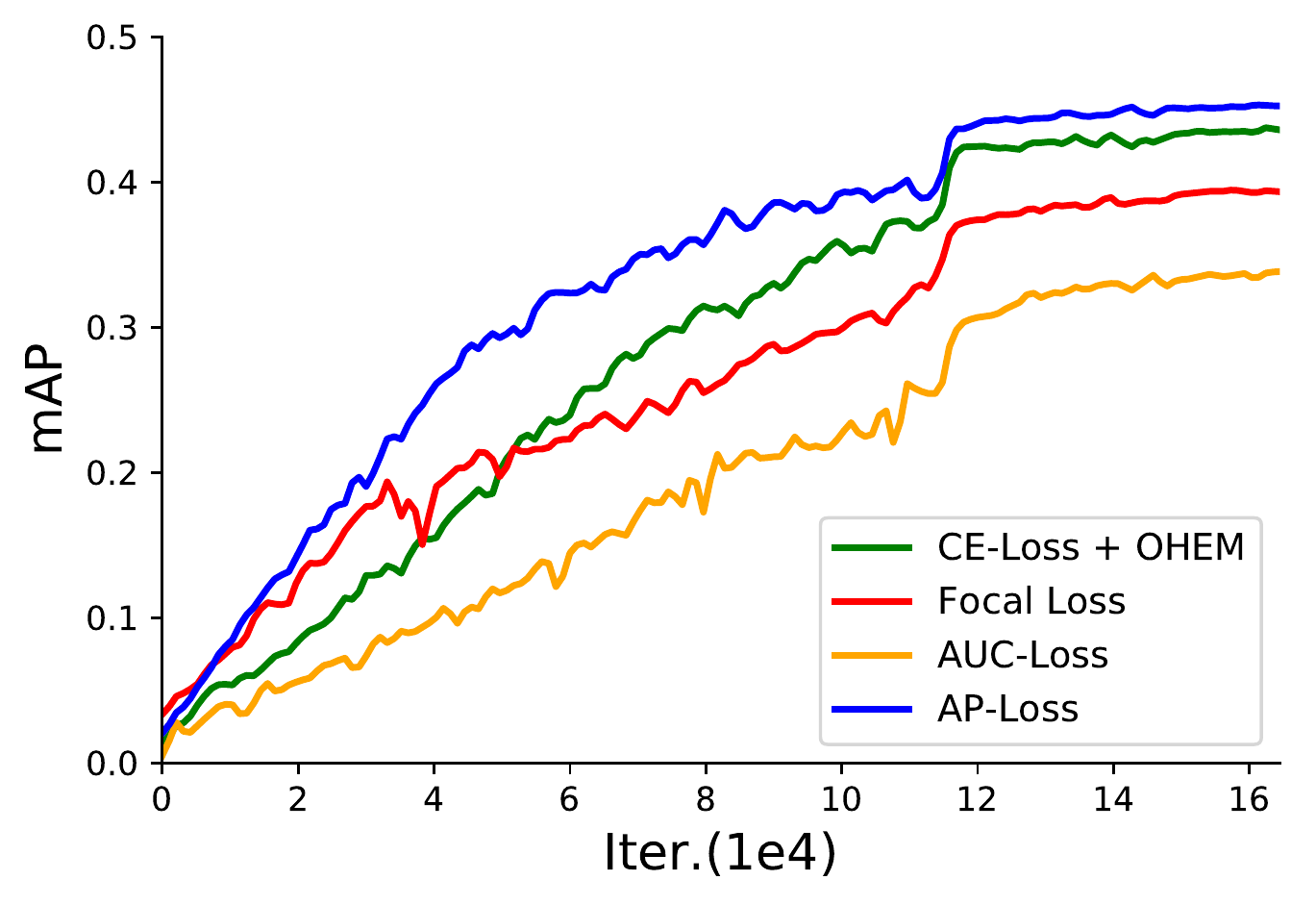}
  }
\vspace{-1mm}
\caption{(a) Detection accuracy (mAP) of RetinaNet on VOC2007 {\tt test} set. (b) Detection accuracy (mAP) of SSD on VOC2007 {\tt test} set. (Best viewed in color)}
\end{figure}

\begin{table*}[t]
\small
\centering
\caption{Comparison through different training losses on robustness. Models are tested on VOC2007 {\tt test} sets. The metric mAP\(_{50}\) is used except for DeepFool~\cite{moosavi2016deepfool}. DeepFool uses the minimal perturbations (for successful attack) to measure the robustness.}
\setlength{\tabcolsep}{1.5mm}{
\begin{tabular}{cccccccc}
\midrule[1pt]
Training Loss & Original & Black Patch & Random Patch & Flip Patch & Advs-Patch~\cite{brown2017adversarial} & Gaussian Noise & DeepFool~\cite{moosavi2016deepfool} \\
\midrule[1pt]
CE-Loss+OHEM & 81.5 & 63.1 & 62.8 & 77.0 & 67.8 & 65.5 & 14.7 \\
Focal Loss & 80.9 & 60.9 & 61.3 & 76.9 & 67.1 & 65.7 & 14.9\\
AUC-Loss & 79.7 & 61.9 & 61.6 & 75.9 & 66.8 & 63.7 & 15.6 \\
AP-Loss & \textbf{82.3} & \textbf{65.2} & \textbf{65.2} & \textbf{78.5} & \textbf{68.9} & \textbf{68.8} & \textbf{16.4} \\
\midrule[1pt]
\end{tabular}}
\label{base-loss-robustness}
\end{table*}

\subsubsection{Comparison on Different Losses}
\begin{figure}[t]
\small
\centering
\subfloat[CE-loss+OHEM]{
  \includegraphics[width=0.5\linewidth]{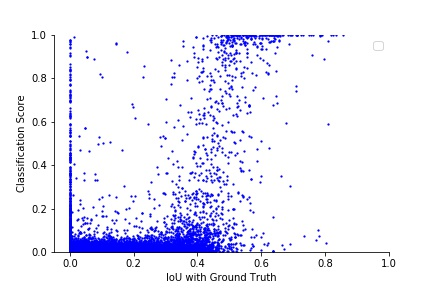}
  }
\subfloat[Focal-loss]{
  \includegraphics[width=0.5\linewidth]{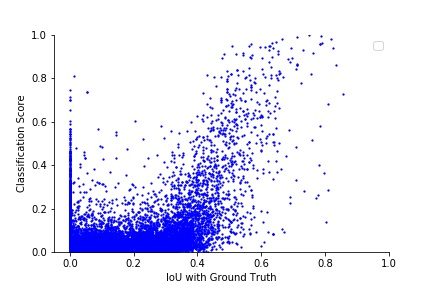}
  }
\par
\subfloat[AUC-loss]{
  \includegraphics[width=0.5\linewidth]{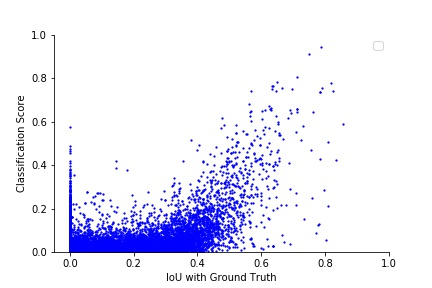}
  }
\subfloat[AP-loss]{
  \includegraphics[width=0.5\linewidth]{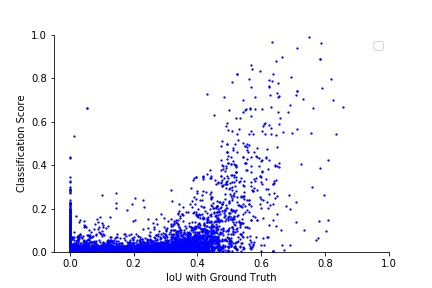}
  }
\vspace{-1mm}
  \caption{The correlation between the IoU of anchor with matched ground-truth and the classification score.}
\vspace{-2mm}
\label{dist}
\end{figure}

We compare traditional classification based losses like Focal-loss~\cite{lin2018focal} and cross entropy loss (CE-loss) with OHEM~\cite{liu2016ssd} to ranking based losses like AUC-loss and AP-loss. We do not use ordinary CE-loss in our experiments since it performs much worse without the anti-imbalance technique. The experiments are conducted on various detectors (\textit{i.e.} RetinaNet and SSD) and datasets (\textit{i.e.} PASCAL VOC and COCO). The results are shown in \autoref{base-loss}. To outline the superiority of the AP-loss for snapshot time points, we also evaluate the detection performances at different training iterations. The results are shown in \autoref{fig-performance_curve} and \autoref{fig-ssd_performance_curve}. Besides, the score distribution (correlation of IoU vs. classification score) also reflects some interesting properties of these losses. These results are shown in \autoref{dist}. We collect the output scores of the models trained with different losses on VOC2007 test set. The x-axis is the IoU between the anchor box and its matched ground-truth box, while the y-axis denotes its classification score based on its ground-truth class. The classification score of CE-loss+OHEM is computed by softmax function, while the other three losses use sigmoid function to compute their classification scores. We observe that the AP-loss has the least false positive rate. The CE-loss+OHEM assigns very high confidence to the true positive samples, but does not perform well when handling the negative samples, especially the negative samples with IoU=0 (notice the thin vertical strip of points). \autoref{fig-result} illustrates some sample detection results by the RetinaNet with Focal-loss and our AP-loss.

{\bf \noindent Performances Across Datasets:}
We first study the performances of RetinaNet across different datasets. Although Focal-loss is significantly better than CE-loss with OHEM on COCO dataset, it is interesting that Focal-loss does not perform better than OHEM at AP\(_{50}\) on PASCAL VOC. This is likely because the hyper-parameters of Focal-loss are designed to suit the imbalance condition on COCO dataset which is not suitable for PASCAL VOC, so that Focal-loss cannot generalize well to PASCAL VOC without tuning its hyper-parameters. However, the proposed AP-loss performs much better than all the other losses on both two datasets (mAP 53.1\% \textit{vs.} 51.3\% on PASCAL VOC, 35.0\% \textit{vs.} 33.9\% on COCO), which demonstrates its effectiveness and stronger generalization ability on handling the imbalance issue. It is worth noting that AUC-loss performs much worse than AP-loss, which may be due to the fact that AUC~\cite{cortes2004auc} has equal penalty for each misordered pair while AP imposes greater penalty for the misordering at higher positions in the predicted ranking. It is obvious that object detection evaluation concerns more on objects with higher confidence, which is why AP provides a better loss measure than AUC.

{\bf \noindent Performances Across Detectors:}
Then we study the performances of different detectors on PASCAL VOC dataset. Note that the Focal-loss performs much worse than OHEM on SSD, though we have carefully tuned its hyper-parameters \(\gamma\) and \(\alpha\). The reason is similar: RetinaNet has much denser anchors than SSD, so the imbalance conditions is very much different from what the Focal-loss was designed for. However, the proposed AP-loss still performs better than the other losses on SSD (mAP 45.2\% \textit{vs.} 43.6\%), which demonstrate the robustness of the proposed loss on different detectors. Together with the results on RetinaNet, we can observe the effectiveness and strong generalization ability of the proposed approach.

{\bf \noindent Robustness:}
Compared to existing classification based loss functions, our AP-loss explicitly models the relationship between samples (\textit{i.e.} anchors). Thus the detection model trained by AP-loss will learn to capture more \textbf{contextual} and \textbf{global} information of the images in the training phase. This makes the model more robust under some local perturbations and noises, or even adversarial attacks. To verify this point, we evaluate the detectors trained with different losses on five different perturbations of PASCAL VOC2007 {\tt test} datasets. Similar to the setting in \cite{li2018auto}, we add four types of patches to the center of each object instance, to destroy the local information inside the object region. The patch is of half width or half height of the ground-truth bounding box annotation. The types of patch include 1) \textit{Black-Patch} with all pixels being zero, 2) \textit{Flip-Patch} that is flipped version of the center patch itself using left/right, top/bottom, or both randomly, 3) \textit{Random-Patch} that randomly sampled from outside of a bounding box, 4) \textit{Adversarial-Patch} generated by \cite{brown2017adversarial}. Besides, the Gaussian noise with zero mean and 0.01 variance is added on each pixel of normalized image to generate the \textit{Gaussian-Noise} dataset. Experimental results are shown in \autoref{base-loss-robustness}. It can be seen that, the proposed AP-loss outperforms all other methods not only on normal images, but also on all these perturbation images. We also use another adversarial attack method DeepFool~\cite{moosavi2016deepfool} to evaluate the models. Different from other perturbation methods in \autoref{base-loss-robustness}, DeepFool can evaluate the robustness by measuring the minimal perturbation that is sufficient to change the predictions of model. Thus, a larger minimal perturbation indicates stronger robustness towards adversarial examples. We slightly generalize this method from attacking classifier to attacking detector. We observe that AP-loss have better results than other methods, which demonstrates its robustness against adversarial attacks.

\subsubsection{Comparison on Different Optimization Methods}

We also compare our optimization method with the approximate gradient method~\cite{song2016training,henderson2016end} and structured hinge loss method~\cite{Mohapatra_2018_CVPR}. Both methods~\cite{song2016training,henderson2016end} approximate the AP-loss with a smooth expectation and envelope function, respectively. Following their guidance, we replace the step function in AP-loss with a sigmoid function to constrain the gradient to neither zero nor undefined, while still keeping the shape similar to the original function. Similar to~\cite{henderson2016end}, we adopt the log space objective function, \textit{i.e.} \(log(\text{AP}+\epsilon)\), to allow the model to quickly escape from the initial state.

\begin{figure}[t]
\small
\centering
\subfloat[\label{fig-convergence_curve}]{
  \includegraphics[width=0.49\linewidth]{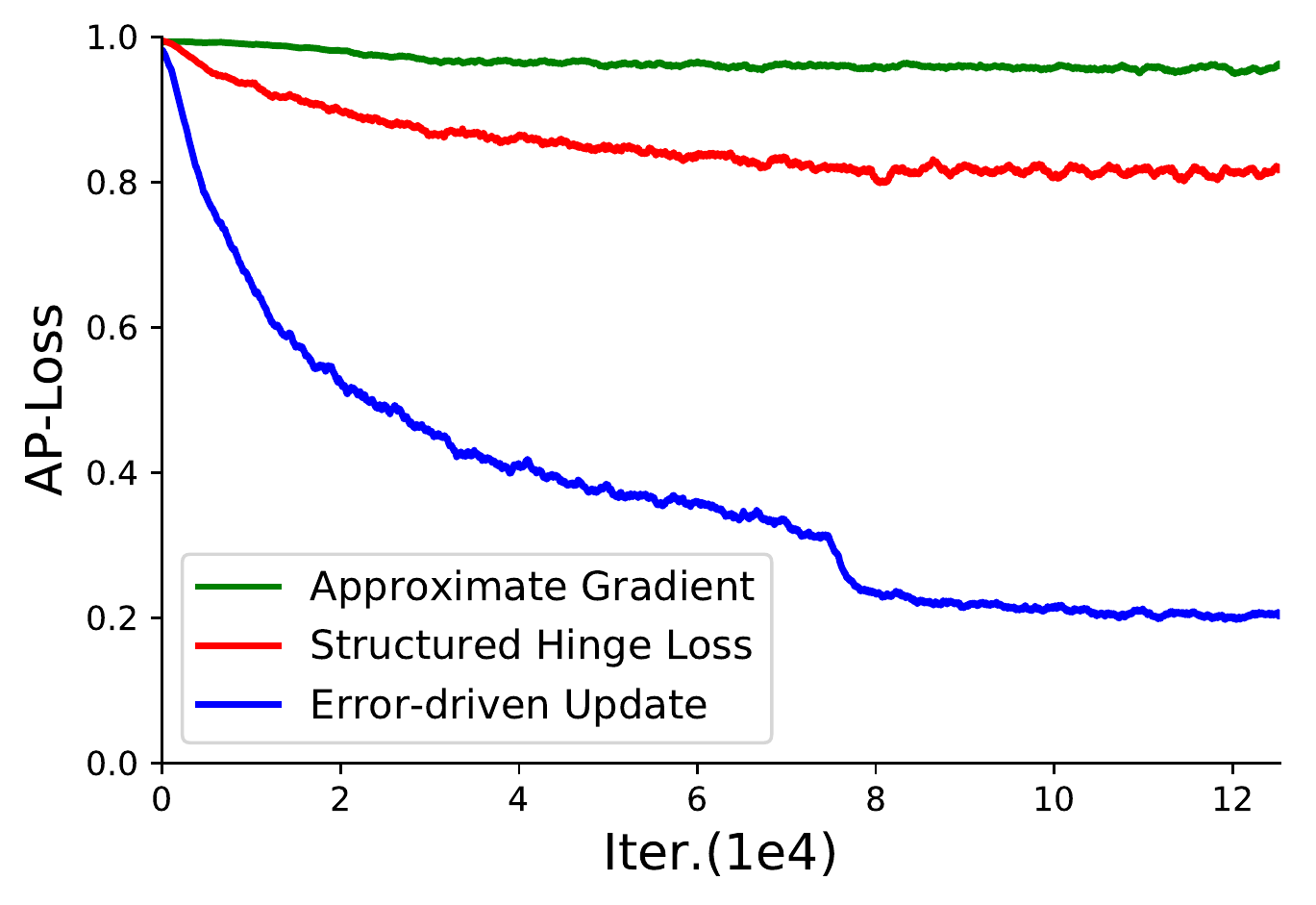}
  }
\subfloat[\label{fig-performance_imbalance}]{
  \includegraphics[width=0.49\linewidth]{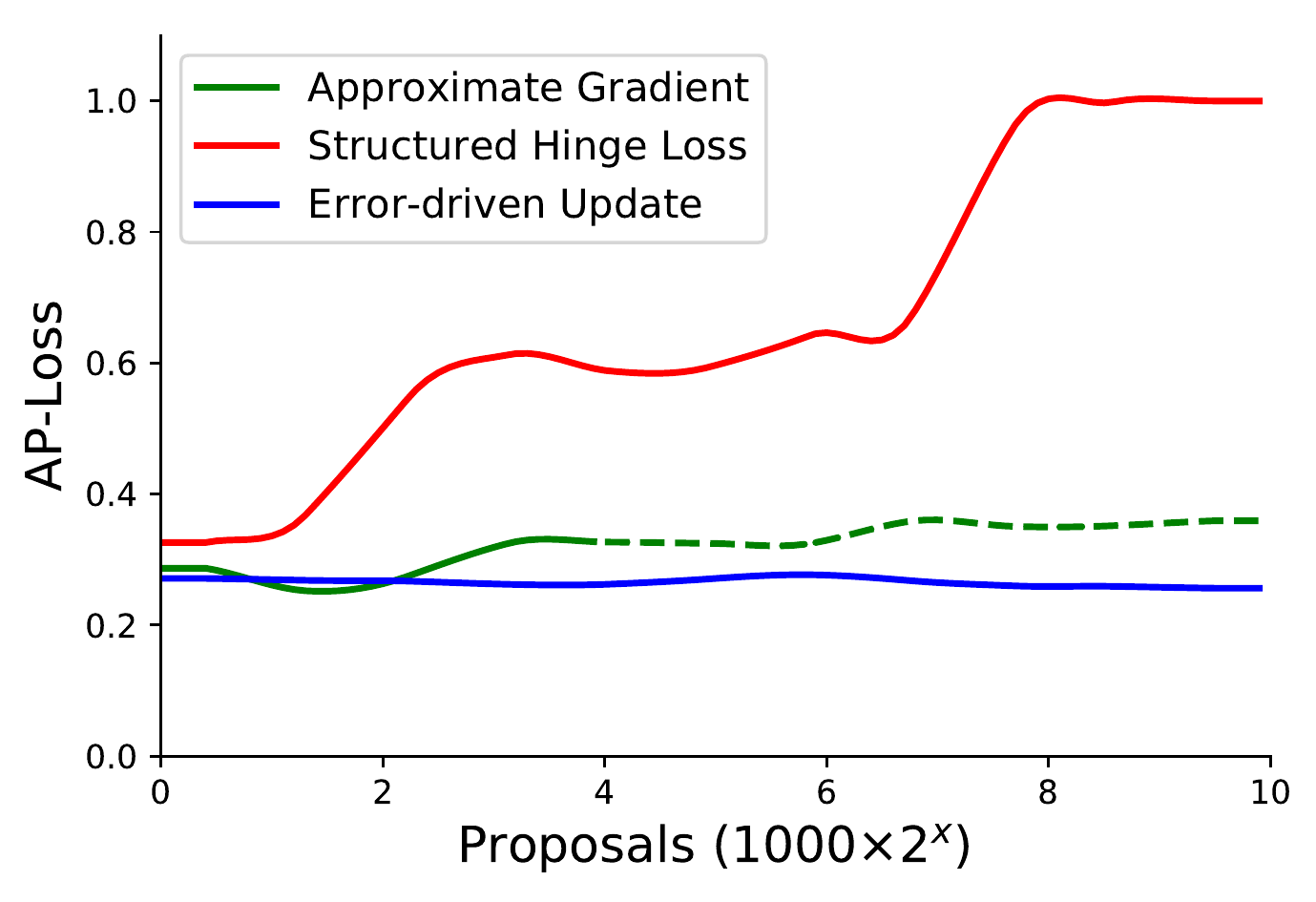}
  }
\vspace{-1mm}
  \caption{(a) Convergence curves of different AP-loss optimizations on VOC2007 {\tt trainval} set. (b) Final convergence values of AP-loss in different imbalance conditions. Dashed line means the loss cannot converge from the ImageNet pre-trained model. (Best viewed in color)}
\vspace{-1mm}
\label{fig-convergence-all}
\end{figure}

{\bf \noindent Convergence Performance:}
We train the detector on VOC2007 {\tt trainval} set and turn off the bounding box regression task. The convergence curves shown in \autoref{fig-convergence_curve} reveal some essential observations. It can be seen that AP-loss optimized by approximate gradient method does not even converge, likely because its non-convexity and non-quasiconvexity fail on a direct gradient descent method. Meanwhile, AP-loss optimized by the structured hinge loss method~\cite{Mohapatra_2018_CVPR} converges slowly and stabilizes near 0.8, which is significantly worse than the asymptotic limit of AP-loss optimized by our error-driven update scheme. The reason is that this method does not optimize the AP-loss directly but rather an upper bound of it. As mentioned before, the structured hinge loss introduces a gap, which is controlled by a discriminant function~\cite{Mohapatra_2018_CVPR}. In ranking task, the discriminant function is hand-picked and has an AUC-like form, which may cause variability in optimization.

{\bf \noindent Study on Imbalance Condition:}
Note that our experimental results in \autoref{fig-convergence_curve} are different from what was reported in \cite{song2016training,Mohapatra_2018_CVPR}, where they observed several improvements compared to the baseline model. In \cite{song2016training,Mohapatra_2018_CVPR}, experiments were conducted on the two stage detector R-CNN with about 2000 pre-computed proposals generated using selective search algorithm~\cite{uijlings2013selective}. However, in our experimental setting, the one-stage detectors have about \(10^5 \sim 10^6\) anchors in one image. Thus, the imbalance issue in \cite{song2016training,Mohapatra_2018_CVPR} are relatively moderate in comparison, which makes it easier for AP-loss optimization. To verify this point, we evaluate these optimization methods at different imbalance conditions. Specifically, we simulate different imbalance cases by using a pre-trained RPN~\cite{ren2015faster} to produce different number of proposals for each training image. Unlike typical two-stage detection, we turn off the regression branch in RPN, so that our proposals are exactly the anchors with high probability of containing an object. The union set over those proposal anchors and the true positive anchors are then treated as the valid training samples for the one-stage detector. We vary the number of proposals generated by RPN in each image to simulate different imbalance conditions. Results are shown in \autoref{fig-performance_imbalance}. The dashed line means the loss cannot converge from the ImageNet pre-trained model, but can converge from some model snapshots which are trained with less proposals. Specifically, we use the snapshot model trained with 2000 proposals for 10 epochs. It can be seen that, as the number of proposals increases, \textit{i.e.} the imbalance rate grows, the final AP-loss after optimized by structured hinge loss converge appears to be increasingly higher than that of the proposed error-driven method. The approximate gradient method keeps a good performance when the number of proposals is less than 4000. However, when the number of proposals is greater than 8000, the loss can only converge from the model snapshot trained with less proposals. This is because the AP-loss is non-convex, a good initial state is important to prevent from falling into local minimum.
Note that the proposed method keeps a steady final convergence value consistently over different imbalance conditions, which demonstrates its consistency and independence against increasing number of training samples.

\subsection{Memory and Time Cost}

As mentioned before, the proposed AP-loss training algorithm somewhat suffers from the high complexity due to the need of pairwise differences computation. However, we notice that both memory and time cost of AP-loss training algorithm can be significantly reduced with the proposed acceleration techniques introduced in Section \ref{complexity_and_acceleration}. To verify its effectiveness, the memory and time cost of AP-loss training algorithm are studied through several experiments.

\begin{figure}[t]
\centering
\small
\includegraphics[width=0.95\linewidth]{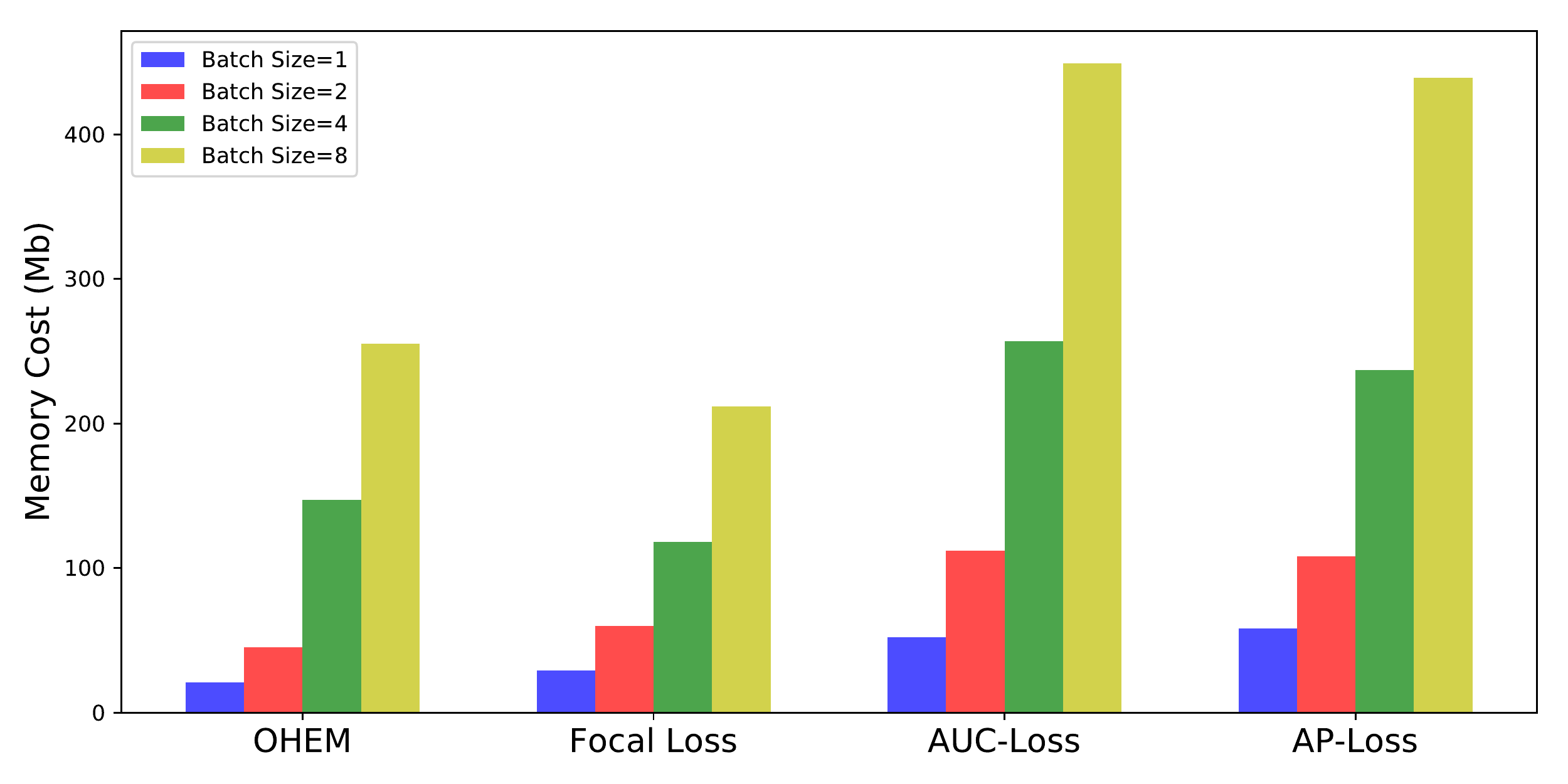}
\caption{Memory cost of different training losses. All models are based on RetinaNet-ResNet50 and evaluated on PASCAL VOC dataset.}
\label{fig_memory_cost}
\end{figure}

\subsubsection{Memory Cost}

Here, we study the memory cost of the proposed method. The experiments are conducted on PASCAL VOC dataset with the detection model fixed to RetinaNet-ResNet50. Without these acceleration techniques, the difference transformation cannot fit into the GPU memory even with small batch sizes. Therefore, we only evaluate our algorithm with these acceleration techniques. Results are shown in \autoref{fig_memory_cost}. It is worth noting that the memory costs are only estimations. We first evaluate the detection model without a loss (with backward propagation still enable). Then, the detection model combined with different losses are evaluated. Therefore, the memory cost for the loss computation can be simply inferred by measuring the differences. Though some increases are observed in the memory cost of our method than that of Focal-Loss~\cite{lin2018focal} and OHEM~\cite{liu2016ssd}, it is still negligible compared to the total memory cost of neural network. We can also find that the memory cost scales linearly with the batch size, which verifies our estimated \(O(|\mathcal{N}|)\) space complexity in Section \ref{complexity_and_acceleration}.

\begin{figure}[t]
\centering
\small
\includegraphics[width=0.95\linewidth]{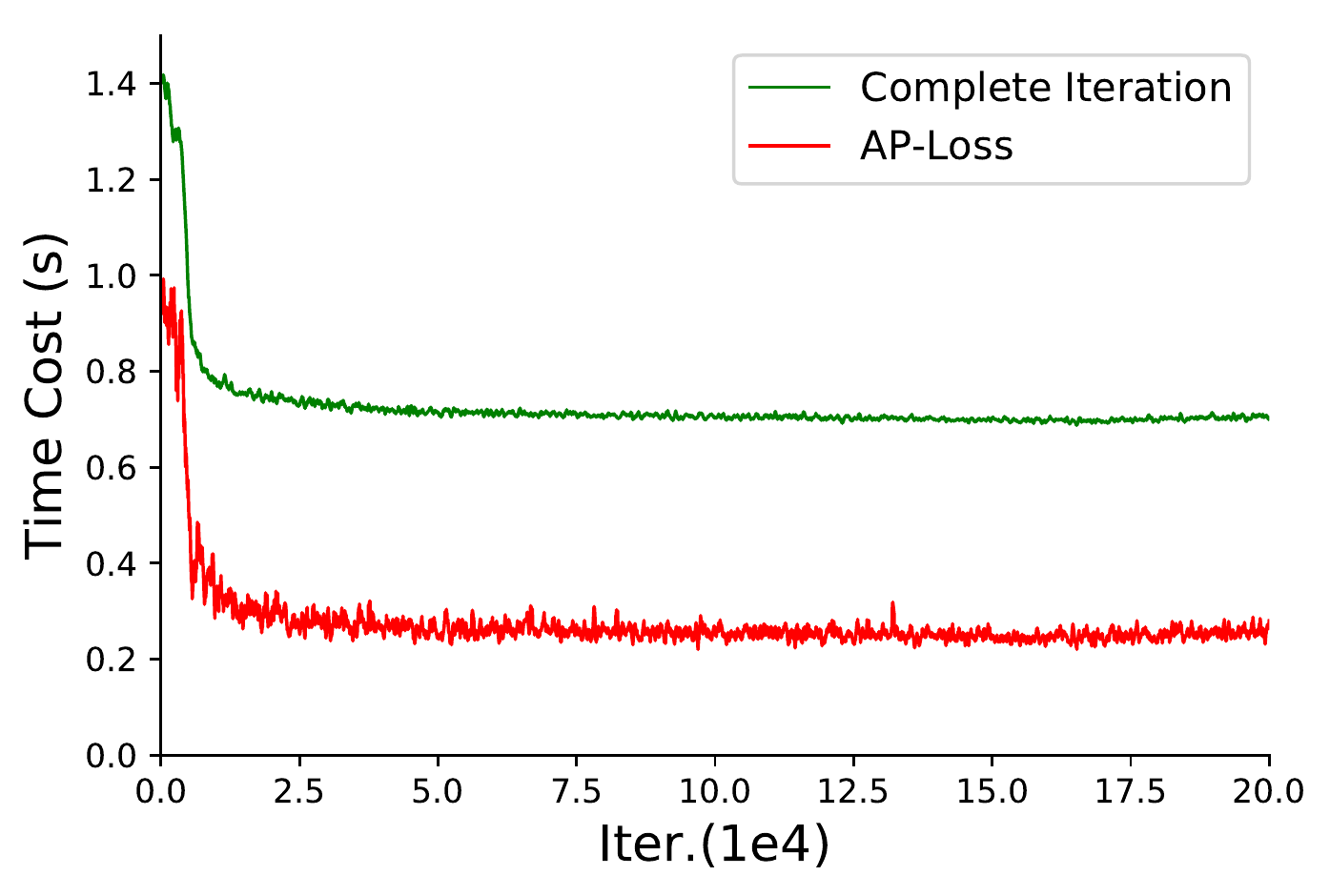}
\caption{Time cost of AP-loss and complete iteration. Batch size equals 8. Only 1 GPU is used. Model are based on RetinaNet-ResNet50 and evaluated on PASCAL VOC dataset.}
\label{fig_time_cost_curve}
\end{figure}

\subsubsection{Time Cost}

In this section, the time costs of training algorithm are evaluated. Experiments are conducted on PASCAL VOC dataset with the detection model fixed to RetinaNet-ResNet50. For simplicity, we use only a single GPU for training here. As studied in \autoref{voc-abaltion-study}, the hyper-parameter \(\delta\) is selected based on the detection performance. As mentioned before, we argue that the training algorithm is going faster along with the increasing performance of detection model. To verify this statement, we trace and report the time cost of both AP-loss and complete training iteration during the whole training phase. Results are shown in \autoref{fig_time_cost_curve}. It shows that the time cost of AP-loss is rapidly reduced to 0.3(s) only after \(1.5\times 10^4\) iterations from the beginning and then stabilized as the minimum. Safe to say, the averaged time cost over the entire training phase is basically close to the minimal time cost at the final convergence stage. We also note that the overall training cost appears to be strongly linked to the loss time cost; any improvements to the loss cost is crucial to the overall time cost.

\begin{figure}[t]
\centering
\small
\includegraphics[width=0.95\linewidth]{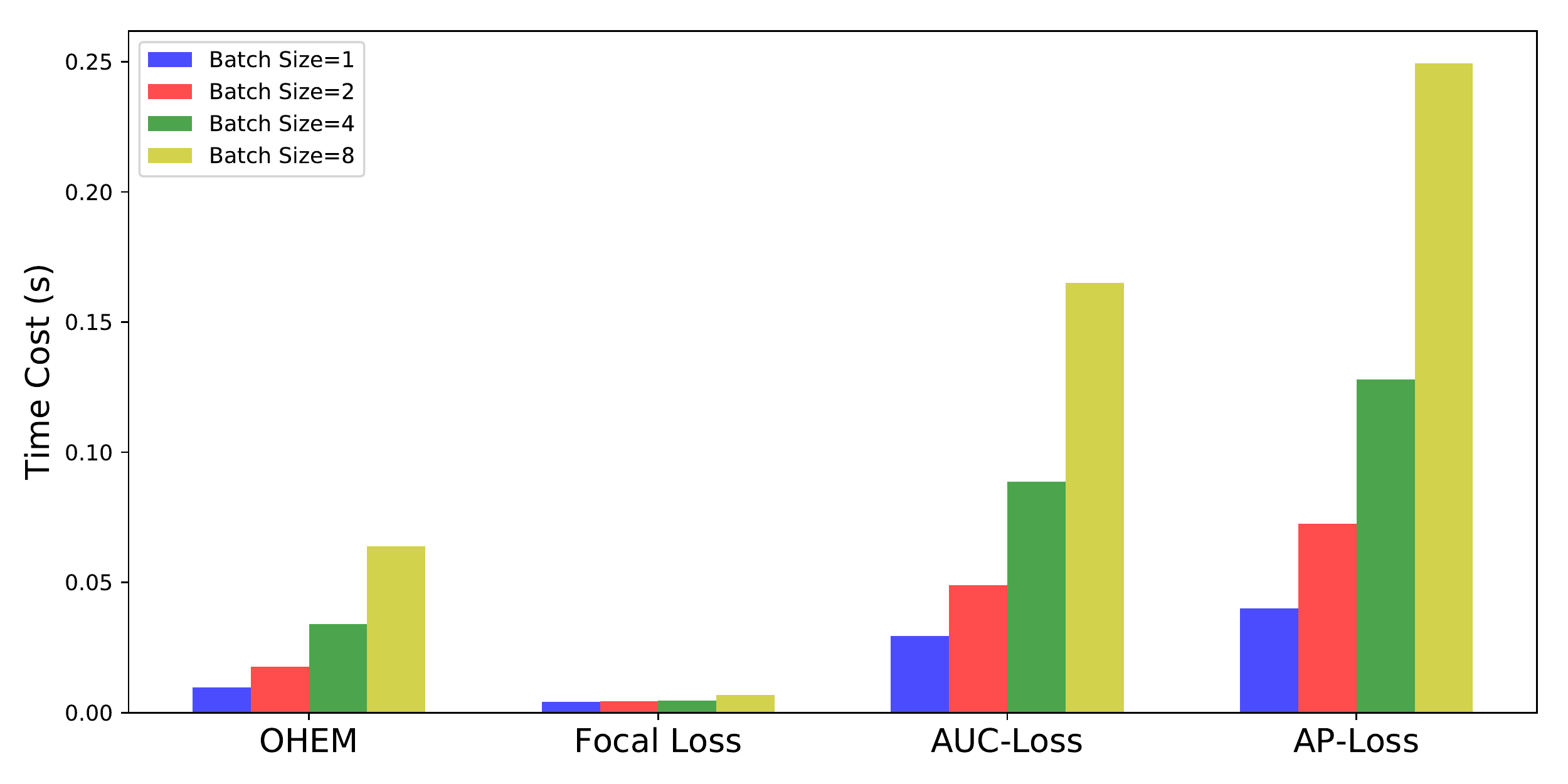}
\caption{Time cost of different training losses. All models are based on RetinaNet-ResNet50 and evaluated on PASCAL VOC dataset.}
\label{fig_time_cost}
\end{figure}

We also evaluate the complexities with different losses and bath sizes. The time costs of AP and other losses are recorded and averaged over the last epoch in the training phase. Results are shown in \autoref{fig_time_cost}. We observe that Focal-Loss has the lowest time cost and the time cost almost stays the same when the batch size increases. The reason is that Focal-Loss treats each sample independently so it likely benefited the most from the parallel computation in GPU. However, the other losses such as OHEM, AUC-loss and AP-loss all require sequential computing, which is the dominant factor of their lower speeds. We emphasize here that, though the AP-loss has higher cost than both OHEM and Focal-Loss, it is still an acceptable option due to the following justifications: \textbf{1)} relatively small impact (about \(1/3\) of cost) to the complete training iteration, \textbf{2)} faster convergence rate (See \autoref{fig-convergence-all}), \textbf{3)} better performance with equal efficiency at test-time.
In addition, we observe good scalability of our algorithm. As mentioned in Section \ref{complexity_and_acceleration}, the time complexity of AP-loss is \(O(|\mathcal{P}|\cdot|\mathcal{\widehat{N}}|+|\mathcal{N}|)\), which indicates that the time cost will increase quadratically with the increase in batch size. However, we only observe almost linear increase in our experiments. The reason for this lies in the computation for \(\{L,x\}_{ij}\) in each iteration which is implemented through parallel computing in GPU. In practice, this translates to a real complexity of about \(O(|\mathcal{P}|+|\mathcal{N}|)\).

\begin{table*}[t]
\footnotesize
\centering
\caption{Detection results on VOC2007 {\tt test} set.}
\vspace{-2mm}
\setlength{\tabcolsep}{1mm}{
\resizebox{1\linewidth}{!}{
\begin{tabular}{ccccccccccccccccccccccc}
\midrule[1pt]
Method & Backbone & mAP\(_{50}\) & aero & bike & bird & boat & bottle & bus & car & cat & chair & cow & table & dog & horse & mbike & person & plant & sheep & sofa & train & tv \\
\midrule[1pt]
Two-Stage: & & & & & & & & & & & & & & & & & & & & & & \\
Faster~\cite{ren2015faster} & ResNet-101 & 76.4 & 79.8 & 80.7 & 76.2 & 68.3 & 55.9 & 85.1 & 85.3 & \textbf{89.8} & 56.7 & 87.8 & 69.4 & 88.3 & 88.9 & 80.9 & 78.4 & 41.7 & 78.6 & 79.8 & 85.3 & 72.0 \\
ION~\cite{bell2016inside} & VGG-16 & 76.5 & 79.2 & 79.2 & 77.4 & 69.8 & 55.7 & 85.2 & 84.2 & \textbf{89.8} & 57.5 & 78.5 & 73.8 & 87.8 & 85.9 & 81.3 & 75.3 & 49.7 & 76.9 & 74.6 & 85.2 & 82.1 \\
MR-CNN~\cite{gidaris2015object} & VGG-16 & 78.2 & 80.3 & 84.1 & 78.5 & 70.8 & 68.5 & 88.0 & 85.9 & 87.8 & 60.3 & 85.2 & 73.7 & 87.2 & 86.5 & 85.0 & 76.4 & 48.5 & 76.3 & 75.5 & 85.0 & 81.0 \\
R-FCN~\cite{dai2016r} & ResNet-101 & 80.5 & 79.9 & 87.2 & 81.5 & 72.0 & 69.8 & 86.8 & 88.5 & \textbf{89.8} & 67.0 & 88.1 & 74.5 & 89.8 & \textbf{90.6} & 79.9 & 81.2 & 53.7 & 81.8 & 81.5 & 85.9 & 79.9 \\
CoupleNet~\cite{zhu2017couplenet} & ResNet-101 & 82.7 & 85.7 & 87.0 & \textbf{84.8} & 75.5 & 73.3 & 88.8 & 89.2 & 89.6 & 69.8 & 87.5 & 76.1 & 88.9 & 89.0 & 87.2 & 86.2 & \textbf{59.1} & 83.6 & 83.4 & \textbf{87.6} & 80.7 \\
Revisiting-RCNN~\cite{cheng2018revisiting} & ResNet-101+152 & \textbf{84.0} & \textbf{89.3} & \textbf{88.7} & 80.5 & \textbf{77.7} & \textbf{76.3} & \textbf{90.1} & \textbf{89.6} & \textbf{89.8} & \textbf{72.9} & \textbf{89.2} & \textbf{77.8} & \textbf{90.1} & 90.0 & \textbf{87.5} & \textbf{87.2} & 58.6 & \textbf{88.2} & \textbf{84.3} & 87.5 & \textbf{85.0} \\
\midrule[1pt]
\midrule[1pt]
One-Stage: & & & & & & & & & & & & & & & & & & & & & & \\
YOLOv2~\cite{redmon2017yolo9000} & DarkNet-19 & 78.6 & - & - & - & - & - & - & - & - & - & - & - & - & - & - & - & - & - & - & - & - \\
DSOD300~\cite{shen2017dsod} & DS/64-192-48-1 & 77.7 & - & - & - & - & - & - & - & - & - & - & - & - & - & - & - & - & - & - & - & -  \\
SSD512*~\cite{liu2016ssd} & VGG-16 & 79.8 & 84.8 & 85.1 & 81.5 & 73.0 & 57.8 & 87.8 & 88.3 & 87.4 & 63.5 & 85.4 & 73.2 & 86.2 & 86.7 & 83.9 & 82.5 & 55.6 & 81.7 & 79.0 & 86.6 & 80.0 \\
SSD513~\cite{fu2017dssd} & ResNet-101 & 80.6 & 84.3 & 87.6 & 82.6 & 71.6 & 59.0 & 88.2 & 88.1 & 89.3 & 64.4 & 85.6 & 76.2 & 88.5 & 88.9 & 87.5 & 83.0 & 53.6 & 83.9 & \textbf{82.2} & 87.2 & 81.3 \\
DSSD513~\cite{fu2017dssd} & ResNet-101 & 81.5 & 86.6 & 86.2 & 82.6 & 74.9 & 62.5 & \textbf{89.0} & 88.7 & 88.8 & 65.2 & 87.0 & \textbf{78.7} & 88.2 & 89.0 & 87.5 & 83.7 & 51.1 & 86.3 & 81.6 & 85.7 & \textbf{83.7} \\
DES512~\cite{Zhang_2018_CVPR} & VGG-16 & 81.7 & 87.7 & 86.7 & 85.2 & 76.3 & 60.6 & 88.7 & 89.0 & 88.0 & 67.0 & 86.9 & 78.0 & 87.2 & 87.9 & 87.4 & 84.4 & 59.2 & 86.1 & 79.2 & \textbf{88.1} & 80.5 \\
RFBNet512~\cite{Liu_2018_ECCV} & VGG-16 & 82.2 & - & - & - & - & - & - & - & - & - & - & - & - & - & - & - & - & - & - & - & - \\
PFPNet-R512~\cite{kim2018parallel} & VGG-16 & 82.3 & - & - & - & - & - & - & - & - & - & - & - & - & - & - & - & - & - & - & - & - \\
RefineDet512~\cite{zhang2018single} & VGG-16 & 81.8 & \textbf{88.7} & 87.0 & 83.2 & 76.5 & 68.0 & 88.5 & 88.7 & 89.2 & 66.5 & 87.9 & 75.0 & 86.8 & 89.2 & 87.8 & 84.7 & 56.2 & 83.2 & 78.7 & \textbf{88.1} & 82.3 \\
AP-loss500 & ResNet-101 & \textbf{83.9} & 87.2 & \textbf{88.3} & \textbf{85.9} & \textbf{80.5} & \textbf{73.6} & 87.9 & \textbf{89.5} & \textbf{89.8} & \textbf{71.6} & \textbf{88.8} & 77.4 & \textbf{88.8} & \textbf{89.8} & \textbf{89.3} & \textbf{87.0} & \textbf{63.3} & \textbf{86.6} & 81.5 & 87.8 & 83.1 \\
\midrule[1pt]
PFPNet-R512+~\cite{kim2018parallel} & VGG-16 & 84.1 & - & - & - & - & - & - & - & - & - & - & - & - & - & - & - & - & - & - & - & - \\
RefineDet512+~\cite{zhang2018single} & VGG-16 & 83.8 & 88.5 & 89.1 & 85.5 & 79.8 & 72.4 & \textbf{89.5} & \textbf{89.5} & \textbf{89.9} & 69.9 & \textbf{88.9} & 75.9 & 87.4 & 89.6 & \textbf{89.0} & 86.2 & 63.9 & 86.2 & 81.0 & \textbf{88.6} & 84.4 \\
AP-loss500+ & ResNet-101 & \textbf{84.9} & \textbf{88.9} & \textbf{89.6} & \textbf{87.8} & \textbf{81.7} & \textbf{76.2} & 89.0 & \textbf{89.5} & 89.8 & \textbf{74.8} & 87.9 & \textbf{79.3} & \textbf{88.6} & \textbf{89.8} & 88.7 & \textbf{87.7} & \textbf{66.5} & \textbf{86.7} & \textbf{84.2} & 88.4 & \textbf{85.4} \\
\midrule[1pt]
\end{tabular}}}
\vspace{-0mm}
\begin{tablenotes}
\item + denotes multi-scale testing.
\end{tablenotes}
\label{state-of-the-art-voc07}
\end{table*}

\begin{table*}[t]
\footnotesize
\centering
\caption{Detection results on VOC2012 {\tt test} set.}
\vspace{-2mm}
\setlength{\tabcolsep}{1mm}{
\resizebox{1\linewidth}{!}{
\begin{tabular}{ccccccccccccccccccccccc}
\midrule[1pt]
Method & Backbone & mAP\(_{50}\) & aero & bike & bird & boat & bottle & bus & car & cat & chair & cow & table & dog & horse & mbike & person & plant & sheep & sofa & train & tv \\
\midrule[1pt]
Two-Stage: & & & & & & & & & & & & & & & & & & & & & & \\
Faster~\cite{ren2015faster} & ResNet-101 & 73.8 & 86.5 & 81.6 & 77.2 & 58.0 & 51.0 & 78.6 & 76.6 & 93.2 & 48.6 & 80.4 & 59.0 & 92.1 & 85.3 & 84.8 & 80.7 & 48.1 & 77.3 & 66.5 & 84.7 & 65.6 \\
ION~\cite{bell2016inside} & VGG-16 & 76.4 & 87.5 & 84.7 & 76.8 & 63.8 & 58.3 & 82.6 & 79.0 & 90.9 & 57.8 & 82.0 & 64.7 & 88.9 & 86.5 & 84.7 & 82.3 & 51.4 & 78.2 & 69.2 & 85.2 & 73.5 \\
MR-CNN~\cite{gidaris2015object} & VGG-16 & 73.9 & 85.5 & 82.9 & 76.6 & 57.8 & 62.7 & 79.4 & 77.2 & 86.6 & 55.0 & 79.1 & 62.2 & 87.0 & 83.4 & 84.7 & 78.9 & 45.3 & 73.4 & 65.8 & 80.3 & 74.0 \\
R-FCN~\cite{dai2016r} & ResNet-101 & 77.6 & 86.9 & 83.4 & 81.5 & 63.8 & 62.4 & 81.6 & 81.1 & 93.1 & 58.0 & 83.8 & 60.8 & 92.7 & 86.0 & 84.6 & 84.4 & 59.0 & 80.8 & 68.6 & 86.1 & 72.9 \\
CoupleNet~\cite{zhu2017couplenet} & ResNet-101 & 80.4 & 89.1 & \textbf{86.7} & 81.6 & 71.0 & 64.4 & \textbf{83.7} & 83.7 & 94.0 & 62.2 & 84.6 & 65.6 & 92.7 & \textbf{89.1} & 87.3 & \textbf{87.7} & \textbf{64.3} & 84.1 & \textbf{72.5} & \textbf{88.4} & 75.3 \\
Revisiting-RCNN~\cite{cheng2018revisiting} & ResNet-101+152 & \textbf{81.2} & \textbf{89.6} & \textbf{86.7} & \textbf{83.8} & \textbf{72.8} & \textbf{68.4} & \textbf{83.7} & \textbf{85.0} & \textbf{94.5} & \textbf{64.1} & \textbf{86.6} & \textbf{66.1} & \textbf{94.3} & 88.5 & \textbf{88.5} & 87.2 & 63.7 & \textbf{85.6} & 71.4 & 88.1 & \textbf{76.1} \\
\midrule[1pt]
\midrule[1pt]
One-Stage: & & & & & & & & & & & & & & & & & & & & & & \\
YOLOv2~\cite{redmon2017yolo9000} & DarkNet-19 & 73.4 & 86.3 & 82.0 & 74.8 & 59.2 & 51.8 & 79.8 & 76.5 & 90.6 & 52.1 & 78.2 & 58.5 & 89.3 & 82.5 & 83.4 & 81.3 & 49.1 & 77.2 & 62.4 & 83.8 & 68.7 \\
DSOD300~\cite{shen2017dsod} & DS/64-192-48-1 & 76.3 & 89.4 & 85.3 & 72.9 & 62.7 & 49.5 & 83.6 & 80.6 & 92.1 & 60.8 & 77.9 & 65.6 & 88.9 & 85.5 & 86.8 & 84.6 & 51.1 & 77.7 & 72.3 & 86.0 & 72.2 \\
SSD512*~\cite{liu2016ssd} & VGG-16 & 78.5 & 90.0 & 85.3 & 77.7 & 64.3 & 58.5 & 85.1 & 84.3 & 92.6 & 61.3 & 83.4 & 65.1 & 89.9 & 88.5 & 88.2 & 85.5 & 54.4 & 82.4 & 70.7 & 87.1 & 75.6 \\
SSD513~\cite{fu2017dssd} & ResNet-101 & 79.4 & 90.7 & 87.3 & 78.3 & 66.3 & 56.5 & 84.1 & 83.7 & 94.2 & 62.9 & 84.5 & 66.3 & 92.9 & 88.6 & 87.9 & 85.7 & 55.1 & 83.6 & 74.3 & 88.2 & 76.8 \\
DSSD513~\cite{fu2017dssd} & ResNet-101 & 80.0 & \textbf{92.1} & 86.6 & 80.3 & 68.7 & 58.2 & 84.3 & 85.0 & 94.6 & 63.3 & 85.9 & 65.6 & 93.0 & 88.5 & 87.8 & 86.4 & 57.4 & 85.2 & 73.4 & 87.8 & 76.8 \\
DES512~\cite{Zhang_2018_CVPR} & VGG-16 & 80.3 & 91.1 & 87.7 & 81.3 & 66.5 & 58.9 & 84.8 & \textbf{85.8} & 92.3 & 64.7 & 84.3 & 67.8 & 91.6 & 89.6 & 88.7 & 86.4 & 57.7 & 85.5 & 74.4 & 89.2 & \textbf{77.6} \\
PFPNet-R512~\cite{kim2018parallel} & VGG-16 & 80.3 & 91.6 & 85.8 & 82.0 & 70.0 & 64.4 & 84.8 & \textbf{85.8} & 91.2 & 63.6 & 85.6 & 64.1 & 90.0 & 88.5 & 87.8 & 87.4 & 59.1 & 87.4 & 73.0 & 88.2 & 76.1 \\
RefineDet512~\cite{zhang2018single} & VGG-16 & 80.1 & 90.2 & 86.8 & 81.8 & 68.0 & 65.6 & 84.9 & 85.0 & 92.2 & 62.0 & 84.4 & 64.9 & 90.6 & 88.3 & 87.2 & 87.8 & 58.0 & 86.3 & 72.5 & 88.7 & 76.6 \\
AP-loss500 & ResNet-101 & \textbf{83.1} & 90.4 & \textbf{88.6} & \textbf{84.6} & \textbf{73.4} & \textbf{69.3} & \textbf{86.2} & \textbf{85.8} & \textbf{94.8} & \textbf{69.2} & \textbf{88.9} & \textbf{68.2} & \textbf{94.2} & \textbf{90.6} & \textbf{90.1} & \textbf{89.9} & \textbf{64.3} & \textbf{88.3} & \textbf{76.8} & \textbf{90.1} & 77.5 \\
\midrule[1pt]
PFPNet-R512+~\cite{kim2018parallel} & VGG-16 & 83.7 & \textbf{93.1} & 89.8 & 85.5 & \textbf{75.0} & 71.6 & 87.7 & \textbf{89.6} & 93.7 & 69.1 & \textbf{88.2} & 66.6 & 92.4 & 90.6 & \textbf{90.7} & 90.1 & 64.0 & \textbf{89.9} & 75.5 & 88.7 & \textbf{81.6} \\
RefineDet512+~\cite{zhang2018single} & VGG-16 & 83.5 & 92.2 & 89.4 & 85.0 & 74.1 & 70.8 & 87.0 & 88.7 & 94.0 & 68.6 & 87.1 & 68.2 & 92.5 & 90.8 & 89.4 & 90.2 & 64.1 & 89.8 & 75.2 & \textbf{90.7} & 81.1 \\
AP-loss500+ & ResNet-101 & \textbf{84.5} & 91.8 & \textbf{90.2} & \textbf{87.2} & \textbf{75.0} & \textbf{73.7} & \textbf{87.9} & 88.5 & \textbf{95.4} & \textbf{72.0} & \textbf{88.2} & \textbf{68.8} & \textbf{94.3} & \textbf{91.3} & 90.3 & \textbf{91.5} & \textbf{67.0} & 89.6 & \textbf{77.8} & 90.2 & 80.3 \\
\midrule[1pt]
\end{tabular}}}
\vspace{-0mm}
\begin{tablenotes}
\item + denotes multi-scale testing.
\end{tablenotes}
\label{state-of-the-art-voc12}
\end{table*}

\begin{table*}[t]
\footnotesize
\centering
\caption{Detection results on COCO {\tt test-dev} set.}
\vspace{-2mm}
\setlength{\tabcolsep}{2mm}{
\begin{tabular}{cccccccccccccc}
\midrule[1pt]
Method & Backbone & AP & AP\(_{50}\) & AP\(_{75}\) & AP\(_S\) & AP\(_M\) & AP\(_L\) & AR\(_1\) & AR\(_{10}\) & AR\(_{100}\) & AR\(_S\) & AR\(_M\) & AR\(_L\) \\
\midrule[1pt]
Two-Stage: & & & & & & & & & & & & & \\
%CoupleNet~\cite{zhu2017couplenet} & ResNet-101 & 34.4 & 54.8 & 37.2 & 13.4 & 38.1 & 50.8 & 30.0 & 45.0 & 46.4 & 20.7 & 53.1 & 68.5 \\
%FPN~\cite{lin2017feature} & ResNet-101 & 36.2 & 59.1 & 39.0 & 18.2 & 39.0 & 48.2 & - & - & - & - & - & - \\
Def-R-FCN~\cite{dai2017deformable} & A-Inc-Res & 37.5 & 58.0 & 40.8 & 19.4 & 40.1 & 52.5 & - & - & - & - & - & - \\
Mask-RCNN~\cite{he2017mask} & ResNet-101 & 38.2 & 60.3 & 41.7 & 20.1 & 41.1 & 50.2 & - & - & - & - & - & - \\
Libra-RCNN~\cite{pang2019libra} & ResNet-101 & 41.1 & 62.1 & 44.7 & 23.4 & 43.7 & 52.5 & - & - & - & - & - & - \\
Cascade-RCNN~\cite{cai2018cascade} & ResNet-101 & 42.8 & 62.1 & 46.3 & 23.7 & 45.5 & 55.2 & - & - & - & - & - & - \\
Revisiting-RCNN~\cite{cheng2018revisiting} & ResNet-101+152 & 43.1 & 66.1 & 47.3 & 25.8 & 45.9 & 55.3 & - & - & - & - & - & - \\
Grid-RCNN~\cite{lu2019grid} & ResNeXt-101 & 43.2 & 63.0 & 46.6 & 25.1 & 46.5 & 55.2 & - & - & - & - & - & - \\
SNIP~\cite{singh2018snip} & DPN-98 & 45.7 & 67.3 & 51.1 & \textbf{29.3} & 48.8 & 57.1 & - & - & - & - & - & - \\
TridentNet~\cite{li2019scale} & ResNet-101-Def & \textbf{46.8} & \textbf{67.6} & \textbf{51.5} & 28.0 & \textbf{51.2} & \textbf{60.5} & - & - & - & - & - & - \\
\midrule[1pt]
\midrule[1pt]
One-Stage: & & & & & & & & & & & & & \\
YOLOv2~\cite{redmon2017yolo9000} & DarkNet-19 & 21.6 & 44.0 & 19.2 & 5.0 & 22.4 & 35.5 & 20.7 & 31.6 & 33.3 & 9.8 & 36.5 & 54.4 \\
DSOD300~\cite{shen2017dsod} & DS/64-192-48-1 & 29.3 & 47.3 & 30.6 & 9.4 & 31.5 & 47.0 & 27.3 & 40.7 & 43.0 & 16.7 & 47.1 & 65.0 \\
SSD512*~\cite{liu2016ssd} & VGG-16 & 28.8 & 48.5 & 30.3 & 10.9 & 31.8 & 43.5 & 26.1 & 39.5 & 42.0 & 16.5 & 46.6 & 60.8 \\
%SSD513~\cite{fu2017dssd} & ResNet-101 & 31.2 & 50.4 & 33.3 & 10.2 & 34.5 & 49.8 & 28.3 & 42.1 & 44.4 & 17.6 & 49.2 & 65.8 \\
DSSD513~\cite{fu2017dssd} & ResNet-101 & 33.2 & 53.3 & 35.2 & 13.0 & 35.4 & 51.1 & 28.9 & 43.5 & 46.2 & 21.8 & 49.1 & 66.4 \\
DES512~\cite{Zhang_2018_CVPR} & VGG-16 & 32.8 & 53.2 & 34.6 & 13.9 & 36.0 & 47.6 & 28.4 & 43.5 & 46.2 & 21.6 & 50.7 & 64.6 \\
RFBNet512~\cite{Liu_2018_ECCV} & VGG-16 & 33.8 & 54.2 & 35.9 & 16.2 & 37.1 & 47.4 & - & - & - & - & - & - \\
PFPNet-R512~\cite{kim2018parallel} & VGG-16 & 35.2 & 57.6 & 37.9 & 18.7 & 38.6 & 45.9 & - & - & - & - & - & - \\
RefineDet512~\cite{zhang2018single} & VGG-16 & 33.0 & 54.5 & 35.5 & 16.3 & 36.3 & 44.3 & 28.3 & 46.4 & 50.6 & 29.3 & 55.5 & 66.0 \\
RefineDet512~\cite{zhang2018single} & ResNet-101 & 36.4 & 57.5 & 39.5 & 16.6 & 39.9 & 51.4 & 30.6 & 49.0 & 53.0 & 30.0 & 58.2 & 70.3 \\
RetinaNet500~\cite{lin2018focal} & ResNet-101 & 34.4 & 53.1 & 36.8 & 14.7 & 38.5 & 49.1 & - & - & - & - & - & - \\
RetinaNet800~\cite{lin2018focal} & ResNet-101 & 39.1 & 59.1 & 42.3 & 21.8 & 42.7 & 50.2 & - & - & - & - & - & - \\
CornerNet~\cite{law2018cornernet} & Hourglass-104 & 40.5 & 56.5 & 43.1 & 19.4 & 42.7 & 53.9 & \textbf{35.3} & \textbf{54.3} & \textbf{59.1} & 37.4 & \textbf{61.9} & \textbf{76.9} \\
GHM800~\cite{li2018gradient} & ResNet-101 & 39.9 & 60.8 & 42.5 & 20.3 & 43.6 & \textbf{54.1} & - & - & - & - & - & - \\
DR-loss\(_{\text{fixed}}\)800~\cite{qian2019dr} & ResNet-101 & 40.6 & 60.7 & \textbf{43.9} & 22.9 & 43.7 & 51.9 & - & - & - & - & - & - \\
AP-loss500 & ResNet-101 & 37.4 & 58.6 & 40.5 & 17.3 & 40.8 & 51.9 & 31.3 & 50.9 & 54.1 & 29.8 & 59.5 & 73.6 \\
AP-loss800 & ResNet-101 & \textbf{40.8} & \textbf{63.7} & 43.7 & \textbf{25.4} & \textbf{43.9} & 50.6 & 33.2 & 53.3 & 56.9 & \textbf{39.5} & 60.3 & 69.9 \\
\midrule[1pt]
PFPNet-R512+~\cite{kim2018parallel} & VGG-16 & 39.4 & 61.5 & 42.6 & 25.3 & 42.3 & 48.8 & - & - & - & - & - & - \\
RefineDet512+~\cite{zhang2018single} & VGG-16 & 37.6 & 58.7 & 40.8 & 22.7 & 40.3 & 48.3 & 31.4 & 52.4 & 61.3 & 41.6 & 65.8 & 75.4 \\
RefineDet512+~\cite{zhang2018single} & ResNet-101 & 41.8 & 62.9 & 45.7 & \textbf{25.6} & \textbf{45.1} & \textbf{54.1} & 34.0 & \textbf{56.3} & \textbf{65.5} & \textbf{46.2} & \textbf{70.2} & \textbf{79.8} \\
AP-loss500+ & ResNet-101 & \textbf{42.1} & \textbf{63.5} & \textbf{46.4} & \textbf{25.6} & 45.0 & 53.9 & \textbf{34.1} & 55.9 & 60.5 & 39.7 & 64.5 & 77.6 \\
\midrule[1pt]
\end{tabular}}
\vspace{-0mm}
\begin{tablenotes}
\item + denotes multi-scale testing.
\end{tablenotes}
\label{state-of-the-art-coco}
\end{table*}

\subsection{Benchmark Results}\label{section-benchmark-results}
\begin{figure*}[t!]
\small
\centering
  \includegraphics[width=0.187\linewidth]{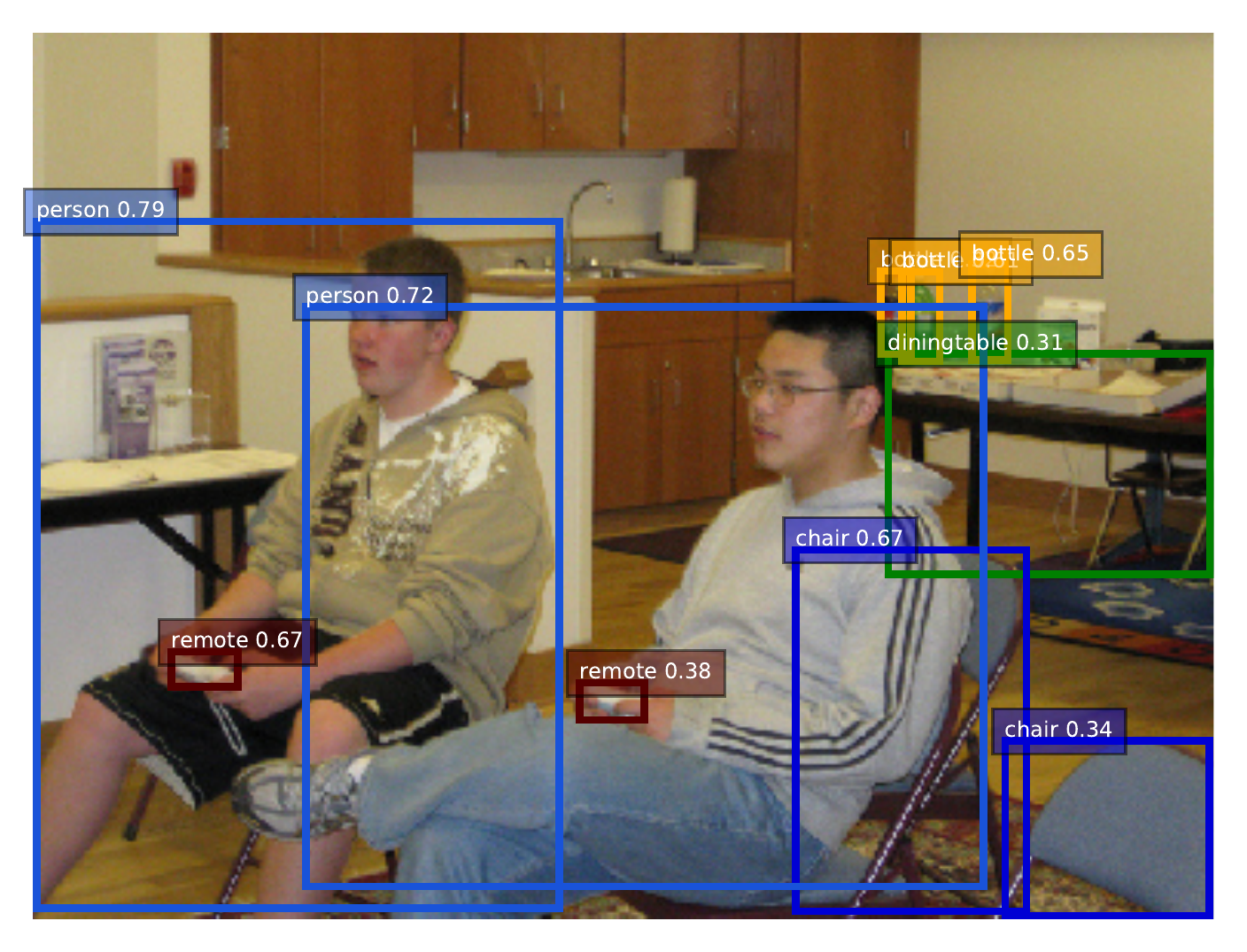}
  \includegraphics[width=0.207\linewidth]{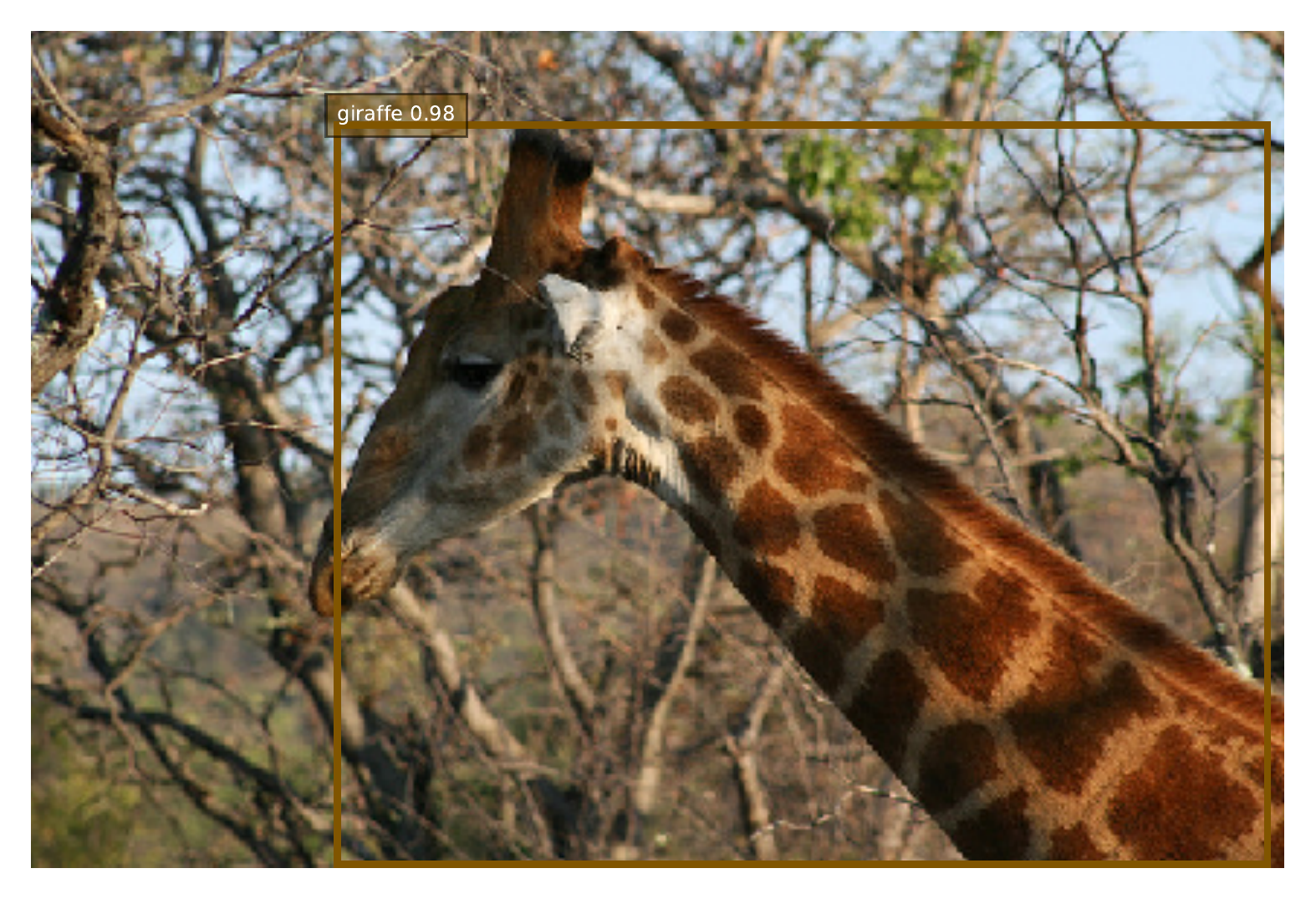}
  \includegraphics[width=0.185\linewidth]{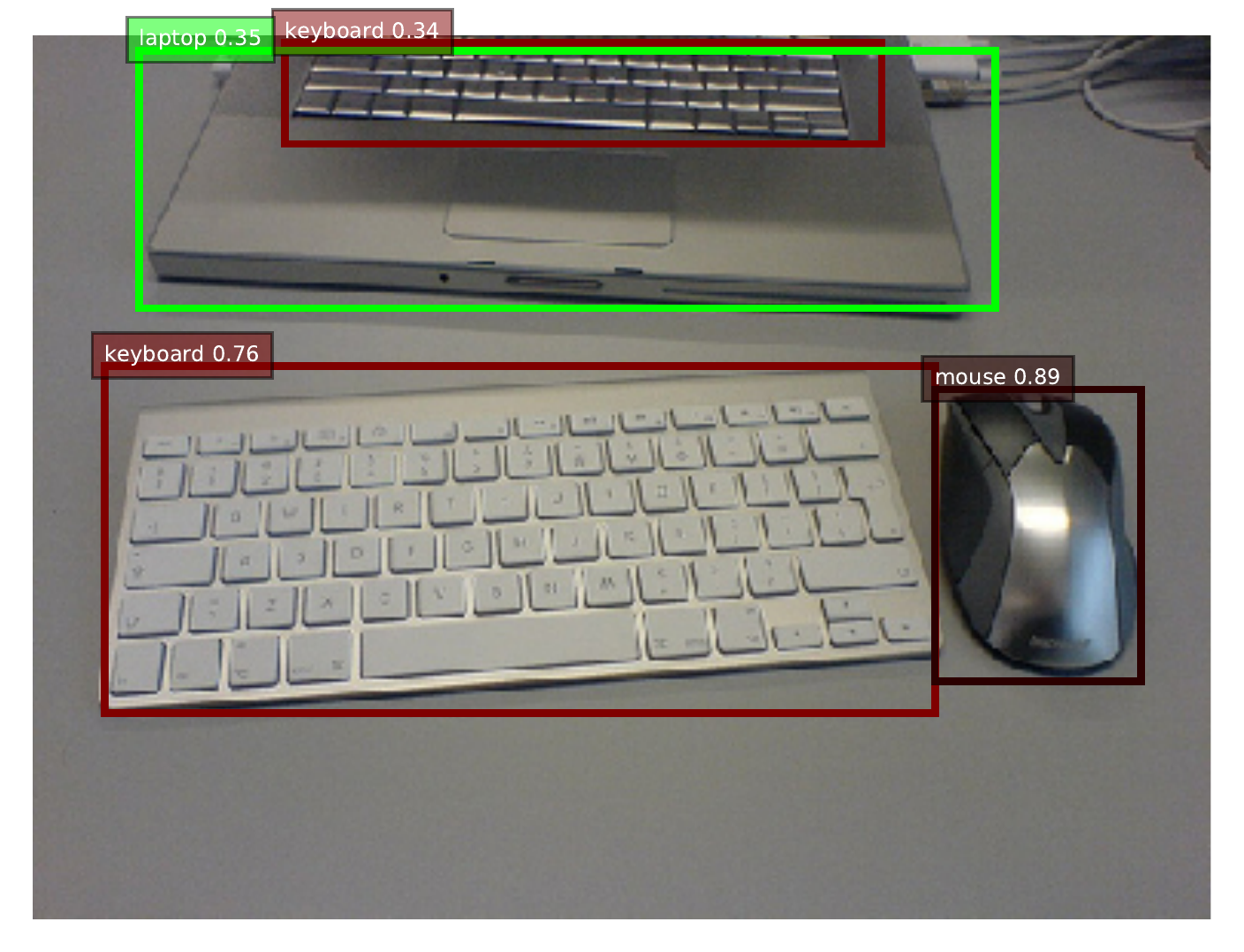}
  \includegraphics[width=0.21\linewidth]{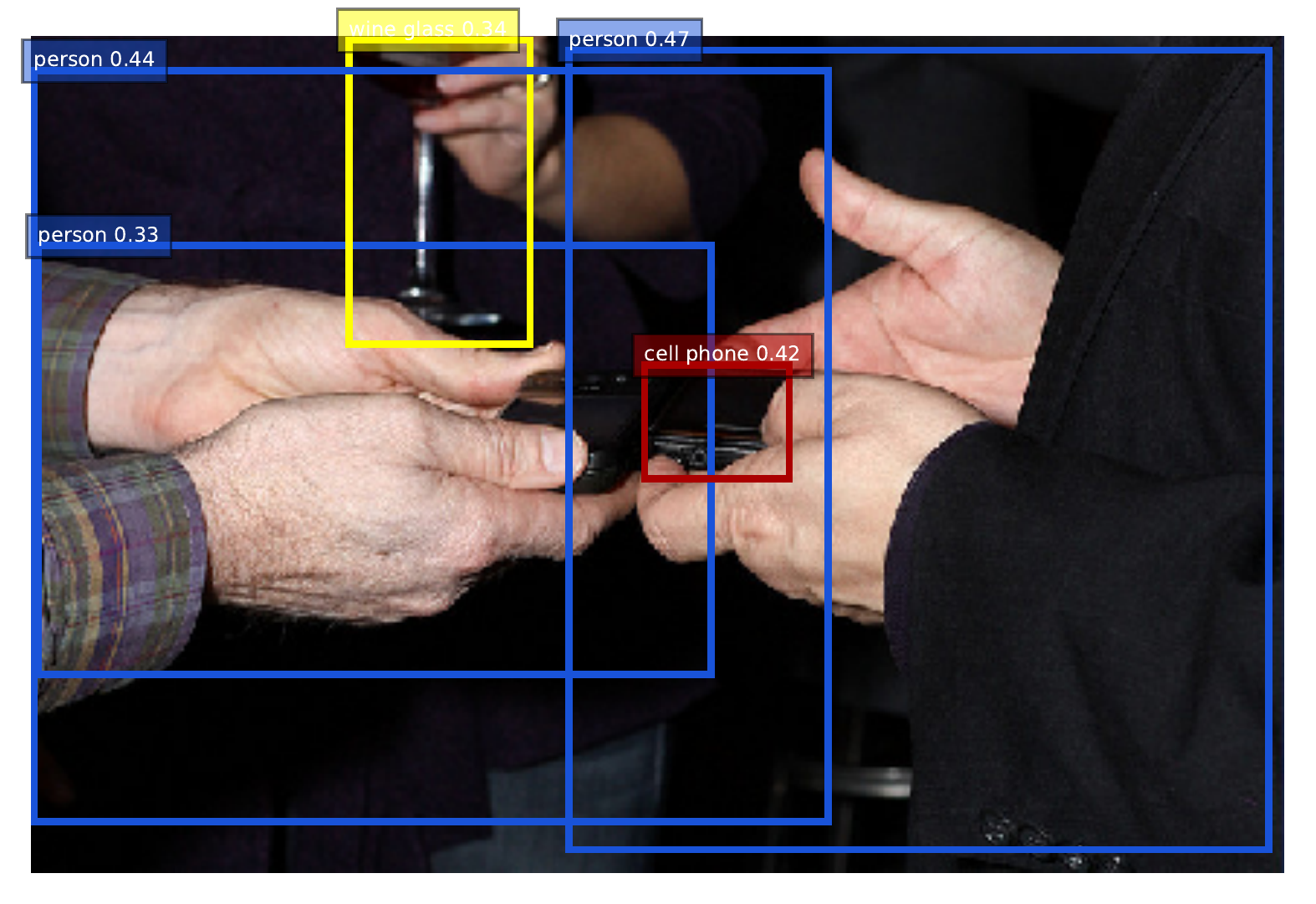}
  \includegraphics[width=0.19\linewidth]{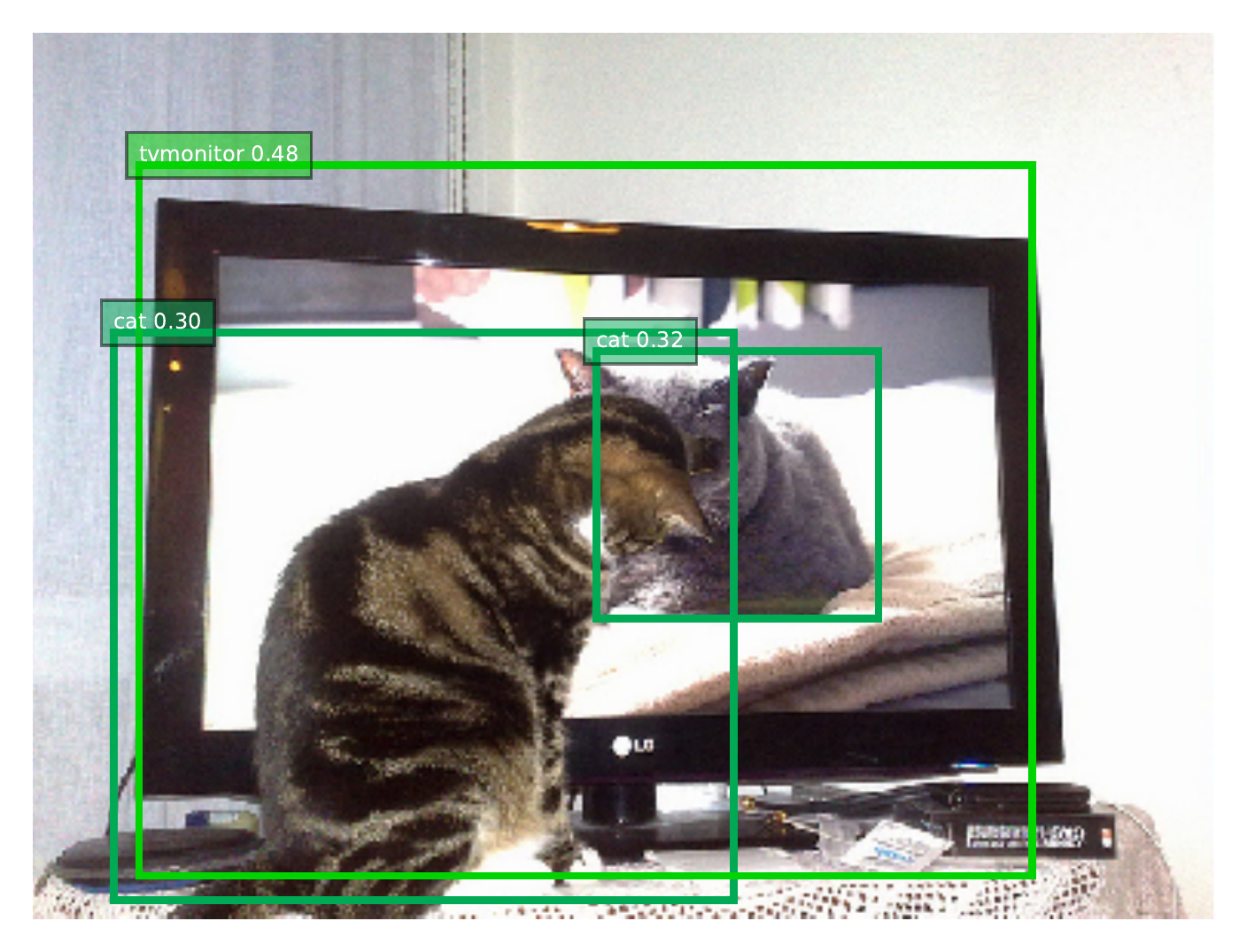}
%\par
%\begin{subfigure}[b]{0.186\linewidth}
%  \centering
%  \includegraphics[width=1.0\linewidth]{figs/fig-results2/ours_42.pdf}
%\end{subfigure}
%\begin{subfigure}[b]{0.21\linewidth}
%  \centering
%  \includegraphics[width=1.0\linewidth]{figs/fig-results2/ours_55.pdf}
%\end{subfigure}
%\begin{subfigure}[b]{0.186\linewidth}
%  \centering
%  \includegraphics[width=1.0\linewidth]{figs/fig-results2/ours_56.pdf}
%\end{subfigure}
%\begin{subfigure}[b]{0.20\linewidth}
%  \centering
%  \includegraphics[width=1.0\linewidth]{figs/fig-results2/ours_89.pdf}
%\end{subfigure}
%\begin{subfigure}[b]{0.20\linewidth}
%  \centering
%  \includegraphics[width=1.0\linewidth]{figs/fig-results2/ours_103.pdf}
%\end{subfigure}
\par
  \includegraphics[width=0.19\linewidth]{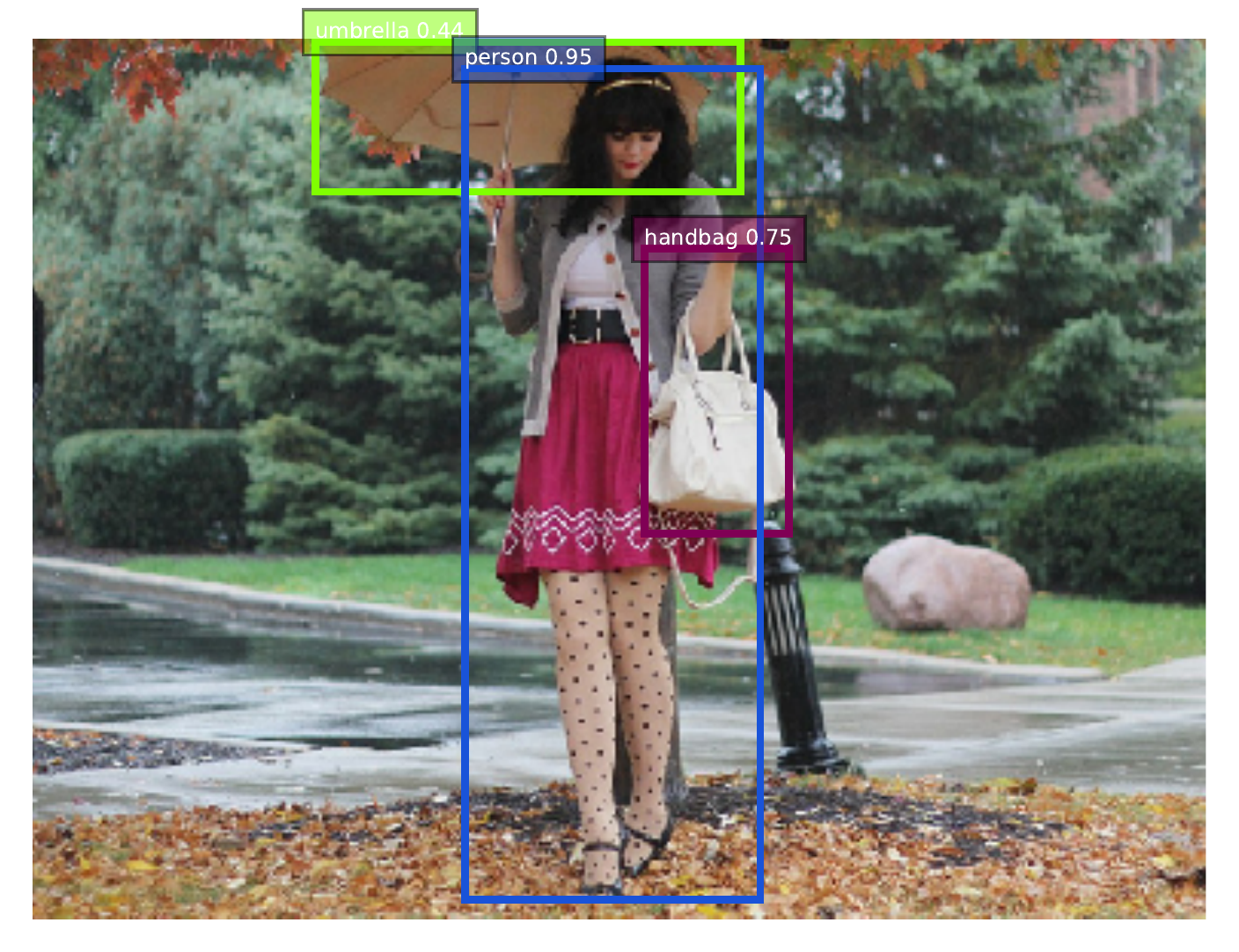}
  \includegraphics[width=0.19\linewidth]{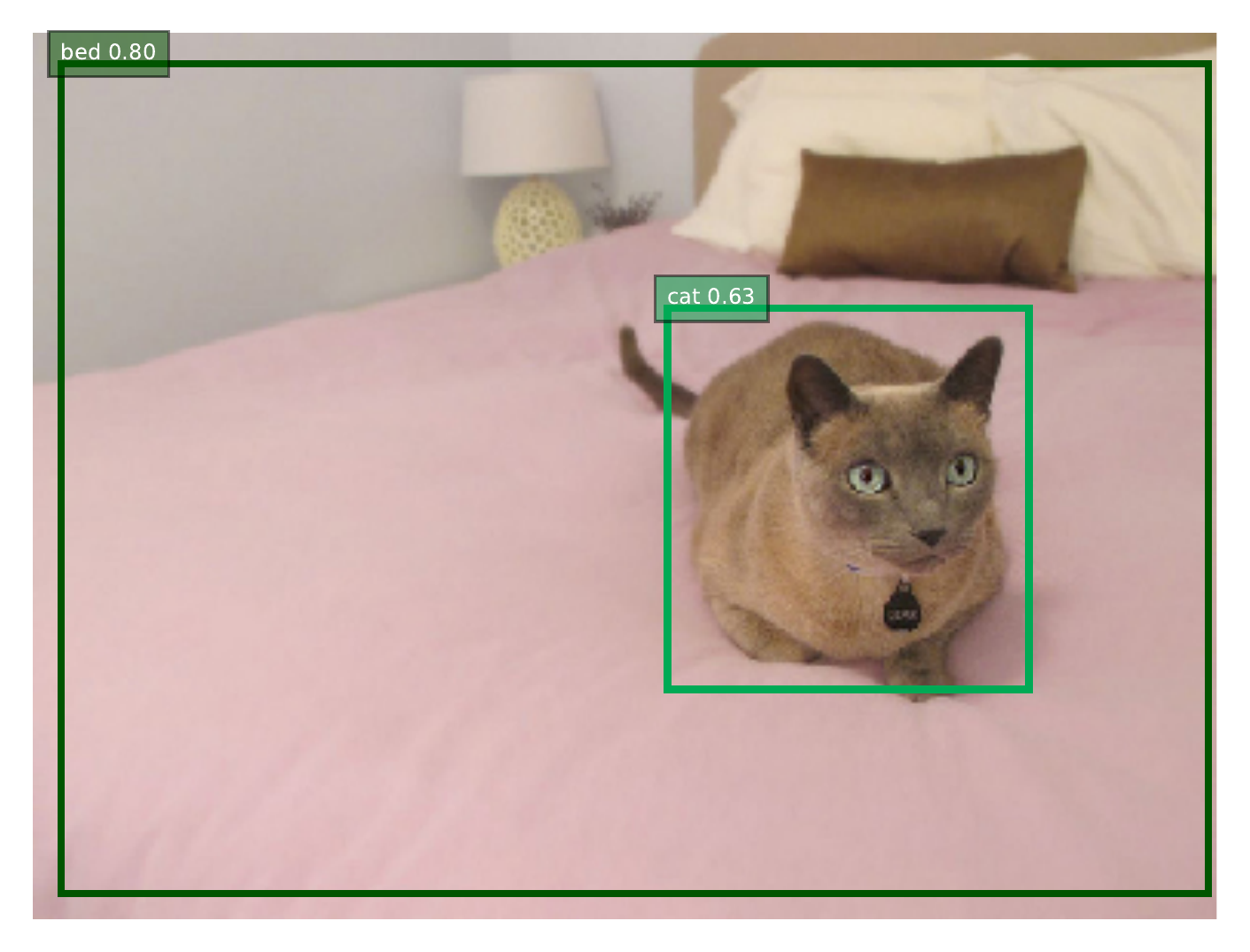}
  \includegraphics[width=0.175\linewidth]{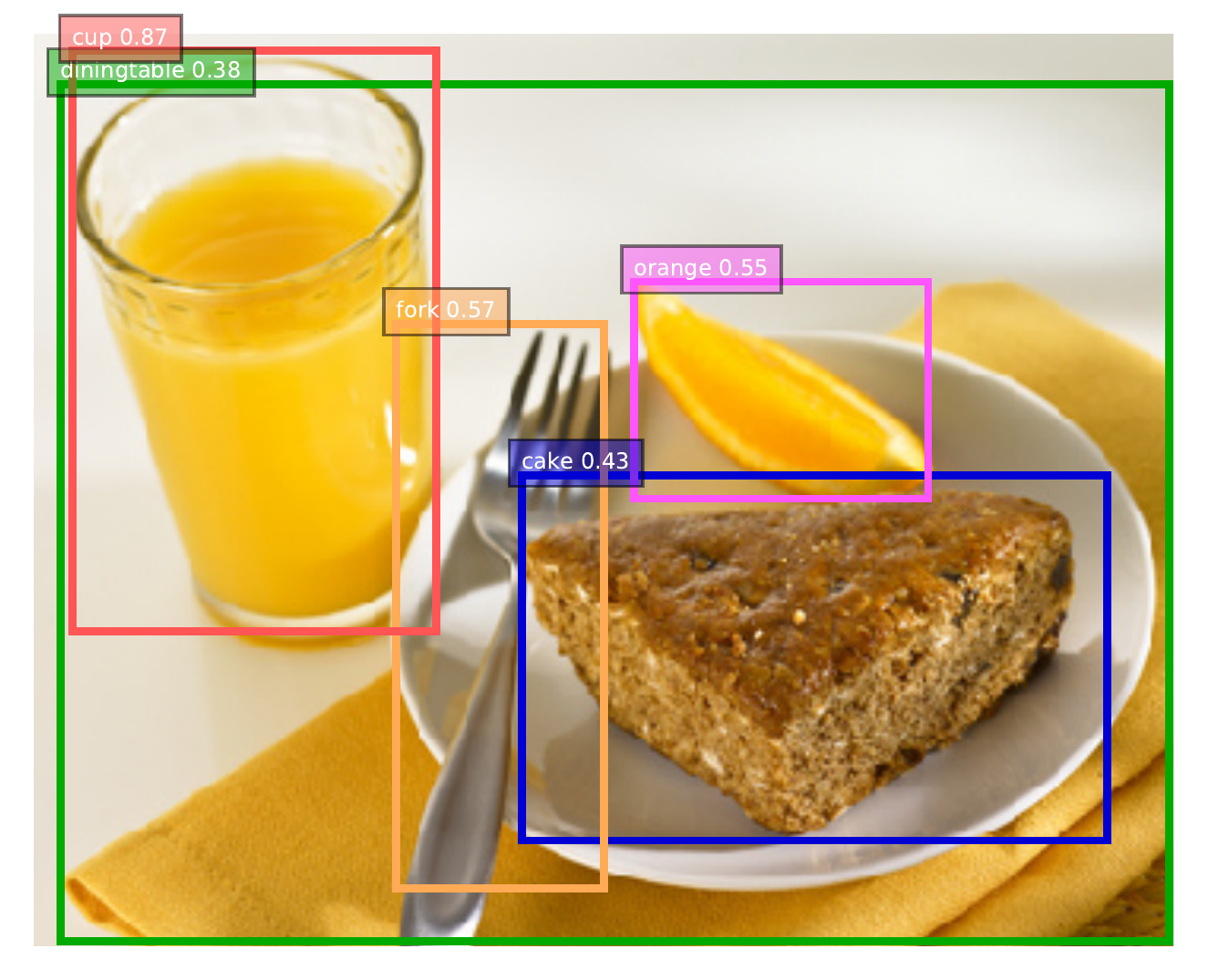}
  \includegraphics[width=0.22\linewidth]{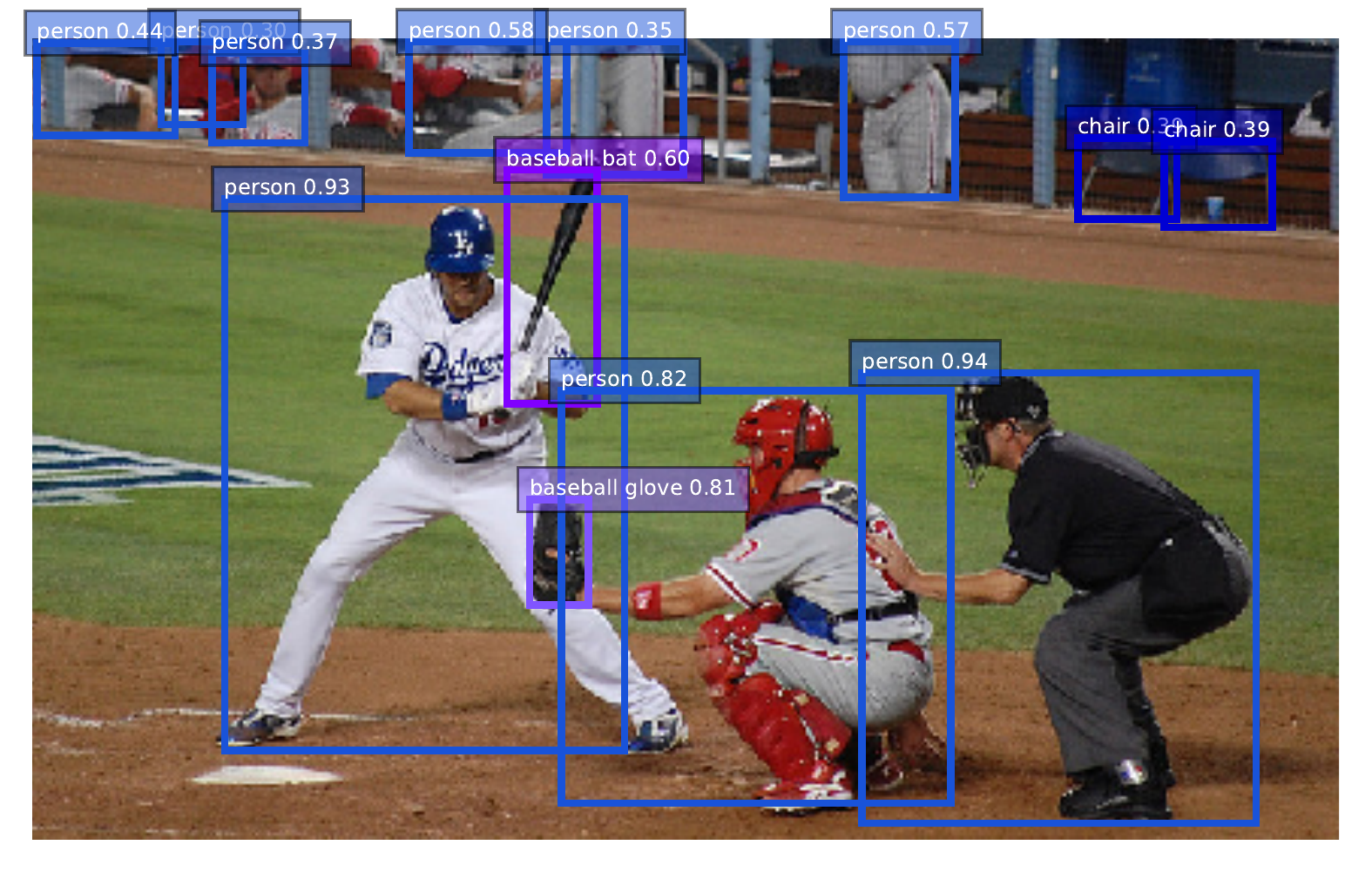}
  \includegraphics[width=0.20\linewidth]{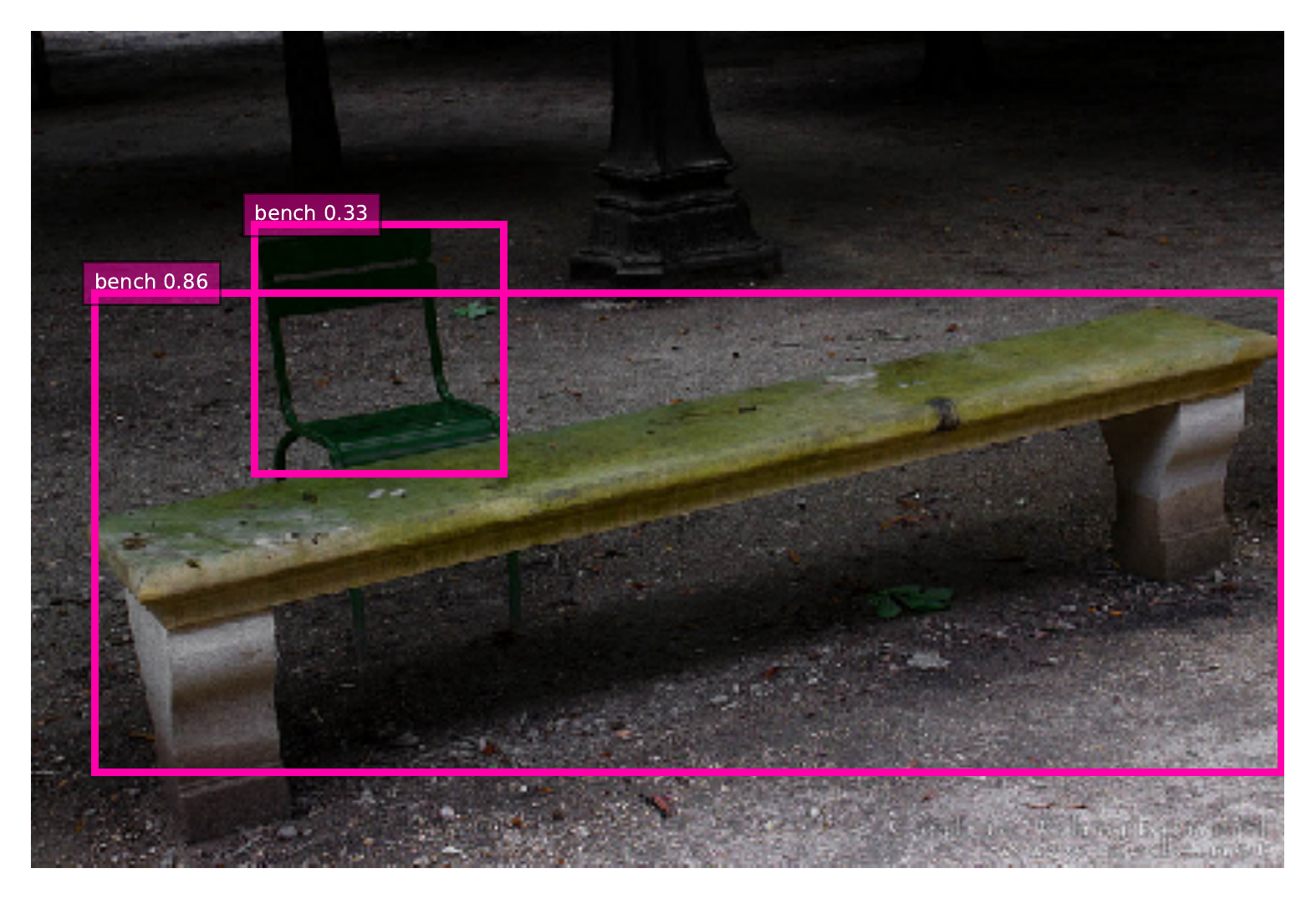}
\par
  \includegraphics[width=0.19\linewidth]{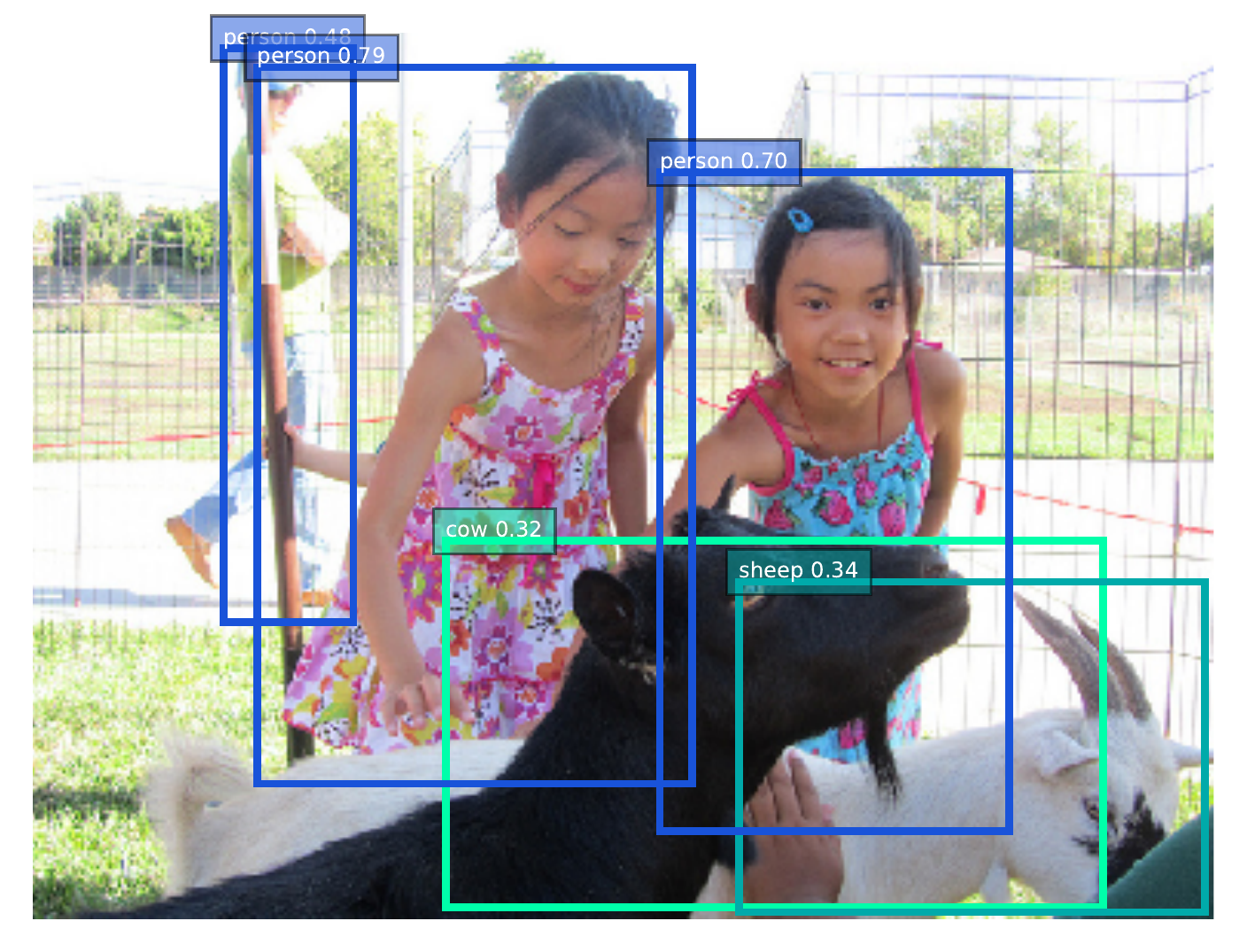}
  \includegraphics[width=0.22\linewidth]{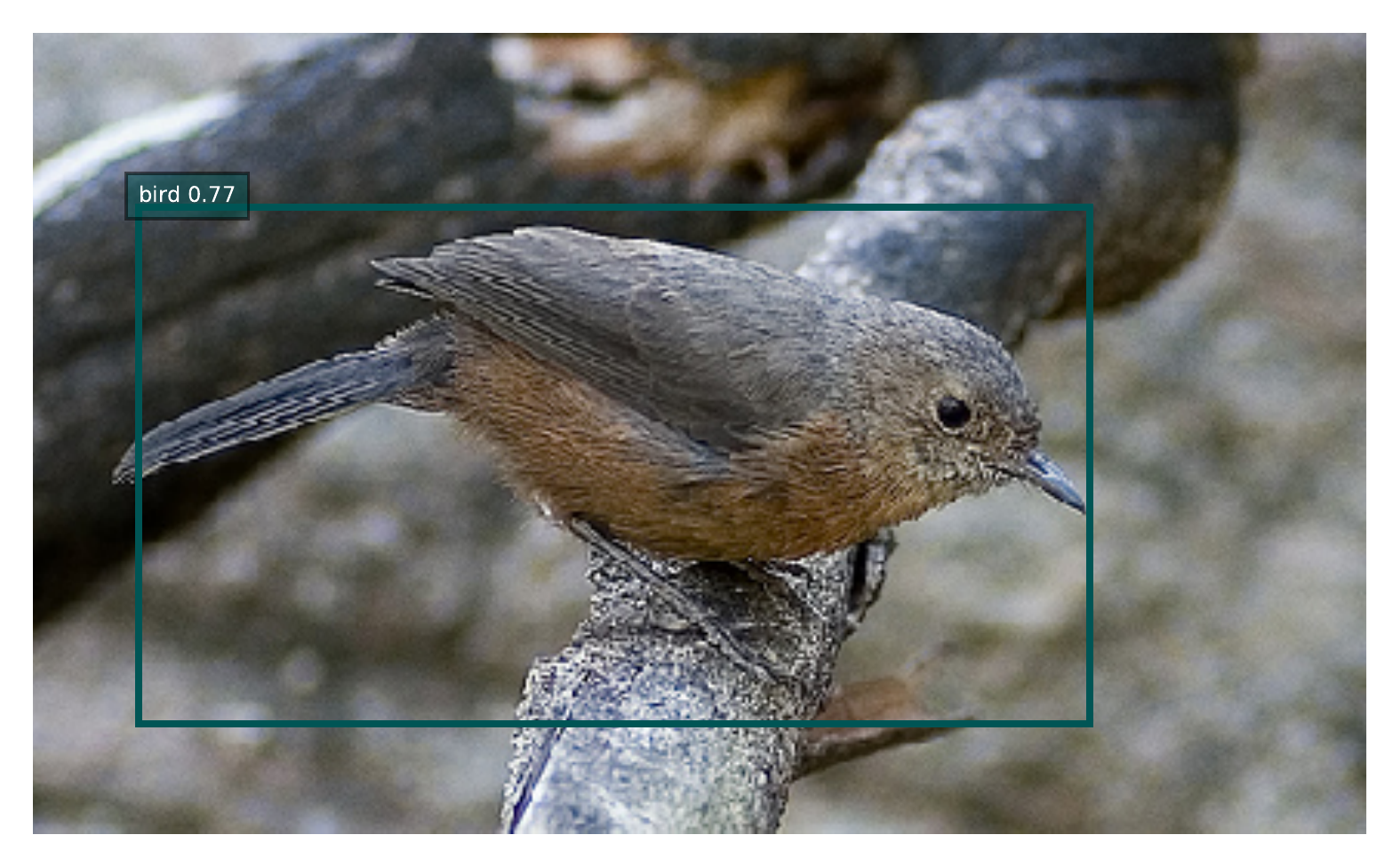}
  \includegraphics[width=0.19\linewidth]{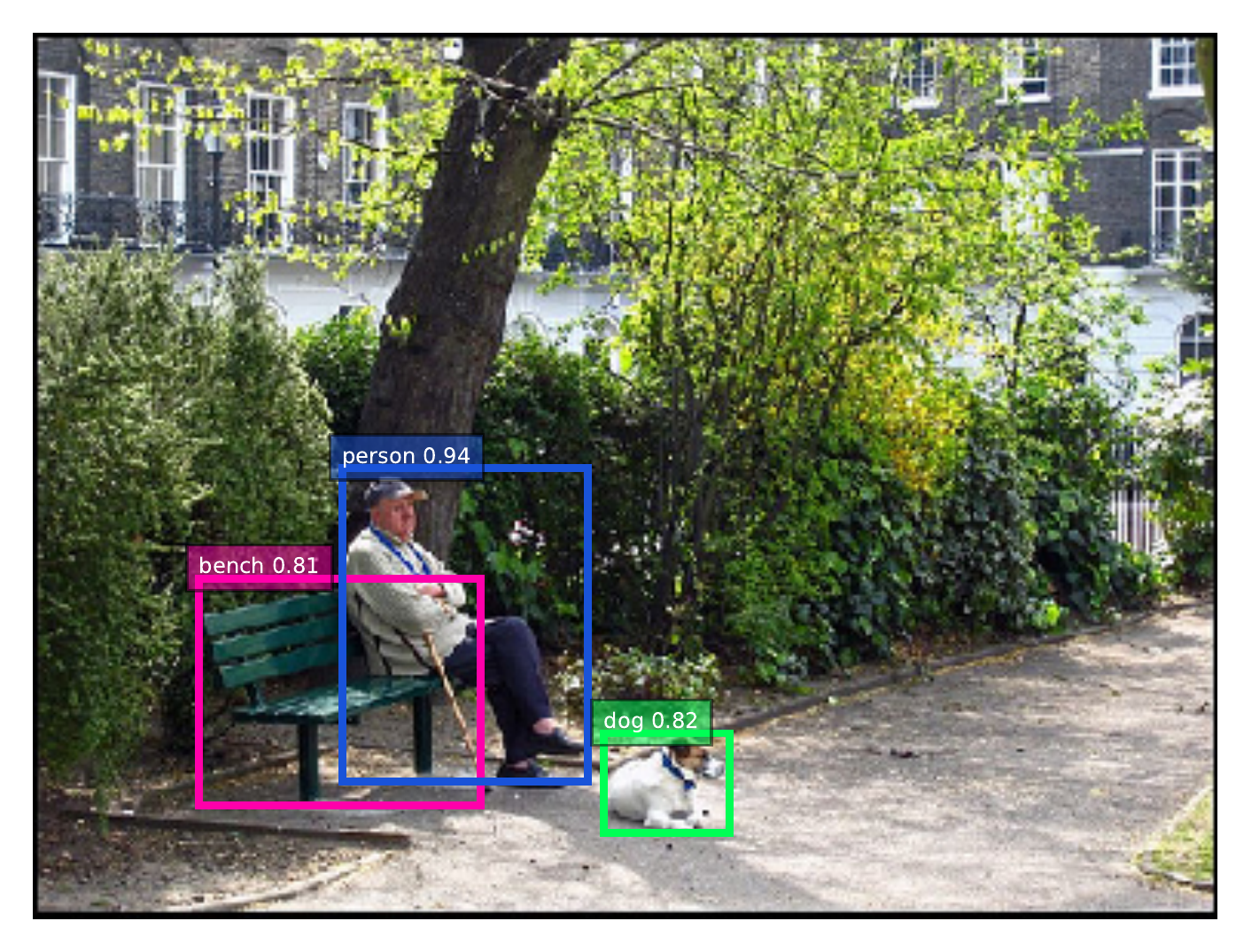}
  \includegraphics[width=0.19\linewidth]{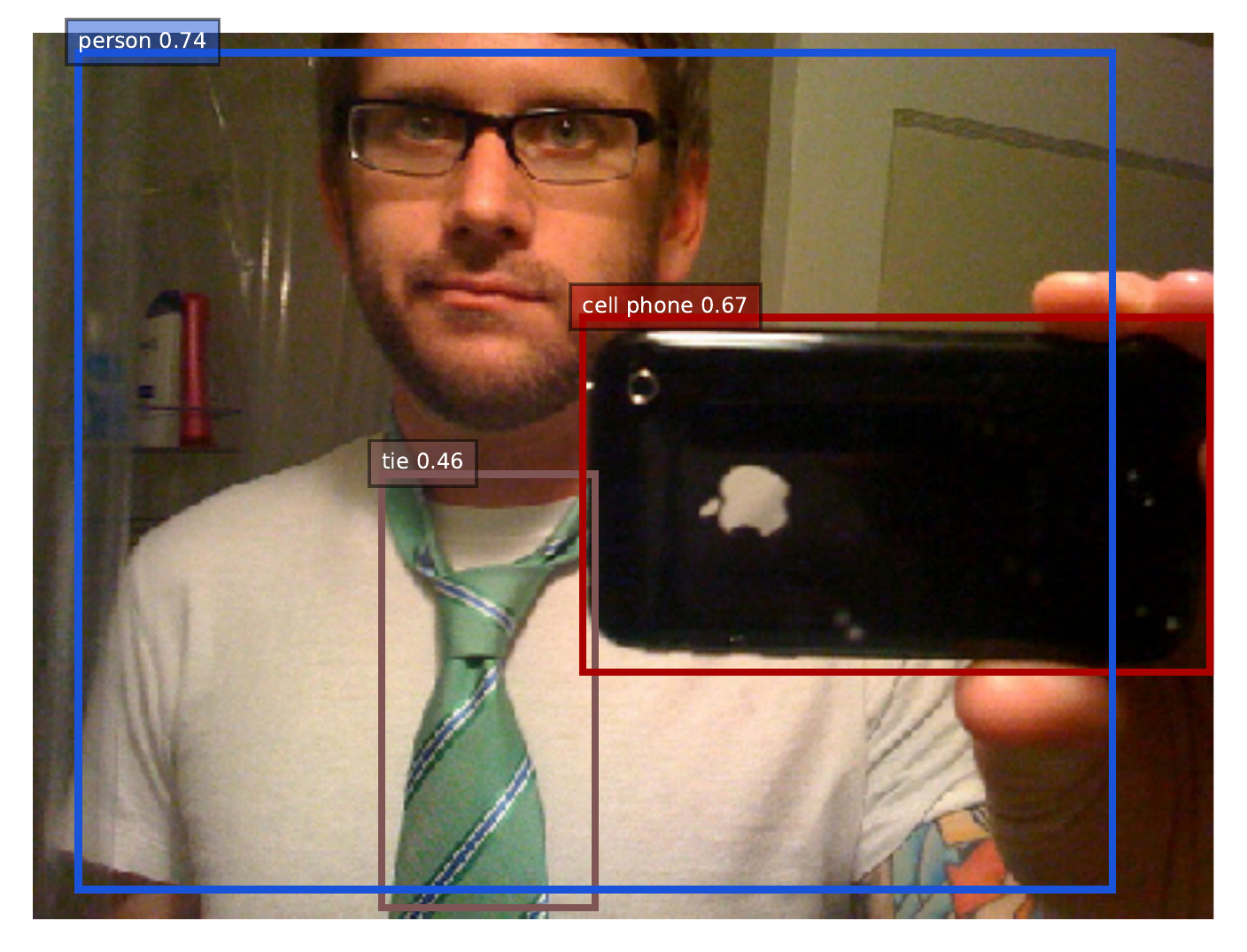}
  \includegraphics[width=0.19\linewidth]{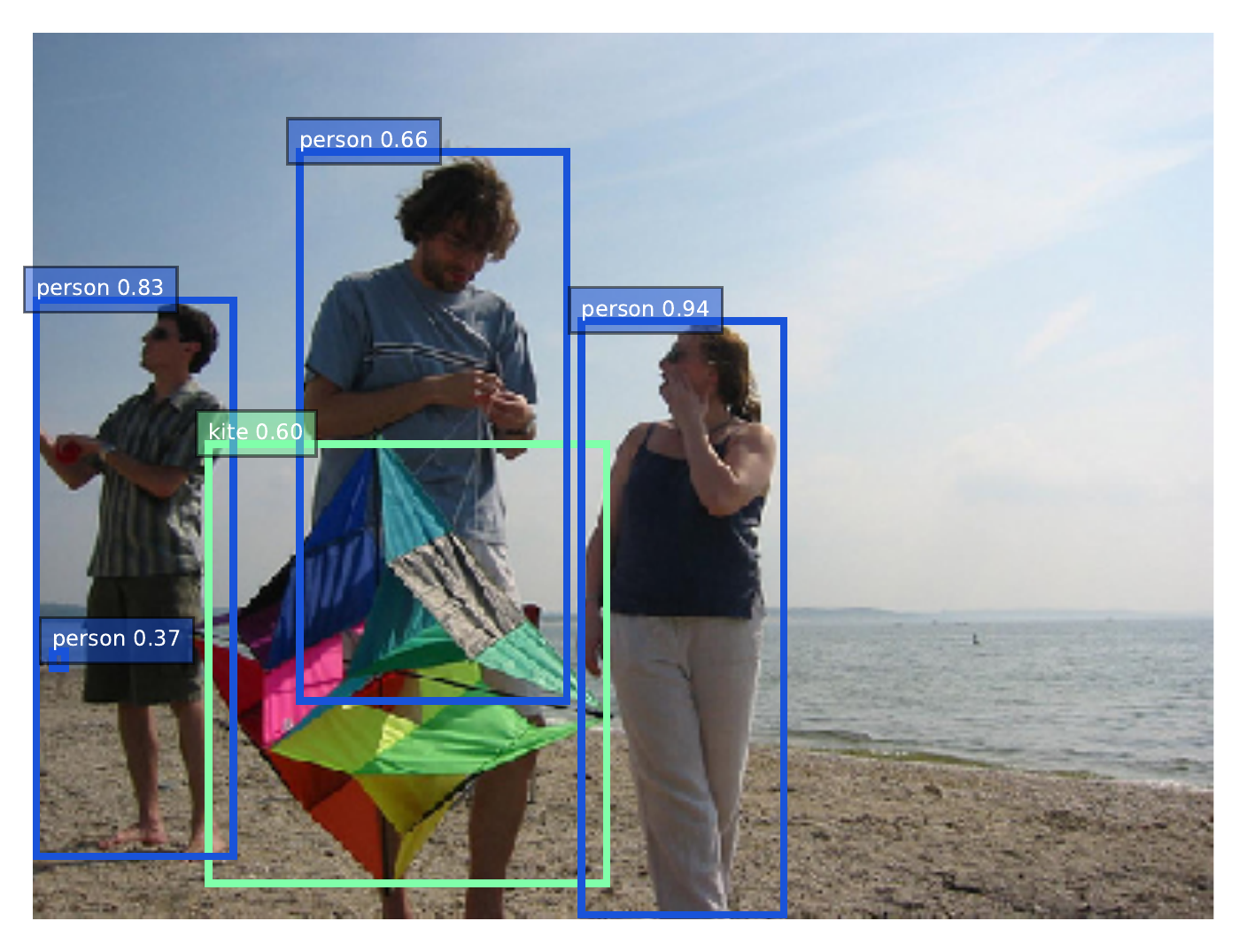}
\par
  \includegraphics[width=0.19\linewidth]{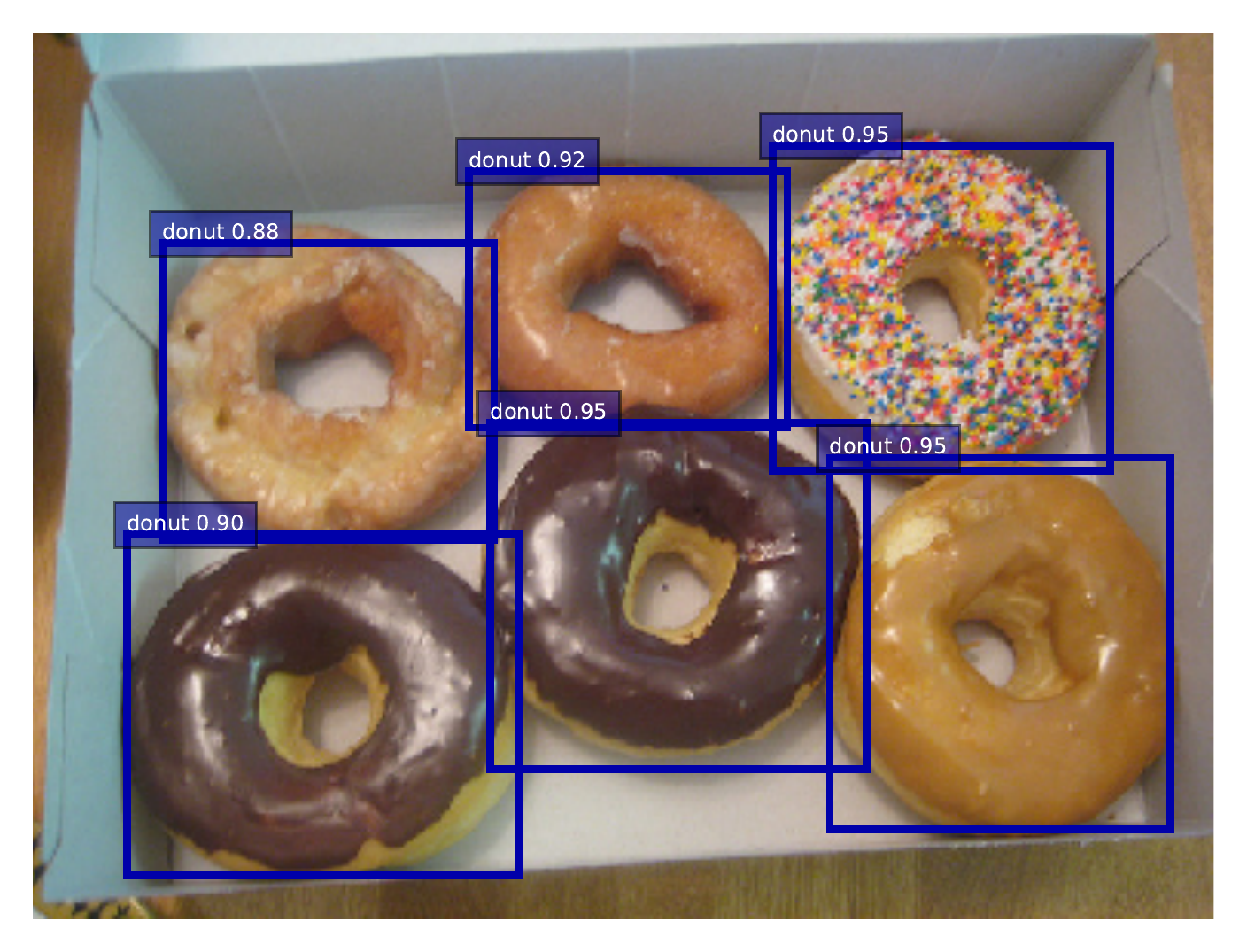}
  \includegraphics[width=0.21\linewidth]{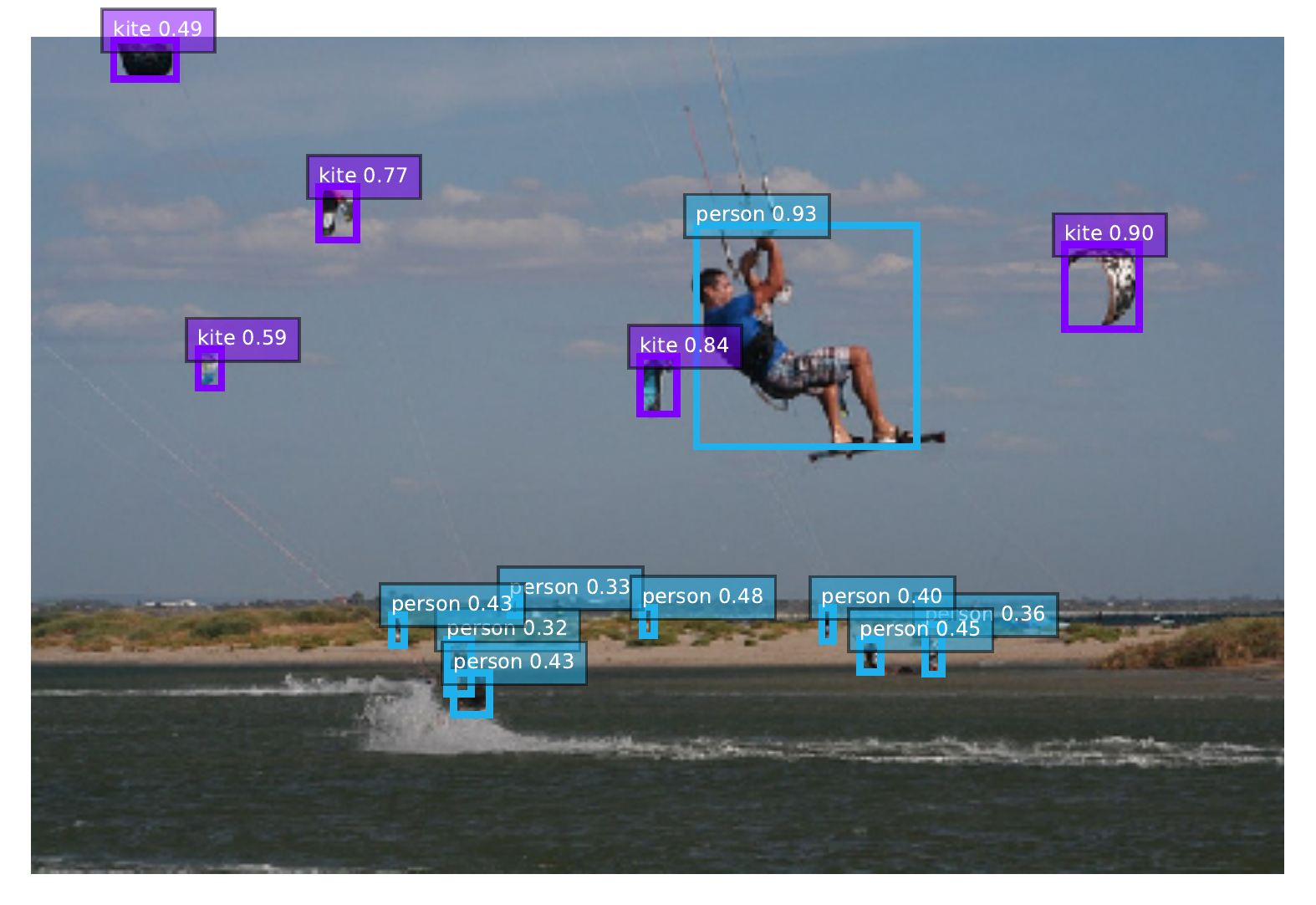}
  \includegraphics[width=0.205\linewidth]{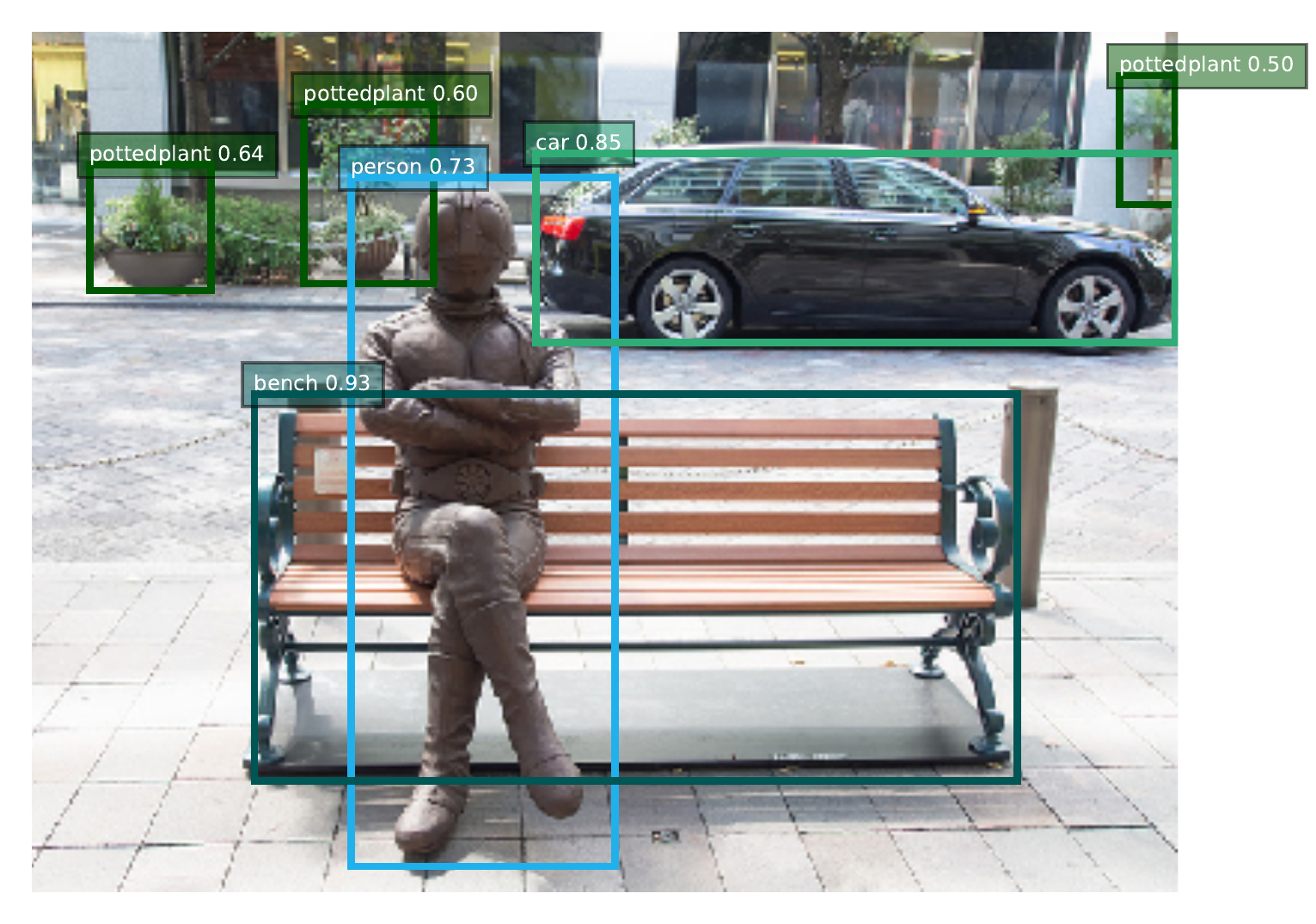}
  \includegraphics[width=0.186\linewidth]{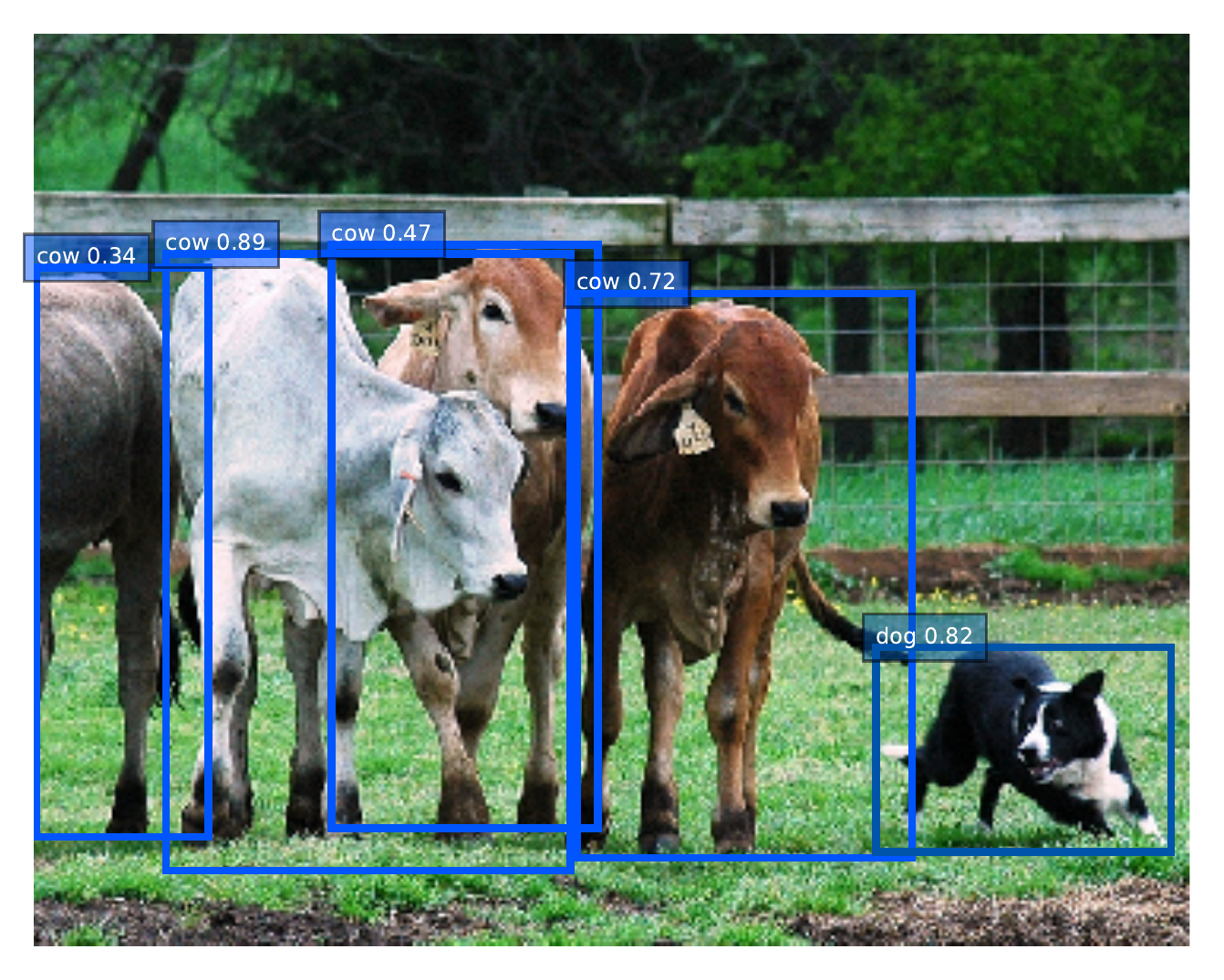}
  \includegraphics[width=0.19\linewidth]{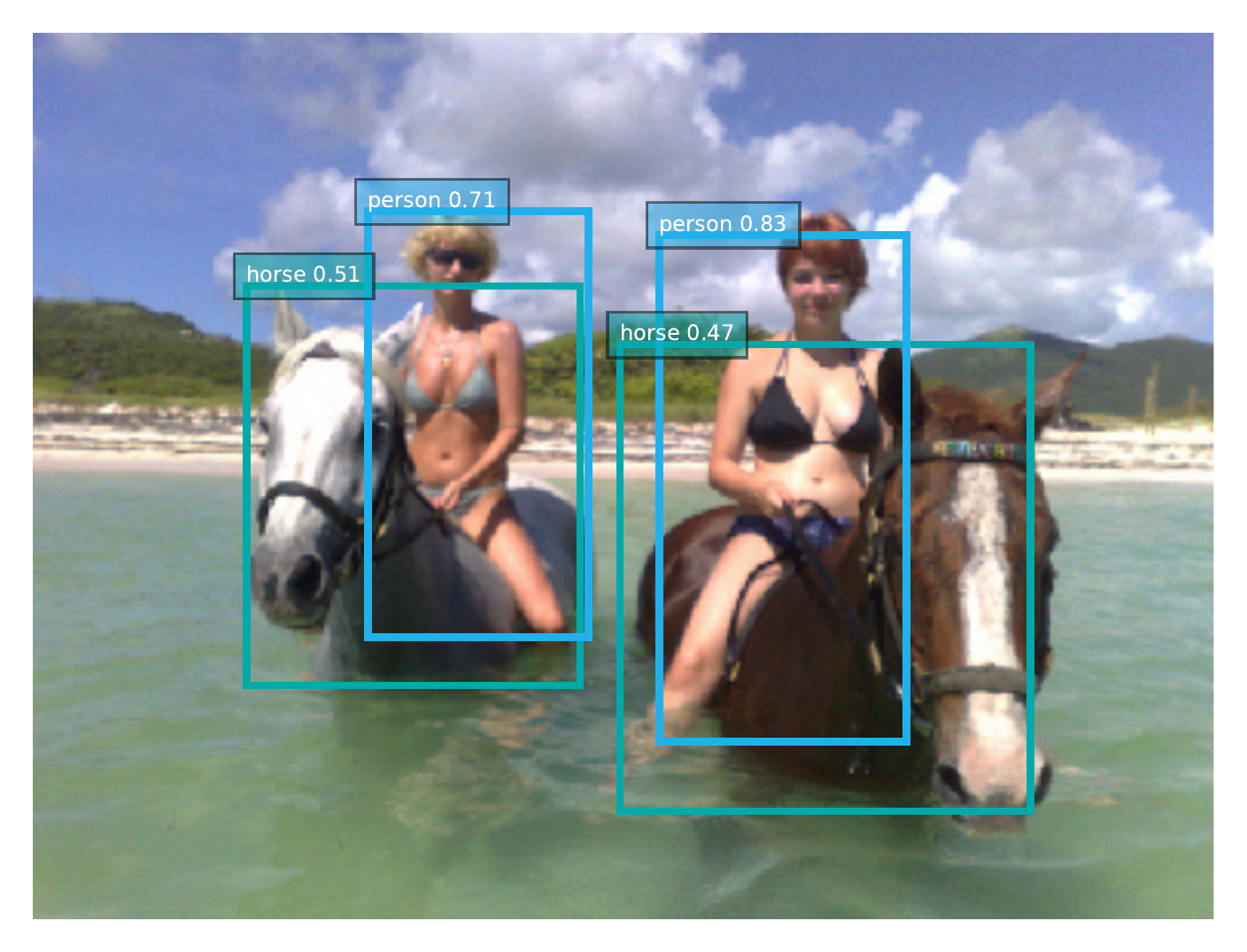}
\vspace{-5mm}
  \caption{Some detection examples on COCO dataset. The detector is RetinaNet-500 trained by AP-loss, and the backbone model is ResNet-101.}
\vspace{-1mm}
\label{fig-result2}
\end{figure*}

With the settings selected in ablation study, we conduct experiments to compare the proposed detector to state-of-the-art one-stage detectors on three widely used benchmark, \textit{i.e.} VOC2007 {\tt test}, VOC2012 {\tt test} and COCO {\tt test-dev} sets. Our detector is RetinaNet trained with AP-loss. We use ResNet-101 as backbone networks instead of ResNet-50 in ablation study. We resize the input image to 500/800 pixels for its shorter side in testing (for the detector AP-loss500/800 respectively). \autoref{state-of-the-art-voc07}, \autoref{state-of-the-art-voc12}, \autoref{state-of-the-art-coco} list the benchmark results comparing to recent state-of-the-art one-stage detectors such as DSOD~\cite{shen2017dsod}, DES~\cite{Zhang_2018_CVPR}, RetinaNet~\cite{lin2018focal}, RefineDet~\cite{zhang2018single}, PFPNet~\cite{kim2018parallel}, RFBNet~\cite{Liu_2018_ECCV}, state-of-the-art two-stage detectors such as Deformable R-FCN~\cite{dai2017deformable}, Mask-RCNN~\cite{he2017mask}, Libra-RCNN~\cite{pang2019libra}, Cascade-RCNN~\cite{cai2018cascade}, SNIP~\cite{singh2018snip}, TridentNet~\cite{li2019scale} and recent anti-imbalance methods - GHM~\cite{li2018gradient} and DR-loss~\cite{qian2019dr} (some of these results are included in Supplementary). After that, \autoref{fig-result2} illustrates some detection results of our detector on COCO dataset. In the testing phase, our detector has the same detection speed (\textit{i.e.}, $\sim$$11~fps$ on one NVidia TitanX GPU) as RetinaNet500~\cite{lin2018focal} since it does not change the network architecture for inference.

\subsubsection{PASCAL VOC}
The experimental results on PASCAL VOC2007 and VOC2012 are shown in \autoref{state-of-the-art-voc07} and \autoref{state-of-the-art-voc12} respectively. It can be seen that our detector surpasses all other methods for both single-scale and multi-scale testing in both two benchmarks. Compared to the closest competitor, PFPNet-R512~\cite{kim2018parallel}, our detector achieves a 1.6\% improvement (83.9\% \textit{vs.} 82.3\%) on VOC2007 dataset, and a 2.8\% improvement (83.1\% \textit{vs.} 80.3\%) on VOC2012 dataset. With the multi-scale testing, our detector achieves a 0.8\% improvement (84.9\% \textit{vs.} 84.1\%) on VOC2007 dataset, a 0.8\% improvement (84.5\% \textit{vs.} 83.7\%) on VOC2012 dataset. Note that both RefineDet~\cite{zhang2018single} and PFPNet~\cite{kim2018parallel} have more advanced detection pipeline than our baseline model RetinaNet~\cite{lin2018focal}. This implies such great improvements are not only from the stronger backbone model (ResNet-101 \textit{vs.} VGG-16), but also from the proposed AP-loss training method. To better understand where the performance gain comes from, we also evaluate the ResNet-101 model with different training losses such as OHEM, Focal Loss and AUC-Loss. Compared to the closest competitor Focal Loss, our AP-loss achieves a 0.9\% improvement (83.9\% \textit{v.s.} 83.0\%) on VOC2007 and a 0.8\% improvement (83.1\% \textit{v.s.} 82.3\%) on VOC2012. More detailed results are shown in Supplementary.

\subsubsection{MS COCO}
The experimental results on MS COCO are shown in \autoref{state-of-the-art-coco}. Our detector outperforms all other methods for both single-scale and multi-scale testing. Compared to the baseline model RetinaNet500~\cite{lin2018focal}, our detector achieves a 3.0\% improvement (37.4\% \textit{vs.} 34.4\%) on COCO dataset. Compared to the closest competitor RefineDet512~\cite{zhang2018single}, our detector achieves a 1.0\% improvement (37.4\% \textit{vs.} 36.4\%) with single-scale testing, and a 0.3\% improvement (42.1\% \textit{vs.} 41.8\%) with multi-scale testing. In 800 pixel resolution, our detector achieves a 1.7\% improvement over the baseline model RetinaNet, and a 0.2\% improvement over the SOTA anti-imbalance method DR-loss. Besides, we also evaluate the ResNet-101 model with other training losses such as OHEM, Focal Loss and AUC-Loss. Compared to the closest competitor OHEM, our AP-loss achieves a 1.0\% improvement (37.4\% \textit{v.s.} 36.4\%) on COCO dataset. More detailed results are shown in Supplementary.

At this juncture, we strongly emphasize that this verifies the effectiveness of our AP-loss since a significant gain in performance can be accomplished by simply replacing the Focal-loss with the AP-loss without whistle and bells, without relying on advanced techniques such as deformable convolution~\cite{dai2017deformable}, SNIP~\cite{singh2018snip}, group normalization~\cite{wu2018group}, etc. We conjecture that the detector performance could be further improved with these additional techniques.

%-------------------------------------------------------------------------
\section{Conclusion}
In this paper, we address the class imbalance issue in one-stage object detectors by replacing the classification sub-task with a ranking sub-task, and proposing to solve the ranking task with AP-Loss. Due to non-differentiability and non-convexity of the AP-loss, we propose a novel algorithm to optimize it based on error-driven update scheme from perceptron learning. We provide a grounded theoretical analysis of the proposed optimization algorithm. Experimental results show that our approach can significantly improve the state-of-the-art one-stage detectors.

%-------------------------------------------------------------------------
{
\small
\bibliographystyle{IEEEtran}
\bibliography{egbib}

\begin{thebibliography}{10}
\providecommand{\url}[1]{#1}
\csname url@samestyle\endcsname
\providecommand{\newblock}{\relax}
\providecommand{\bibinfo}[2]{#2}
\providecommand{\BIBentrySTDinterwordspacing}{\spaceskip=0pt\relax}
\providecommand{\BIBentryALTinterwordstretchfactor}{4}
\providecommand{\BIBentryALTinterwordspacing}{\spaceskip=\fontdimen2\font plus
\BIBentryALTinterwordstretchfactor\fontdimen3\font minus
  \fontdimen4\font\relax}
\providecommand{\BIBforeignlanguage}[2]{{%
\expandafter\ifx\csname l@#1\endcsname\relax
\typeout{** WARNING: IEEEtran.bst: No hyphenation pattern has been}%
\typeout{** loaded for the language `#1'. Using the pattern for}%
\typeout{** the default language instead.}%
\else
\language=\csname l@#1\endcsname
\fi
#2}}
\providecommand{\BIBdecl}{\relax}
\BIBdecl

\bibitem{ren2015faster}
S.~Ren, K.~He, R.~Girshick, and J.~Sun, ``Faster {R-CNN}: Towards real-time
  object detection with region proposal networks,'' in \emph{Advances in Neural
  Information Processing Systems (NeurIPS)}, 2015, pp. 91--99.

\bibitem{liu2016ssd}
W.~Liu, D.~Anguelov, D.~Erhan, C.~Szegedy, S.~Reed, C.-Y. Fu, and A.~C. Berg,
  ``{SSD}: Single shot multibox detector,'' in \emph{European Conference on
  Computer Vision (ECCV)}, 2016, pp. 21--37.

\bibitem{redmon2016you}
J.~Redmon, S.~Divvala, R.~Girshick, and A.~Farhadi, ``You only look once:
  Unified, real-time object detection,'' in \emph{IEEE Conference on Computer
  Vision and Pattern Recognition (CVPR)}, 2016, pp. 779--788.

\bibitem{he2017mask}
K.~He, G.~Gkioxari, P.~Doll{\'a}r, and R.~Girshick, ``Mask r-cnn,'' in
  \emph{International Conference on Computer Vision (ICCV)}, 2017, pp.
  2961--2969.

\bibitem{cai2018cascade}
Z.~Cai and N.~Vasconcelos, ``Cascade r-cnn: Delving into high quality object
  detection,'' in \emph{IEEE Conference on Computer Vision and Pattern
  Recognition (CVPR)}, 2018, pp. 6154--6162.

\bibitem{lin2018focal}
T.-Y. Lin, P.~Goyal, R.~Girshick, K.~He, and P.~Doll{\'a}r, ``Focal loss for
  dense object detection,'' in \emph{International Conference on Computer
  Vision (ICCV)}, 2017, pp. 2980--2988.

\bibitem{girshick2015fast}
R.~Girshick, ``Fast {R-CNN},'' in \emph{International Conference on Computer
  Vision (ICCV)}, 2015, pp. 1440--1448.

\bibitem{uijlings2013selective}
J.~R. Uijlings, K.~E. Van De~Sande, T.~Gevers, and A.~W. Smeulders, ``Selective
  search for object recognition,'' \emph{International Journal of Computer
  Vision}, vol. 104, no.~2, pp. 154--171, 2013.

\bibitem{carreira2012cpmc}
J.~Carreira and C.~Sminchisescu, ``Cpmc: Automatic object segmentation using
  constrained parametric min-cuts,'' \emph{IEEE Trans. Pattern Analysis and
  Machine Intelligence}, vol.~34, no.~7, pp. 1312--1328, 2012.

\bibitem{arbelaez2014multiscale}
P.~Arbel{\'a}ez, J.~Pont-Tuset, J.~T. Barron, F.~Marques, and J.~Malik,
  ``Multiscale combinatorial grouping,'' in \emph{IEEE Conference on Computer
  Vision and Pattern Recognition (CVPR)}, 2014, pp. 328--335.

\bibitem{girshick2014rich}
R.~Girshick, J.~Donahue, T.~Darrell, and J.~Malik, ``Rich feature hierarchies
  for accurate object detection and semantic segmentation,'' in \emph{IEEE
  Conference on Computer Vision and Pattern Recognition (CVPR)}, 2014, pp.
  580--587.

\bibitem{dai2016r}
J.~Dai, Y.~Li, K.~He, and J.~Sun, ``R-{FCN}: Object detection via region-based
  fully convolutional networks,'' in \emph{Advances in Neural Information
  Processing Systems (NeurIPS)}, 2016, pp. 379--387.

\bibitem{lin2017feature}
T.-Y. Lin, P.~Doll{\'a}r, R.~Girshick, K.~He, B.~Hariharan, and S.~Belongie,
  ``Feature pyramid networks for object detection,'' in \emph{IEEE Conference
  on Computer Vision and Pattern Recognition (CVPR)}, 2017, pp. 2117--2125.

\bibitem{redmon2017yolo9000}
J.~Redmon and A.~Farhadi, ``{YOLO}9000: better, faster, stronger,'' in
  \emph{IEEE Conference on Computer Vision and Pattern Recognition (CVPR)},
  2017, pp. 7263--7271.

\bibitem{redmon2018yolov3}
------, ``{YOLO}v3: An incremental improvement,'' \emph{arXiv preprint
  arXiv:1804.02767}, 2018.

\bibitem{fu2017dssd}
C.-Y. Fu, W.~Liu, A.~Ranga, A.~Tyagi, and A.~C. Berg, ``{DSSD}: Deconvolutional
  single shot detector,'' \emph{arXiv preprint arXiv:1701.06659}, 2017.

\bibitem{shen2017dsod}
Z.~Shen, Z.~Liu, J.~Li, Y.-G. Jiang, Y.~Chen, and X.~Xue, ``{DSOD}: Learning
  deeply supervised object detectors from scratch,'' in \emph{International
  Conference on Computer Vision (ICCV)}, 2017, pp. 1919--1927.

\bibitem{zhang2018single}
S.~Zhang, L.~Wen, X.~Bian, Z.~Lei, and S.~Z. Li, ``Single-shot refinement
  neural network for object detection,'' in \emph{IEEE Conference on Computer
  Vision and Pattern Recognition (CVPR)}, 2018, pp. 4203--4212.

\bibitem{law2018cornernet}
H.~Law and J.~Deng, ``Cornernet: Detecting objects as paired keypoints,'' in
  \emph{European Conference on Computer Vision (ECCV)}, 2018, pp. 734--750.

\bibitem{zhao2018m2det}
Q.~Zhao, T.~Sheng, Y.~Wang, Z.~Tang, Y.~Chen, L.~Cai, and H.~Ling, ``M2det: A
  single-shot object detector based on multi-level feature pyramid network,''
  in \emph{AAAI Conference on Artificial Intelligence (AAAI)}, vol.~33, 2019,
  pp. 9259--9266.

\bibitem{shrivastava2016ohem}
A.~Shrivastava, A.~Gupta, and R.~Girshick, ``Training region-based object
  detectors with online hard example mining,'' in \emph{IEEE Conference on
  Computer Vision and Pattern Recognition (CVPR)}, 2016, pp. 761--769.

\bibitem{li2018gradient}
B.~Li, Y.~Liu, and X.~Wang, ``Gradient harmonized single-stage detector,'' in
  \emph{AAAI Conference on Artificial Intelligence (AAAI)}, 2019, pp.
  8577--8584.

\bibitem{cortes2004auc}
C.~Cortes and M.~Mohri, ``{AUC} optimization vs. error rate minimization,'' in
  \emph{Advances in Neural Information Processing Systems (NeurIPS)}, 2004, pp.
  313--320.

\bibitem{cruz2016tackling}
R.~Cruz, K.~Fernandes, J.~S. Cardoso, and J.~F.~P. Costa, ``Tackling class
  imbalance with ranking,'' in \emph{International Joint Conference on Neural
  Networks (IJCNN)}, 2016, pp. 2182--2187.

\bibitem{cruz2017combining}
R.~Cruz, K.~Fernandes, J.~F.~P. Costa, M.~P. Ortiz, and J.~S. Cardoso,
  ``Combining ranking with traditional methods for ordinal class imbalance,''
  in \emph{International Work-conference on Artificial Neural Networks
  (IWANN)}, 2017, pp. 538--548.

\bibitem{natole2019stochastic}
M.~A. Natole, Y.~Ying, and S.~Lyu, ``Stochastic {AUC} optimization algorithms
  with linear convergence,'' \emph{Frontiers in Applied Mathematics and
  Statistics}, vol.~5, p.~30, 2019.

\bibitem{everingham2015pascal}
M.~Everingham, S.~A. Eslami, L.~Van~Gool, C.~K. Williams, J.~Winn, and
  A.~Zisserman, ``The pascal visual object classes challenge: A
  retrospective,'' \emph{International Journal of Computer Vision}, vol. 111,
  no.~1, pp. 98--136, 2015.

\bibitem{lin2014microsoft}
T.-Y. Lin, M.~Maire, S.~Belongie, J.~Hays, P.~Perona, D.~Ramanan,
  P.~Doll{\'a}r, and C.~L. Zitnick, ``Microsoft {COCO}: Common objects in
  context,'' in \emph{European Conference on Computer Vision (ECCV)}, 2014, pp.
  740--755.

\bibitem{salton1986introduction}
G.~Salton and M.~J. McGill, \emph{Introduction to modern information
  retrieval}.\hskip 1em plus 0.5em minus 0.4em\relax McGraw-Hill, Inc., 1986.

\bibitem{song2016training}
Y.~Song, A.~Schwing, R.~Urtasun \emph{et~al.}, ``Training deep neural networks
  via direct loss minimization,'' in \emph{International Conference on Machine
  Learning (ICML)}, 2016, pp. 2169--2177.

\bibitem{yue2007support}
Y.~Yue, T.~Finley, F.~Radlinski, and T.~Joachims, ``A support vector method for
  optimizing average precision,'' in \emph{International ACM SIGIR Conference
  on Research and Development in Information Retrieval (SIGIR)}, 2007, pp.
  271--278.

\bibitem{mohapatra2014efficient}
P.~Mohapatra, C.~Jawahar, and M.~P. Kumar, ``Efficient optimization for average
  precision {SVM},'' in \emph{Advances in Neural Information Processing Systems
  (NeurIPS)}, 2014, pp. 2312--2320.

\bibitem{tsochantaridis2005large}
I.~Tsochantaridis, T.~Joachims, T.~Hofmann, and Y.~Altun, ``Large margin
  methods for structured and interdependent output variables,'' \emph{Journal
  of Machine Learning Research}, vol.~6, no. Sep, pp. 1453--1484, 2005.

\bibitem{Mohapatra_2018_CVPR}
P.~Mohapatra, M.~Rolinek, C.~Jawahar, V.~Kolmogorov, and M.~Pawan~Kumar,
  ``Efficient optimization for rank-based loss functions,'' in \emph{IEEE
  Conference on Computer Vision and Pattern Recognition (CVPR)}, 2018, pp.
  3693--3701.

\bibitem{henderson2016end}
P.~Henderson and V.~Ferrari, ``End-to-end training of object class detectors
  for mean average precision,'' in \emph{Asian Conference on Computer Vision
  (ACCV)}, 2016, pp. 198--213.

\bibitem{chen2019towards}
K.~Chen, J.~Li, W.~Lin, J.~See, J.~Wang, L.~Duan, Z.~Chen, C.~He, and J.~Zou,
  ``Towards accurate one-stage object detection with {AP-loss},'' in \emph{IEEE
  Conference on Computer Vision and Pattern Recognition (CVPR)}, 2019, pp.
  5119--5127.

\bibitem{rosenblatt1957perceptron}
F.~Rosenblatt, \emph{The perceptron, a perceiving and recognizing automaton
  Project Para}.\hskip 1em plus 0.5em minus 0.4em\relax Cornell Aeronautical
  Laboratory, 1957.

\bibitem{sermanet2013overfeat}
P.~Sermanet, D.~Eigen, X.~Zhang, M.~Mathieu, R.~Fergus, and Y.~LeCun,
  ``Overfeat: Integrated recognition, localization and detection using
  convolutional networks,'' \emph{arXiv preprint arXiv:1312.6229}, 2013.

\bibitem{li2018tiny}
Y.~Li, J.~Li, W.~Lin, and J.~Li, ``{Tiny-DSOD}: Lightweight object detection
  for resource-restricted usages,'' in \emph{British Machine Vision Conference
  (BMVC)}, 2018.

\bibitem{Zhang_2018_CVPR}
Z.~Zhang, S.~Qiao, C.~Xie, W.~Shen, B.~Wang, and A.~L. Yuille, ``Single-shot
  object detection with enriched semantics,'' in \emph{IEEE Conference on
  Computer Vision and Pattern Recognition (CVPR)}, 2018, pp. 5813--5821.

\bibitem{kim2018parallel}
S.-W. Kim, H.-K. Kook, J.-Y. Sun, M.-C. Kang, and S.-J. Ko, ``Parallel feature
  pyramid network for object detection,'' in \emph{European Conference on
  Computer Vision (ECCV)}, 2018, pp. 234--250.

\bibitem{Liu_2018_ECCV}
S.~Liu, D.~Huang \emph{et~al.}, ``Receptive field block net for accurate and
  fast object detection,'' in \emph{European Conference on Computer Vision
  (ECCV)}, 2018, pp. 385--400.

\bibitem{wang2017fast}
X.~Wang, A.~Shrivastava, and A.~Gupta, ``A-fast-rcnn: Hard positive generation
  via adversary for object detection,'' in \emph{IEEE Conference on Computer
  Vision and Pattern Recognition (CVPR)}, 2017, pp. 2606--2615.

\bibitem{pang2019libra}
J.~Pang, K.~Chen, J.~Shi, H.~Feng, W.~Ouyang, and D.~Lin, ``Libra {R-CNN}:
  Towards balanced learning for object detection,'' in \emph{IEEE Conference on
  Computer Vision and Pattern Recognition (CVPR)}, 2019, pp. 821--830.

\bibitem{qian2019dr}
Q.~Qian, L.~Chen, H.~Li, and R.~Jin, ``{DR Loss}: improving object detection by
  distributional ranking,'' \emph{arXiv preprint arXiv:1907.10156}, 2019.

\bibitem{oksuz2019imbalance}
K.~Oksuz, B.~C. Cam, S.~Kalkan, and E.~Akbas, ``Imbalance problems in object
  detection: A review,'' \emph{IEEE Trans. Pattern Analysis and Machine
  Intelligence}, 2020.

\bibitem{rao2018learning}
Y.~Rao, D.~Lin, J.~Lu, and J.~Zhou, ``Learning globally optimized object
  detector via policy gradient,'' in \emph{IEEE Conference on Computer Vision
  and Pattern Recognition (CVPR)}, 2018, pp. 6190--6198.

\bibitem{krauth1987learning}
W.~Krauth and M.~M{\'e}zard, ``Learning algorithms with optimal stability in
  neural networks,'' \emph{Journal of Physics A: Mathematical and General},
  vol.~20, no.~11, p. L745, 1987.

\bibitem{anlauf1989adatron}
J.~Anlauf and M.~Biehl, ``The adatron: an adaptive perceptron algorithm,''
  \emph{Europhysics Letters}, vol.~10, no.~7, p. 687, 1989.

\bibitem{wendemuth1995learning}
A.~Wendemuth, ``Learning the unlearnable,'' \emph{Journal of Physics A:
  Mathematical and General}, vol.~28, no.~18, p. 5423, 1995.

\bibitem{li2013learning}
J.~Li and Y.~Zhang, ``Learning surf cascade for fast and accurate object
  detection,'' in \emph{IEEE Conference on Computer Vision and Pattern
  Recognition (CVPR)}, 2013.

\bibitem{he2016deep}
K.~He, X.~Zhang, S.~Ren, and J.~Sun, ``Deep residual learning for image
  recognition,'' in \emph{IEEE Conference on Computer Vision and Pattern
  Recognition (CVPR)}, 2016, pp. 770--778.

\bibitem{deng2009imagenet}
J.~Deng, W.~Dong, R.~Socher, L.-J. Li, K.~Li, and L.~Fei-Fei, ``Imagenet: A
  large-scale hierarchical image database,'' in \emph{IEEE Conference on
  Computer Vision and Pattern Recognition (CVPR)}, 2009, pp. 248--255.

\bibitem{simonyan2014very}
K.~Simonyan and A.~Zisserman, ``Very deep convolutional networks for
  large-scale image recognition,'' \emph{arXiv preprint arXiv:1409.1556}, 2014.

\bibitem{goyal2017accurate}
P.~Goyal, P.~Doll{\'a}r, R.~Girshick, P.~Noordhuis, L.~Wesolowski, A.~Kyrola,
  A.~Tulloch, Y.~Jia, and K.~He, ``Accurate, large minibatch sgd: Training
  imagenet in 1 hour,'' \emph{arXiv preprint arXiv:1706.02677}, 2017.

\bibitem{moosavi2016deepfool}
S.-M. Moosavi-Dezfooli, A.~Fawzi, and P.~Frossard, ``Deepfool: a simple and
  accurate method to fool deep neural networks,'' in \emph{IEEE Conference on
  Computer Vision and Pattern Recognition (CVPR)}, 2016, pp. 2574--2582.

\bibitem{brown2017adversarial}
T.~B. Brown, D.~Man{\'e}, A.~Roy, M.~Abadi, and J.~Gilmer, ``Adversarial
  patch,'' \emph{arXiv preprint arXiv:1712.09665}, 2017.

\bibitem{li2018auto}
B.~Li, T.~Wu, L.~Zhang, and R.~Chu, ``Auto-context r-cnn,'' \emph{arXiv
  preprint arXiv:1807.02842}, 2018.

\bibitem{bell2016inside}
S.~Bell, C.~Lawrence~Zitnick, K.~Bala, and R.~Girshick, ``Inside-outside net:
  Detecting objects in context with skip pooling and recurrent neural
  networks,'' in \emph{IEEE Conference on Computer Vision and Pattern
  Recognition (CVPR)}, 2016, pp. 2874--2883.

\bibitem{gidaris2015object}
S.~Gidaris and N.~Komodakis, ``Object detection via a multi-region and semantic
  segmentation-aware cnn model,'' in \emph{International Conference on Computer
  Vision (ICCV)}, 2015, pp. 1134--1142.

\bibitem{zhu2017couplenet}
Y.~Zhu, C.~Zhao, J.~Wang, X.~Zhao, Y.~Wu, and H.~Lu, ``Couplenet: Coupling
  global structure with local parts for object detection,'' in
  \emph{International Conference on Computer Vision (ICCV)}, 2017, pp.
  4126--4134.

\bibitem{cheng2018revisiting}
B.~Cheng, Y.~Wei, H.~Shi, R.~Feris, J.~Xiong, and T.~Huang, ``Revisiting
  {RCNN}: On awakening the classification power of {Faster} {RCNN},'' in
  \emph{European Conference on Computer Vision (ECCV)}, 2018, pp. 453--468.

\bibitem{dai2017deformable}
J.~Dai, H.~Qi, Y.~Xiong, Y.~Li, G.~Zhang, H.~Hu, and Y.~Wei, ``Deformable
  convolutional networks,'' in \emph{International Conference on Computer
  Vision (ICCV)}, 2017, pp. 764--773.

\bibitem{lu2019grid}
X.~Lu, B.~Li, Y.~Yue, Q.~Li, and J.~Yan, ``Grid r-cnn,'' in \emph{IEEE
  Conference on Computer Vision and Pattern Recognition (CVPR)}, 2019, pp.
  7363--7372.

\bibitem{singh2018snip}
B.~Singh and L.~S. Davis, ``An analysis of scale invariance in object detection
  snip,'' in \emph{IEEE Conference on Computer Vision and Pattern Recognition
  (CVPR)}, 2018, pp. 3578--3587.

\bibitem{li2019scale}
Y.~Li, Y.~Chen, N.~Wang, and Z.~Zhang, ``Scale-aware trident networks for
  object detection,'' in \emph{International Conference on Computer Vision
  (ICCV)}, 2019, pp. 6054--6063.

\bibitem{wu2018group}
Y.~Wu and K.~He, ``Group normalization,'' in \emph{European Conference on
  Computer Vision (ECCV)}, 2018.

\bibitem{Novikoff1963ON}
A.~Novikoff, ``On convergence proofs for perceptrons,''
  \emph{Proc.sympos.math.theory of Automata}, pp. 615--622, 1963.

\bibitem{shalev2012online}
S.~Shalev-Shwartz \emph{et~al.}, ``Online learning and online convex
  optimization,'' \emph{Foundations and Trends{\textregistered} in Machine
  Learning}, vol.~4, no.~2, pp. 107--194, 2012.

\bibitem{kong2017ron}
T.~Kong, F.~Sun, A.~Yao, H.~Liu, M.~Lu, and Y.~Chen, ``Ron: Reverse connection
  with objectness prior networks for object detection,'' in \emph{IEEE
  Conference on Computer Vision and Pattern Recognition (CVPR)}, 2017, pp.
  5936--5944.

\bibitem{wu2019detectron2}
Y.~Wu, A.~Kirillov, F.~Massa, W.-Y. Lo, and R.~Girshick, ``Detectron2,''
  \url{https://github.com/facebookresearch/detectron2}, 2019.

\end{thebibliography}
}

\vspace{-15mm}
\begin{IEEEbiography}[{\includegraphics[width=1in,height=1.2in,clip,keepaspectratio]{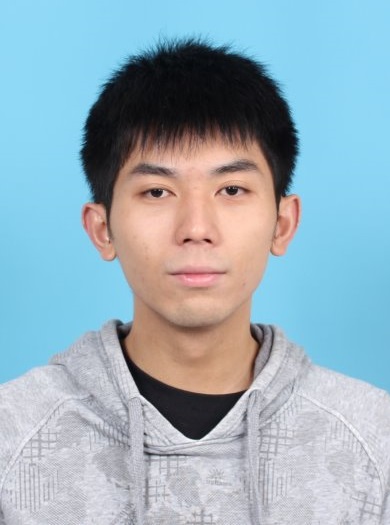}}]{Kean Chen}
received his B.E. degree of Information Engineering from Shanghai Jiao Tong University, China, in 2017. He is currently a master student in Department of Electronic Engineering in Shanghai Jiao Tong University. His current research interests include computer vision and machine learning.
\end{IEEEbiography}

\vspace{-10mm}
\begin{IEEEbiography}[{\includegraphics[width=1in,height=1.2in,clip,keepaspectratio]{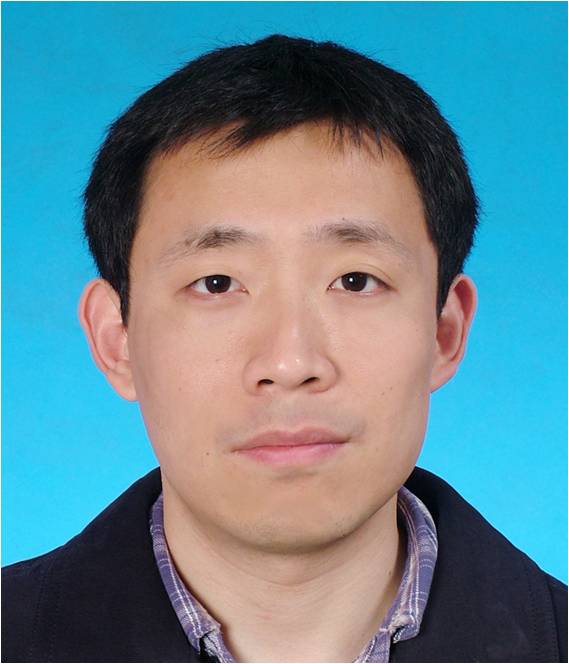}}]{Weiyao Lin}
received the B.E. and M.E. degrees from Shanghai Jiao Tong University, in 2003 and 2005 respectively, and the Ph.D degree from the University of Washington, Seattle, USA, in 2010, all in electrical engineering. He is currently a Professor with the Department of Electronic Engineering, Shanghai Jiao Tong University, China. He has authored or coauthored 100+ technical papers on top journals/conferences including TPAMI, IJCV, TIP, CVPR, and ICCV. He holds 20+ patents. His research interests include video/image analysis, computer vision, and video/image processing applications.

Dr. Lin served as an associate editor for IEEE Trans. Image Processing, IEEE Trans. Circuits \& Systems for Video Technology, and IEEE Trans. Intelligent Transportation Systems. He was an area chair of BMVC'2019, ICIP'2019, ACM MM'2018, ICME'2018-19, and VCIP'2017, and an organizer of 6+ workshops in ICME, ICCV, ECCV. He is a member of IEEE Multimedia Signal Processing (MMSP TC), a member of the IEEE Visual Signal Processing and Communication Technical Committee (VSPC TC), and a member of the the IEEE Multimedia Communications Technical Committee (MMTC). He received the ICME'19 Multimedia Rising Star award. He is a senior member of IEEE.
\end{IEEEbiography}

\vspace{-10mm}
\begin{IEEEbiography}[{\includegraphics[width=1in,height=1.2in,clip,keepaspectratio]{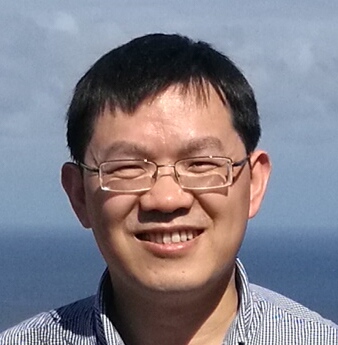}}]{Jianguo Li}
is a senior staff researcher with Intel Labs. His research focuses on computer vision, machine learning and their applications. He has served as TPC members/reviewers for top-tier conferences like CVPR, ICCV, ECCV, AAAI, IJCAI, etc. He is co-organizer of the workshop and challenge on computer vision for wildlife conservation (CVWC) at ICCV 2019. He has published 40+ top-tier peer-reviewed papers, and has led team to win or perform top on several computer vision and machine learning related academic challenges. He has 50+ issued patents, and dozens of technique transfers to Intel products. He got his PhD from Tsinghua University in July 2006. He is a senior member of IEEE.
\end{IEEEbiography}

\vspace{-10mm}
\begin{IEEEbiography}[{\includegraphics[width=1in,height=1.2in,clip,keepaspectratio]{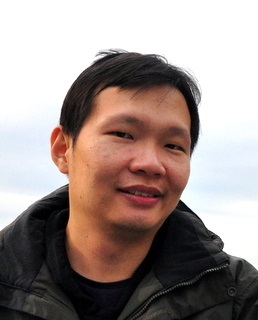}}]{John See}
received his B.Eng., M.Eng.Sc. and Ph.D degrees from Multimedia University, Malaysia. He is currently a Senior Lecturer and Chair of the Centre for Visual Computing. He also leads the Visual Processing Lab at the Faculty of Computing and Informatics, Multimedia University. His current research interests covers a diverse range of topics in computer vision and pattern recognition particularly in visual surveillance, image aesthetics and affective computing. He served as an Associate Editor of EURASIP Journal of Image \& Video Processing and a Program Chair of MMSP 2019. He is a Senior Member of IEEE.
\end{IEEEbiography}

\vspace{-10mm}
\begin{IEEEbiography}[{\includegraphics[width=1in,height=1.2in,clip,keepaspectratio]{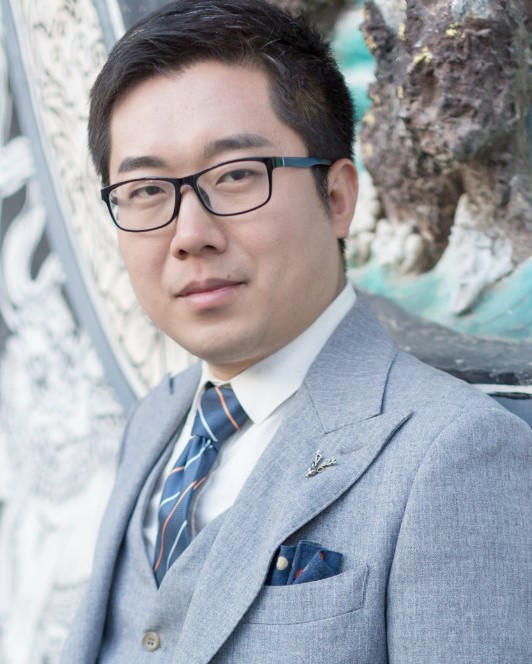}}]{Ji Wang}
received his PhD degree in applied computer science from East China University of Science \& Technology, China, in 2015. Since 2017, he is a researcher at Tencent Youtu Labs, Shanghai. His research interests include  face recognition and object detection.
\end{IEEEbiography}

\vspace{-10mm}
\begin{IEEEbiography}[{\includegraphics[width=1in,height=1.2in,clip,keepaspectratio]{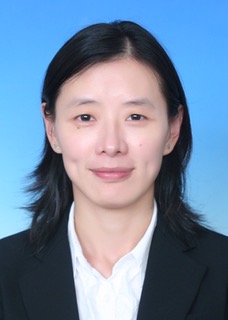}}]{Junni Zou} received the M.S. and the Ph.D. degrees in Communication and Information System from Shanghai University, Shanghai, China, in 2004 and 2006, respectively. From March 2006 to January 2017, she was with the School of Communication and Information Engineering, Shanghai University, Shanghai, where she was a full Professor. Since February 2017, she has been a full Professor with the Department of Computer Science and Engineering, Shanghai Jiao Tong University, China. From June 2011 to June 2012, she was with the Department of Electrical and Computer Engineering, University of California, San Diego (UCSD), as a visiting Professor.

Her research interests include multimedia networking and communications, distributed resource allocation, and machine learning. She has published over 90 international journal/conference papers on these topics. Dr. Zou received the Best 10\% Paper Award in IEEE VCIP 2016. She was granted the National Science Fund for Outstanding Young Scholar in 2016. She was the recipient of Shanghai Young Rising-Star Scientist in 2011. Also, she has served as the Associate Editor for Digital Signal Processing Since April 2019.
\end{IEEEbiography}

\vspace{70mm}

\section*{A1. Convergence Proof}
We provide proof for the proposition mentioned in Section 3.3.1 of the paper. The proof is generalized from the original convergence proof~\cite{Novikoff1963ON} for perceptron learning algorithm.
\begin{proposition}
The AP-loss optimizing algorithm is guaranteed to converge in finite steps if below conditions hold: \par
{\noindent (1) the learning model is linear;} \par
{\noindent (2) the training data is linearly separable.}
\end{proposition}

Let \(\bm{\theta}\) denote the weights of the linear model. Let \(\bm{f}_{k}^{(n)}\) denote the feature vector of \(k\)-th box in \(n\)-th training sample. Assume the number of training samples is finite and each training sample contains at most \(M\) boxes. Hence the score of \(k\)-th box is \(s_{k}^{(n)}=\langle \bm{f}_{k}^{(n)} , \bm{\theta} \rangle\). Define \(x_{ij}^{(n)}=-(s_{i}^{(n)}-s_{j}^{(n)})\). Note that the training data is separable, which means there are \(\epsilon>0\) and \(\bm{\theta}^{*}\) that satisfy:
\begin{equation}
\small
\forall n, \,\, \forall i\in \mathcal{P}^{(n)}, \,\, \forall j\in \mathcal{N}^{(n)}, \,\, \langle \bm{f}_{i}^{(n)} , \bm{\theta}^{*} \rangle \geq \langle \bm{f}_{j}^{(n)}, \bm{\theta}^{*}\rangle + \epsilon
\end{equation}

In the \(t\)-th step, a training sample which makes an error (if there is no such training sample, the model is already optimal and algorithm will stop) is randomly chosen. Then the update of \(\bm{\theta}\) is:
\begin{equation}
\small
\bm{\theta}^{(t+1)}=\bm{\theta}^{(t)}+\sum_{i\in \mathcal{P}}\sum_{j\in \mathcal{N}}L_{ij}(\bm{x})\cdot (\bm{f}_i-\bm{f}_j)
\label{update}
\end{equation}
where
\begin{equation}
\small
L_{ij}(\bm{x})=\frac{H(x_{ij})}{1+\sum_{k\neq i} H(x_{ik})}
\end{equation}
Here, since the discussion centers on the current training sample, we omit the superscript hereon.

From (\ref{update}), we have
\begin{equation}
\small
\begin{split}
\langle \bm{\theta}^{(t+1)}, \bm{\theta}^{*} \rangle &=\langle \bm{\theta}^{(t)},\bm{\theta}^{*}\rangle+\sum_{i\in \mathcal{P}}\sum_{j\in \mathcal{N}}L_{ij}\langle (\bm{f}_i-\bm{f}_j),\bm{\theta}^{*} \rangle \\
&\geq \langle \bm{\theta}^{(t)},\bm{\theta}^{*}\rangle+\sum_{i\in \mathcal{P}}\sum_{j\in \mathcal{N}}L_{ij}\epsilon \\
&\geq \langle \bm{\theta}^{(t)},\bm{\theta}^{*}\rangle+\max_{i\in \mathcal{P},j\in \mathcal{N}}\{L_{ij}\}\epsilon \\
&\geq \langle \bm{\theta}^{(t)},\bm{\theta}^{*} \rangle + \frac{1}{|\mathcal{P}|+|\mathcal{N}|}\epsilon \\
&\geq \langle \bm{\theta}^{(t)},\bm{\theta}^{*} \rangle + \frac{1}{M}\epsilon
\end{split}
\end{equation}
For convenience, let \(\bm{\theta}^{(0)}=0\) (if \(\bm{\theta}^{(0)}\neq0\), we can still find a \(c>0\) that satisfies (\ref{lowerbound}) for sufficiently large \(t\)), we have
\begin{equation}
\small
\langle \bm{\theta}^{(t)},\bm{\theta}^{*} \rangle \geq \frac{1}{M}\epsilon \cdot t
\end{equation}
Then
\begin{equation}
\small
\|\bm{\theta}^{(t)}\| \geq \frac{\langle \bm{\theta}^{(t)},\bm{\theta}^{*} \rangle}{\|\bm{\theta}^{*}\|} \geq \frac{1}{M\cdot \|\bm{\theta}^{*}\|}\epsilon \cdot t \geq c \cdot t
\label{lowerbound}
\end{equation}
Here, \(c\) is a positive constant.

From (\ref{update}), we also have
\begin{equation}
\small
\begin{split}
&\|\bm{\theta}^{(t+1)}\|^2 \\
= & \|\bm{\theta}^{(t)}\|^2+\|\sum_{i\in \mathcal{P}}\sum_{j\in \mathcal{N}}L_{ij}(\bm{f}_i-\bm{f}_j)\|^2 \\
& \,\,\,\,\,\,\,\,\,\,\,\,\,\,\,\,\,\,\,\,\,\,\,\,\,\,\,\,\,\,\,\,\,\,\,\,\,\,\,\,\,\,\,\,\,\,\, +2\langle \sum_{i\in \mathcal{P}}\sum_{j\in \mathcal{N}}L_{ij}(\bm{f}_i-\bm{f}_j),\bm{\theta}^{(t)} \rangle \\
= & \|\bm{\theta}^{(t)}\|^2 + \|\sum_{i\in \mathcal{P}}\sum_{j\in \mathcal{N}}L_{ij}(\bm{f}_i-\bm{f}_j)\|^2+2\sum_{i\in \mathcal{P}}\sum_{j\in \mathcal{N}}L_{ij} x_{ji} \\
\leq & \|\bm{\theta}^{(t)}\|^2 + \|\sum_{i \in \mathcal{P}}\sum_{j\in \mathcal{N}} L_{ij}(\bm{f}_i-\bm{f}_j)\|^2 \\
\leq & \|\bm{\theta}^{(t)}\|^2 + |\mathcal{P}|\cdot |\mathcal{N}| \cdot \max_{i\in \mathcal{P},j\in \mathcal{N}}\{\|\bm{f}_i-\bm{f}_j\|^2\} \\
\leq & \|\bm{\theta}^{(t)}\|^2 + M^2\cdot \max_{n,i,j}\{\|\bm{f}^{(n)}_i-\bm{f}^{(n)}_j\|^2\} \\
\leq & \|\bm{\theta}^{(t)}\|^2 + C
\end{split}
\end{equation}
Here, \(C\) is a positive constant. Let \(\bm{\theta}^{(0)}=0\) (again, if \(\bm{\theta}^{(0)}\neq0\), we can still find a \(C>0\) that satisfies (\ref{upperbound}) for sufficiently large \(t\)), we arrive at:
\begin{equation}
\small
\|\bm{\theta}^{(t)}\|^2 \leq C \cdot t
\label{upperbound}
\end{equation}
Then, combining (\ref{lowerbound}) and (\ref{upperbound}), we have
\begin{equation}
\small
c^2\cdot t^2\leq \|\bm{\theta}^{(t)}\|^2 \leq C\cdot t
\end{equation}
which means
\begin{equation}
\small
t \leq \frac{C}{c^2}
\end{equation}
It shows that the algorithm will stop at most after \(C/c^2\) steps, which means that the training model will achieve the optimal solution in finite steps.

\section*{A2. An Example of Gradient Descent Failing on Smoothed AP-loss}

We approximate the step function in AP-loss by sigmoid function to make it amenable to gradient descent. Specifically, the smoothed AP-loss function is given by:
\begin{equation}
\small
F =\frac{1}{|\mathcal{P}|}\sum_{i\in \mathcal{P}}\sum_{j\in \mathcal{N}}\frac{S(x_{ij})}{1+\sum_{k\neq i}S(x_{ik})}
\end{equation}
where
\begin{equation}
\small
S(x)=\frac{e^x}{1+e^x}
\end{equation}
Consider a linear model \(s=f_1 \theta_1+f_2 \theta_2\) and three training samples \((0,0),(1,0),(-3,1)\) (the first one is negative sample, others are positive samples). Then we have
\begin{equation}
\small
\begin{split}
&s^{(1)}=0\cdot \theta_1+0 \cdot \theta_2 \\
&s^{(2)}=1\cdot \theta_1+0\cdot \theta_2 \\
&s^{(3)}=-3\cdot \theta_1 + 1 \cdot \theta_2
\end{split}
\end{equation}
Note that the training data is separable since we have \(s^{(2)} > s^{(1)}\) and \(s^{(3)} > s^{(1)}\) when \(0< \theta_1 < \frac{1}{3} \cdot \theta_2 \).

Under this setting, the smoothed AP-loss become
\begin{equation}
\small
\begin{split}
& F(\theta_1,\theta_2)=\frac{1}{2}(\frac{S(-\theta_1)}{1+S(-\theta_1)+S(\theta_2-4\theta_1)} \\
& \,\,\,\,\,\,\,\,\,\,\,\,\,+\frac{S(3\theta_1-\theta_2)}{1+S(4\theta_1-\theta_2)+S(3\theta_1-\theta_2)})
\end{split}
\end{equation}
If \(\theta_1\) is sufficiently large and \(\theta_1>\theta_2>0\), then the partial derivatives satisfy the following condition:
\begin{equation}
\small
\frac{\partial F}{\partial \theta_1}<\frac{\partial F}{\partial \theta_2}<0
\label{partial}
\end{equation}
which means \(\theta_1\) and \(\theta_2\) will keep increasing with the inequality \(\theta_1>\theta_2\) according to the gradient descent algorithm. Hence the objective function \(F\) will approach \(1/6\) here. However, the objective function \(F\) approaches the global minimum value \(0\) if and only if \(\theta_1 \rightarrow +\infty\) and \( \theta_2 - 3 \theta_1 \rightarrow +\infty\). This shows that the gradient descent fails to converge to global minimum in this case.

\section*{A3. Inseparable Case}

In this section, we will provide analysis for our algorithm with inseparable training data. We demonstrate that the bound of accumulated AP-loss depends on the best performance of learning model. The analysis is based on online learning bounds~\cite{shalev2012online}.

\subsection*{A3.1. Preliminary}
To handle the inseparable case, a mild modification on the proposed algorithm is needed, \textit{i.e.} in the error-driven update scheme, \(L_{ij}\) is modified to
\begin{equation}
\small
\widetilde{L}_{ij}=\frac{\widetilde{H}(x_{ij})}{1+\sum_{k\in \mathcal{P}\cup\mathcal{N},k\neq i}H(x_{ik})}
\end{equation}
where \(\widetilde{H}(\cdot)\) is defined in Section 3.4.2 (Piecewise Step Function) of the paper. The purpose is to introduce a non-zero decision margin for the pairwise score \(x_{ij}\) which makes the algorithm more robust in the inseparable case. In contrast to the case in Section 3.4.2, here we only change \(H(\cdot)\) to \(\widetilde{H}(\cdot)\) in the numerator for the convenience of theoretical anaysis. However, such algorithm still suffers from the discontinuity of \(H(\cdot)\) in the denominator. Hence the strategy in Section 3.4.2 is also practical consideration, necessary for good performance. Then, consider the AP-loss:
\begin{equation}
\small
\mathcal{L}_{AP}(\bm{x};\mathcal{P},\mathcal{N})=\frac{1}{|\mathcal{P}|}\sum_{i\in \mathcal{P}}\frac{\sum_{j\in \mathcal{N}} H(x_{ij})}{1+\sum_{j\in \mathcal{P} \cup\mathcal{N},j\neq i}H(x_{ij})}
\end{equation}
and define a surrogate loss function:
\begin{equation}
\small
l(\bm{x},\hat{\bm{x}};\mathcal{P},\mathcal{N})=\frac{1}{|\mathcal{P}|}\sum_{i\in \mathcal{P}}\frac{\sum_{j\in \mathcal{N}}Q(x_{ij})}{1+\sum_{j\in \mathcal{P}\cup\mathcal{N},j \neq i}H(\hat{x}_{ij})}
\end{equation}
where \(Q(x)=\int_{-\infty}^{x}{\widetilde{H}(\upsilon)d\upsilon}\). Note that the AP-loss is upper bounded by the surrogate loss:
\begin{equation}
\small
l(\bm{x},\bm{x};\mathcal{P},\mathcal{N})\geq \frac{\delta}{4} \mathcal{L}_{AP}(\bm{x};\mathcal{P},\mathcal{N})\
\end{equation}
The learning model can be written as \(\bm{x}=\bm{X}_{d}(\bm{\theta})\), where \(d\in \mathcal{D}\) denotes the training data for one iteration and \(D\) is the whole training set. Then, the modified error-driven algorithm is equivalent to gradient descent on surrogate loss \(l(\bm{X}_{d^{(t)}}(\bm{\theta}),\bm{X}_{d^{(t)}}(\bm{\theta}^{(t)});\mathcal{P}_{d^{(t)}},\mathcal{N}_{d^{(t)}})\) at each step \(t\).
We further suppose below conditions are satisfied:\par
\noindent(1) For all \(\hat{\bm{\theta}}\) and \(d \in \mathcal{D}\), \(l(\bm{X}_{d}(\bm{\theta}),\bm{X}_{d}(\hat{\bm{\theta}});\mathcal{P}_d,\mathcal{N}_d)\) is convex \textit{w.r.t} \(\bm{\theta}\).\par
\noindent(2) For all \(d \in \mathcal{D}\), \(\|\partial \bm{X}_{d}(\bm{\theta}) / \partial \bm{\theta}\|\) is upper bounded by a constant \(R\). Here \(\|\cdot\|\) is the matrix norm induced by the 2-norm for vectors.

\noindent \textit{\textbf{Remark 1.}} Note that these two conditions are satisfied if the learning model is linear.\par

\subsection*{A3.2. Bound of Accumulated Loss}

By the convexity, we have:
\begin{equation}
\small
l^{(t)}(\bm{\theta})\leq l^{(t)}(\bm{u})+\langle\bm{\theta}-\bm{u},\frac{\partial l^{(t)}(\bm{\theta})}{\partial \bm{\theta}}\rangle
\end{equation}
where \(l^{(t)}(\bm{\theta})\) denotes \(l(\bm{X}_{d^{(t)}}(\bm{\theta}),\bm{X}_{d^{(t)}}(\bm{\theta}^{(t)});\mathcal{P}_{d^{(t)}},\mathcal{N}_{d^{(t)}})\) and \(\bm{u}\) can be any vector of model weights.
Then, let \(\bm{\theta}=\bm{\theta}^{(t)}\) and compute the sum over \(t=1 \sim T\), we have:
\begin{equation}
\small
\begin{split}
&\sum_{t=1}^{T} l^{(t)}(\bm{\theta}^{(t)}) - \sum_{t=1}^{T}l^{(t)}(\bm{u})
\leq \sum_{t=1}^{T}\langle \bm{\theta}^{(t)}-\bm{u},\frac{\partial l^{(t)}(\bm{\theta}^{(t)})}{\partial \bm{\theta}}\rangle \\
&=\sum_{t=1}^{T} \langle \bm{\theta}^{(t)}-\bm{u},\frac{1}{\eta}(\bm{\theta}^{(t)}-\bm{\theta}^{(t+1)})\rangle \\
&\leq \frac{1}{2\eta} \|\bm{u}-\bm{\theta}^{(1)}\|^2 +
\frac{1}{2\eta} \sum_{t=1}^{T} \|\bm{\theta}^{(t)}-\bm{\theta}^{(t+1)}\|^2\\
&=\frac{1}{2\eta} \|\bm{u}-\bm{\theta}^{(1)}\|^2 + \frac{\eta}{2}\sum_{t=1}^{T}\|\frac{\partial l^{(t)}(\bm{\theta}^{(t)})}{\partial \bm{\theta}}\|^2
\end{split}
\end{equation}
where \(\eta\) is the step size of gradient descent. Note that
\begin{equation}
\small
\frac{\partial l^{(t)}(\bm{\theta}^{(t)})}{\partial \bm{\theta}}=\frac{\partial \bm{X}(\bm{\theta})}{\partial {\bm{\theta}}} \big|_{\bm{\theta}=\bm{\theta}^{(t)}} \cdot
\frac{\partial l(\bm{x},\bm{x}^{(t)})}{\partial \bm{x}} \big|_{\bm{x}=\bm{X}(\bm{\theta}^{(t)})}
\end{equation}
and
\begin{equation}
\small
\begin{split}
&\|\frac{\partial l(\bm{x},\bm{x}^{(t)})}{\partial \bm{x}}\|^{2}
=\frac{1}{|\mathcal{P}|^{2}}\sum_{i\in \mathcal{P}}\frac{\sum_{j\in \mathcal{N}} \widetilde{H}^{2}(x_{ij})}{(1+\sum_{j\in \mathcal{P}\cup\mathcal{N},j\neq i} H(x^{(t)}_{ij}))^{2}}\\
&\leq \frac{1}{|\mathcal{P}|^{2}}\sum_{i\in \mathcal{P}} \frac{\frac{1}{\delta} \sum_{j\in \mathcal{N}} Q(x_{ij})}{1+\sum_{j\in \mathcal{P}\cup\mathcal{N},j\neq i} H(x^{(t)}_{ij})}\leq\frac{1}{\delta } l(\bm{x},\bm{x}^{(t)})\\
\end{split}
\end{equation}
Note that both \(\mathcal{P}_{d}\) and \(\mathcal{N}_{d}\) depend on \(d\). However, we omit the subscript \(d\) here since the discussion only centers on the current training sample \(d^{(t)}\).

Hence we have:
\begin{equation}
\small
\begin{split}
&\sum_{t=1}^{T} l^{(t)}(\bm{\theta}^{(t)}) - \sum_{t=1}^{T}l^{(t)}(\bm{u})\\
&\leq \frac{1}{2\eta} \|\bm{u}-\bm{\theta}^{(1)}\|^2 + \frac{\eta R^2}{2 \delta} \sum_{t=1}^{T} l^{(t)}(\bm{\theta}^{(t)}).
\end{split}
\end{equation}
Let \(\eta=\delta/R^2\), rearrange and get the expression:
\begin{equation}
\small
\frac{1}{2}\sum_{t=1}^{T}l^{(t)}(\bm{\theta}^{(t)})\leq \frac{R^2}{2\delta} \|\bm{u}-\bm{\theta}^{(1)}\|^2 + \sum_{t=1}^{T}l^{(t)}(\bm{u})
\end{equation}
This entails the bound of surrogate loss \(l\):
\begin{equation}
\small
\begin{split}
\sum_{t=1}^{T}l^{(t)}(\bm{\theta}^{(t)})\leq 2\sum_{t=1}^{T}l^{(t)}(\bm{u})+ \frac{R^2}{\delta} \|\bm{u}-\bm{\theta}^{(1)}\|^2
\end{split}
\end{equation}
which implies the bound of AP-loss \(\mathcal{L}_{AP}\):
\begin{equation}
\small
\sum_{t=1}^{T} \mathcal{L}_{AP}(\bm{X}(\bm{\theta}^{(t)})) \leq \frac{8}{\delta}\sum_{t=1}^{T}l^{(t)}(\bm{u})+ \frac{4R^2}{\delta^2} \|\bm{u}-\bm{\theta}^{(1)}\|^2
\label{regret_bound1}
\end{equation}
As a special case, if there exists a \(\bm{u}\) such that \(l^{(t)}(\bm{u})=0\) for all \(t\), then the accumulated AP-loss is bounded by a constant, which implies that convergence can be achieved with finite steps (similar to that of the separable case). Otherwise, with sufficiently large \(T\), the average AP-loss mainly depends on \(\frac{1}{T}\frac{8}{\delta}\sum_{t=1}^{T}l^{(t)}(\bm{u})\). This implies that the bound is meaningful if there still exists a sufficiently good solution \(\bm{u}\) in such inseparable case.

\subsection*{A3.3. Offline Setting}
With the offline setting (\(d^{(t)}=d\) for all \(t\)), a bound with simpler form can be revealed. For simplicity, we will omit the subscript \(d\) of \(\bm{X}_{d}(\bm{u}),\mathcal{P}_d, \mathcal{N}_d\) and define \(A_{i}(\bm{u})=\sum_{j\in\mathcal{N}}Q(X_{ij}(\bm{u}))\), \(Z(\bm{u})=\max_{i\in \mathcal{P}}\{A_{i}(\bm{u})\}\).
Then,
\begin{equation}
\small
\begin{split}
&l^{(t)}(\bm{u})= \frac{1}{|\mathcal{P}|}\sum_{i\in \mathcal{P}}\frac{\sum_{j\in \mathcal{N}}Q(X_{ij}(\bm{u}))}{1+\sum_{j\in \mathcal{P}\cup\mathcal{N},j\neq i}H(X_{ij}(\bm{\theta}^{(t)}))}\\
&= \frac{1}{|\mathcal{P}|}\sum_{i\in \mathcal{P}}\frac{A_i(\bm{u})}{1+\sum_{j\in \mathcal{P}\cup\mathcal{N},j\neq i}H(X_{ij}(\bm{\theta}^{(t)}))}\\
&\leq \frac{Z(\bm{u})}{|\mathcal{P}|}\sum_{i=1}^{|\mathcal{P}|}\frac{1}{i}\leq \frac{\ln|\mathcal{P}|+1}{|\mathcal{P}|}Z(\bm{u})
\end{split}
\label{bound1}
\end{equation}
The second last inequality is based on the fact that \((1+\sum_{j\in \mathcal{P}\cup\mathcal{N},j\neq i}H(X_{ij}(\bm{\theta}^{(t)})))\) are picked from \(1\sim (|\mathcal{P}|+|\mathcal{N}|)\) without replacement (assume no ties; if ties exist, this inequality still holds).
Combining the results from \autoref{bound1} and \autoref{regret_bound1}, we have:
\begin{equation}
\footnotesize
\frac{1}{T}\sum_{t=1}^{T}\mathcal{L}_{AP}(\bm{X}(\bm{\theta}^{(t)}))\leq \frac{\ln|\mathcal{P}|+1}{|\mathcal{P}|}\cdot\frac{8}{\delta}Z(\bm{u})+ \frac{1}{T}\frac{4R^2\|\bm{u}-\bm{\theta}^{(1)}\|^2}{\delta^2}
\label{final_bound1}
\end{equation}
Next,
\begin{equation}
\footnotesize
\begin{split}
&l^{(t)}(\bm{u})=\frac{1}{|\mathcal{P}|}\sum_{i\in \mathcal{P}}\frac{A_i(\bm{u})}{1+\sum_{j\in \mathcal{P}\cup\mathcal{N},j\neq i}H(X_{ij}(\bm{\theta}^{(t)}))} \\
&=\frac{1}{|\mathcal{P}|}\sum_{i\in \mathcal{P}}\frac{1+\sum_{j\in\mathcal{P},j\neq i}H(X_{ij}(\bm{\theta}^{(t)}))}{1+\sum_{j\in \mathcal{P}\cup\mathcal{N},j\neq i}H(X_{ij}(\bm{\theta}^{(t)}))}\\
&\quad\quad\quad\quad\quad\quad\quad\quad\quad\quad\quad \cdot\frac{A_i(\bm{u})}{1+\sum_{j\in\mathcal{P},j\neq i}H(X_{ij}(\bm{\theta}^{(t)}))}\\
&\leq \frac{1}{|\mathcal{P}|}\sum_{i\in \mathcal{P}}\frac{1+\sum_{j\in\mathcal{P},j\neq i}H(X_{ij}(\bm{\theta}^{(t)}))}{1+\sum_{j\in \mathcal{P}\cup\mathcal{N},j\neq i}H(X_{ij}(\bm{\theta}^{(t)}))}\cdot Z(\bm{u})\\
&=(1-\mathcal{L}_{AP}(\bm{X}(\bm{\theta}^{(t)})))\cdot Z(\bm{u})
\end{split}
\label{bound2}
\end{equation}
Combining the results from \autoref{bound2} and \autoref{regret_bound1}, we have:
\begin{equation}
\footnotesize
\frac{1}{T}\sum_{t=1}^{T}\mathcal{L}_{AP}(\bm{X}(\bm{\theta}^{(t)}))\leq \frac{\frac{8}{\delta}Z(\bm{u})}{1+\frac{8}{\delta}Z(\bm{u})}+ \frac{1}{T}\frac{4R^2\|\bm{u}-\bm{\theta}^{(1)}\|^2}{\delta^2}
\label{final_bound2}
\end{equation}
If \(Z(\bm{u})\) is small, the bound in \autoref{final_bound1} is active, otherwise the bound in \autoref{final_bound2} is active. Consequently, we have:
\begin{equation}
\footnotesize
\overline{\mathcal{L}_{AP}}\leq \min\{\frac{\ln|\mathcal{P}|+1}{|\mathcal{P}|}\frac{8}{\delta}Z(\bm{u}),\frac{\frac{8}{\delta}Z(\bm{u})}{1+\frac{8}{\delta}Z(\bm{u})}\}+\epsilon
\end{equation}
where \(\overline{\mathcal{L}_{AP}}\) denotes the average AP-loss, \(\epsilon\rightarrow 0\) as \(T\) increases.

\section*{A4. Additional Experiments}
\subsection*{A4.1. From MS COCO to PASCAL VOC}

\begin{table}[t]
\footnotesize
\centering
\caption{Models are pre-trained on COCO and fine tuned on PASCAL VOC.}
\vspace{-2mm}
\setlength{\tabcolsep}{3.5mm}{
\begin{tabular}{cccc}
\midrule[1pt]
Method & Backbone & VOC07 & VOC12 \\
\midrule[1pt]
Single-Scale: & & & \\
SSD300~\cite{liu2016ssd} & VGG-16 & 81.2 & 79.3 \\
SSD512~\cite{liu2016ssd} & VGG-16 & 83.2 & 82.2 \\
RON384++~\cite{kong2017ron} & VGG-16 & 81.3 & 80.7 \\
DSOD300~\cite{shen2017dsod} & DS/64-192-48-1 & 81.7 & 79.3 \\
RefineDet512~\cite{zhang2018single} & VGG-16 & 85.2 & 85.0 \\
Ours & ResNet-101 & \textbf{86.2} & \textbf{87.2} \\
\midrule[1pt]
Multi-Scale: & & & \\
RefineDet512~\cite{zhang2018single} & VGG-16 & 85.8 & 86.8 \\
Ours & ResNet-101 & \textbf{87.0} & \textbf{88.6} \\
\midrule[1pt]
\end{tabular}}
\label{state-of-the-art-coco-to-voc}
\end{table}

In this section, we investigate how the pre-trained model affects the detection performance. We first train the detector on COCO dataset, and then fine-tune the model on PASCAL VOC dataset. COCO has much more training images and contains all classes in PASCAL VOC. Results are shown in \autoref{state-of-the-art-coco-to-voc}. We observe that our detector outperforms all other methods. Compared to the closest competitor RefineDet512~\cite{zhang2018single}, our detector achieves a 1.0\% improvement (86.2\% \textit{vs.} 85.2\%) on VOC2007 dataset and a 2.2\% improvement (87.2\% \textit{vs.} 85.0\%) on VOC2012. With multi-scale testing, our detector achieves a 1.2\% improvement (87.0\% \textit{vs.} 85.8\%) on VOC2007 dataset and a 1.8\% improvement (88.6\% \textit{vs.} 86.8\%) on VOC2012 dataset.

\subsection*{A4.2. More Ablation Experiments}
\subsubsection*{A4.2.1. OHEM and Random Sampling for AP-loss}
\begin{table}[h]
\small
\centering
\caption{Models are tested on VOC2007 {\tt test} set.}
\vspace{-2mm}
\setlength{\tabcolsep}{2.5mm}{
\begin{tabular}{ccccccc}
\midrule[1pt]
Method & AP & AP\(_{50}\) & AP\(_{75}\) \\
\midrule[1pt]
OHEM + AP-loss & 53.0 & 81.1 & 57.4 \\
Random Sampling + AP-loss & 39.8 & 68.5 & 40.5 \\
Original AP-Loss & \textbf{53.1} & \textbf{82.3} & \textbf{58.1} \\
\midrule[1pt]
\end{tabular}}
\label{OHEM_Randomsampling}
\end{table}
We also conduct an experiment to evaluate the Hard-Example Miming (OHEM) and Random Sampling for AP-loss optimization. We first use OHEM or Random Sampling to select a subset of negative samples, and then use these samples to compute AP-loss. Other experimental settings are the same as those in Section 4.2 of the paper. The experimental results are shown in \autoref{OHEM_Randomsampling}. Note that ``Random Sampling + AP-loss'' is much worse than original AP-loss. This is because there are a large number of easy negative samples, and the random sampling is very likely to only choose these easy negative samples. Then, the training will be sub-optimal since the easy negatives contribute little useful learning signals. The ``OHEM + AP-loss'' is slightly worse than original AP-loss. We argue that the OHEM is redundant for AP-loss, since the AP-loss itself can handle the imbalance issue through a ranking strategy, while the OHEM procedure could in fact, filter some potentially valuable negative samples and thus impact the ranking task.

\subsubsection*{A4.2.2. Parameter Settings}
\noindent\textbf{Batch Size:}
In addition to the AP-loss model, we also evaluate the detectors trained with Focal-Loss on different batch-sizes, and compare the results with those of AP-loss detectors, to demonstrate that the gain from enlarging the batch-size in AP-loss training is non-trivial. The experimental settings are similar to those in Section 4.2.1 of the paper. When varying the batch-size, we follow the linear scaling rule~\cite{goyal2017accurate} to adjust the learning rate. The experimental results are shown in \autoref{focal-batch-size}.
We observe that batch-size 4 is better than batch-size 8 for the Focal-loss model. Thus, we conjecture that generally, larger batch-sizes do not necessarily help the ordinary detection task (note that the BN layers are frozen in our experiments). In contrast, we found that though the ``score-shift'' does not seriously impact the performance, it still has a non-trivial effect on our ranking task, which demands larger batch-sizes for better performance (especially on AP$_{50}$).
\begin{table}[t]
\small
\centering
\caption{Models trained with Focal-Loss and AP-Loss on different batch-sizes. The models are evaluated on VOC2007 {\tt test} dataset.}
\setlength{\tabcolsep}{2mm}{
\begin{tabular}{cccccccccccccccc}
\midrule[1pt]
\multirow{2}*{Batch-size} & \multicolumn{3}{c} {Focal-Loss} & \multicolumn{3}{c}{AP-Loss} \\
\cmidrule{2-4}\cmidrule{5-7}
& AP & AP\(_{50}\) & AP\(_{75}\) & AP & AP\(_{50}\) & AP\(_{75}\)  \\
\midrule[1pt]
1 & 50.6 & 79.8 & 54.5 & 52.4 & 80.2 & 56.7\\
2 & 51.1 & 80.5 & 55.0 & 53.0 & 81.7 & 57.8\\
4 & \textbf{51.5} & \textbf{81.2} & \textbf{55.9} & 52.8 & 82.2 & 58.0\\
8 & 51.3 & 80.9 & 55.3 & \textbf{53.1} & \textbf{82.3} & \textbf{58.1}\\
\midrule[1pt]
\end{tabular}}
\label{focal-batch-size}
\end{table}

\noindent\textbf{Sigmoid Function:}
As an alternative of the piecewise step function, the sigmoid function can also achieve competitive results. In details, we implement the sigmoid function as
\begin{equation}
Sigmoid(x)=\frac{e^{x/k}}{1+e^{x/k}}
\end{equation}
where \(k\) is designated as a tuning hyper-parameter to control the slope near zero point. This is similar to the parameter \(\delta\) in piecewise step function. The experimental settings are the same as those in Section 4.2 of the paper and the results are shown in \autoref{sigmoid}. The Sigmoid function achieves its best performance when \(k=0.5\), and is competitive to the piecewise step function. This verifies our hypothesis that the precise form of piecewise step function is not crucial. The sigmoid function is a good example to replace piecewise step function, and this also demonstrates the robustness of the proposed method.
\begin{table}[t]
\small
\centering
\caption{Sigmoid function with different \(k\). Models are tested on VOC2007 {\tt test} set.}
\vspace{-2mm}
\setlength{\tabcolsep}{2.5mm}{
\begin{tabular}{cccc}
\midrule[1pt]
\(k\) & AP & AP\(_{50}\) & AP\(_{75}\) \\
\midrule[1pt]
0.25 & 52.5 & 81.4 & 57.0 \\
0.5 & \textbf{53.5} & \textbf{81.9} & \textbf{58.4} \\
1 & 51.3 & 80.6 & 55.9 \\
2 & 42.7 & 72.1 & 43.9 \\
\midrule[1pt]
\end{tabular}}
\vspace{-2mm}
\label{sigmoid}
\end{table}

\subsubsection*{A4.2.3. Ablation Experiments on Stronger Baselines}
\begin{table}[t]
\small
\centering
\caption{Ablation experiments on ResNet-101. Models are tested on VOC2007 {\tt test} set, VOC2012 {\tt test} set and COCO {\tt test-dev} set.}
\vspace{-2mm}
\begin{tabular}{cccccc}
\midrule[1pt]
\multirow{2}*{Training Loss} & \multicolumn{1}{c}{VOC07} & \multicolumn{1}{c}{VOC12} & \multicolumn{3}{c}{COCO} \\
\cmidrule{2-3}\cmidrule{4-6}
 & AP\(_{50}\) & AP\(_{50}\) & AP & AP\(_{50}\) & AP\(_{75}\) \\
\midrule[1pt]
CE-Loss+OHEM & 82.8 & 81.5 & 36.4 & 58.0 & 38.8 \\
Focal Loss & 83.0 & 82.3 & 36.2 & 57.5 & 38.7 \\
AUC-Loss & 81.9 & 81.8 & 33.1 & 54.7 & 34.8 \\
AP-Loss & \textbf{83.9} & \textbf{83.1} & \textbf{37.4} & \textbf{58.6} & \textbf{40.5} \\
\midrule[1pt]
\end{tabular}
\vspace{-4mm}
\label{Resnet-101}
\end{table}

\begin{table}[t]
\small
\centering
\caption{Results of different losses on ResNeXt-101-32\(\times\)8d. The models are evaluated on VOC2007 {\tt test} dataset and COCO {\tt test-dev} dataset.}
\setlength{\tabcolsep}{1.5mm}{
\begin{tabular}{cccccccccccccccc}
\midrule[1pt]
\multirow{2}*{Training Loss} & \multicolumn{3}{c} {PASCAL VOC2007} & \multicolumn{3}{c}{COCO {\tt test-dev}} \\
\cmidrule{2-4}\cmidrule{5-7}
& AP & AP\(_{50}\) & AP\(_{75}\) & AP & AP\(_{50}\) & AP\(_{75}\)  \\
\midrule[1pt]
CE-Loss + OHEM & 55.9 & 83.8 & 62.3 & 38.1 & 60.3 & 40.7\\
Focal Loss & 56.8 & 83.6 & 62.2 & 38.5 & 60.5 & 41.4\\
AUC-Loss & 54.0 & 82.6 & 59.0 & 34.5 & 56.2 & 36.6\\
AP-Loss & \textbf{57.5} & \textbf{84.5} & \textbf{63.5} & \textbf{39.2} & \textbf{62.5} & \textbf{41.6}\\
\midrule[1pt]
\end{tabular}}
\label{resnext}
\end{table}

We evaluate the ResNet-101 and ResNeXt-101 models with different training losses such as OHEM, Focal-Loss, AUC-Loss. The experimental settings are the same as those in Section 4.4 of the paper. The results are shown in \autoref{Resnet-101} and \autoref{resnext}. Compared to other methods, the proposed AP-loss achieves consistent improvements on different datasets and backbones, which demonstrates its effectiveness and strong generalization ability.

\begin{table}[t]
\small
\centering
\caption{Ablation experiments for GHM and DR-loss. The models are evaluated on VOC2007 {\tt test} and VOC2012 {\tt test} dataset.}
\vspace{-2mm}
\setlength{\tabcolsep}{1.5mm}{
\begin{tabular}{cccccccccccccccc}
\midrule[1pt]
\multirow{2}*{Method} & \multicolumn{3}{c}{VOC2007} & \multicolumn{1}{c}{VOC2012} \\
\cmidrule{2-5}
 & AP & AP\(_{50}\) & AP\(_{75}\) & AP\(_{50}\) \\
\midrule[1pt]
RetinaNet500+GHM-C & 54.6 & 81.7 & 60.1 & 80.9\\
RetinaNet500+DR-loss\(_{\text{fixed}}\) & 56.1 & 82.0 & 61.6 & 80.6\\
RetinaNet500+AP-loss & \textbf{58.0} & \textbf{83.9} & \textbf{64.2} & \textbf{83.1} \\
\midrule[1pt]
\end{tabular}}
\vspace{-2mm}
\label{GHM-DRLOSS}
\end{table}

Besides, we also compare the AP-loss with other anti-imbalance methods such as GHM~\cite{li2018gradient} and DR-loss~\cite{qian2019dr}. The experimental settings are the same as those used in Section 4.4 of the paper, and the results are shown in \autoref{GHM-DRLOSS}. Our method performs better than GHM and DR-loss on both COCO (\textit{c.f.} Benchmark Results in the paper) and PASCAL VOC datasets. This also verifies the good generalization ability of the AP-loss.

\subsection*{A4.3. AP-loss in Two-stage Detector}

%\begin{table}[t]
%\small
%\centering
%\caption{Experiments on two-stage detectors. The models are evaluated on COCO dataset.}
%\vspace{-2mm}
%\setlength{\tabcolsep}{1mm}{
%\begin{tabular}{cccccccccccccccc}
%\midrule[1pt]
%Method & AP & AP\(_{50}\) & AP\(_{75}\) & AP\(_S\) & AP\(_M\) & AP\(_L\) \\
%\midrule[1pt]
%Faster-RCNN & 37.1 & 58.2 & 40.4 & 21.0 & \textbf{41.1} & 48.7 \\
%Faster-RCNN+AP-loss & \textbf{37.3} & \textbf{59.2} & \textbf{40.6} & \textbf{21.5} & 40.5 & \textbf{49.8} \\
%\midrule[1pt]
%\end{tabular}}
%\vspace{-4mm}
%\label{two-stage}
%\end{table}
\begin{table}[t]
\small
\centering
\caption{Faster R-CNN trained with different number of RoI samples in the second stage. Models are evaluated on VOC2007 {\tt test} dataset.}
\setlength{\tabcolsep}{1.5mm}{
\begin{tabular}{cccccccccccccccc}
\midrule[1pt]
\multirow{2}*{\shortstack{\\Num of Samples\\ (per im)}} & \multicolumn{3}{c}{Faster} & \multicolumn{3}{c}{Faster+AP-Loss} \\
\cmidrule{2-4}\cmidrule{5-7}
 & AP & AP\(_{50}\) & AP\(_{75}\) & AP & AP\(_{50}\) & AP\(_{75}\)  \\
\midrule[1pt]
512  & \textbf{54.09} & \textbf{81.94} & \textbf{59.86} & 54.62 & 82.04 & 59.85 \\
1024 & 53.66 & 81.52 & 59.75 & 54.70 & 82.20 & 60.20 \\
1536 & 53.45 & 81.09 & 58.88 & 54.72 & \textbf{82.47} & 60.54 \\
2048 & 52.21 & 79.53 & 57.12 & 54.88 & 82.39 & \textbf{60.70} \\
3072 & 50.16 & 78.07 & 55.15 & \textbf{55.04} & 82.05 & 60.69 \\
4096 & 48.36 & 75.69 & 52.15 & 54.83 & 82.23 & 60.23 \\
\midrule[1pt]
\end{tabular}}
\label{two-stage}
\end{table}

\begin{figure}[t!]
\centering
\subfloat[AP]{
  \hspace{-3mm}
  \includegraphics[width=0.33\linewidth]{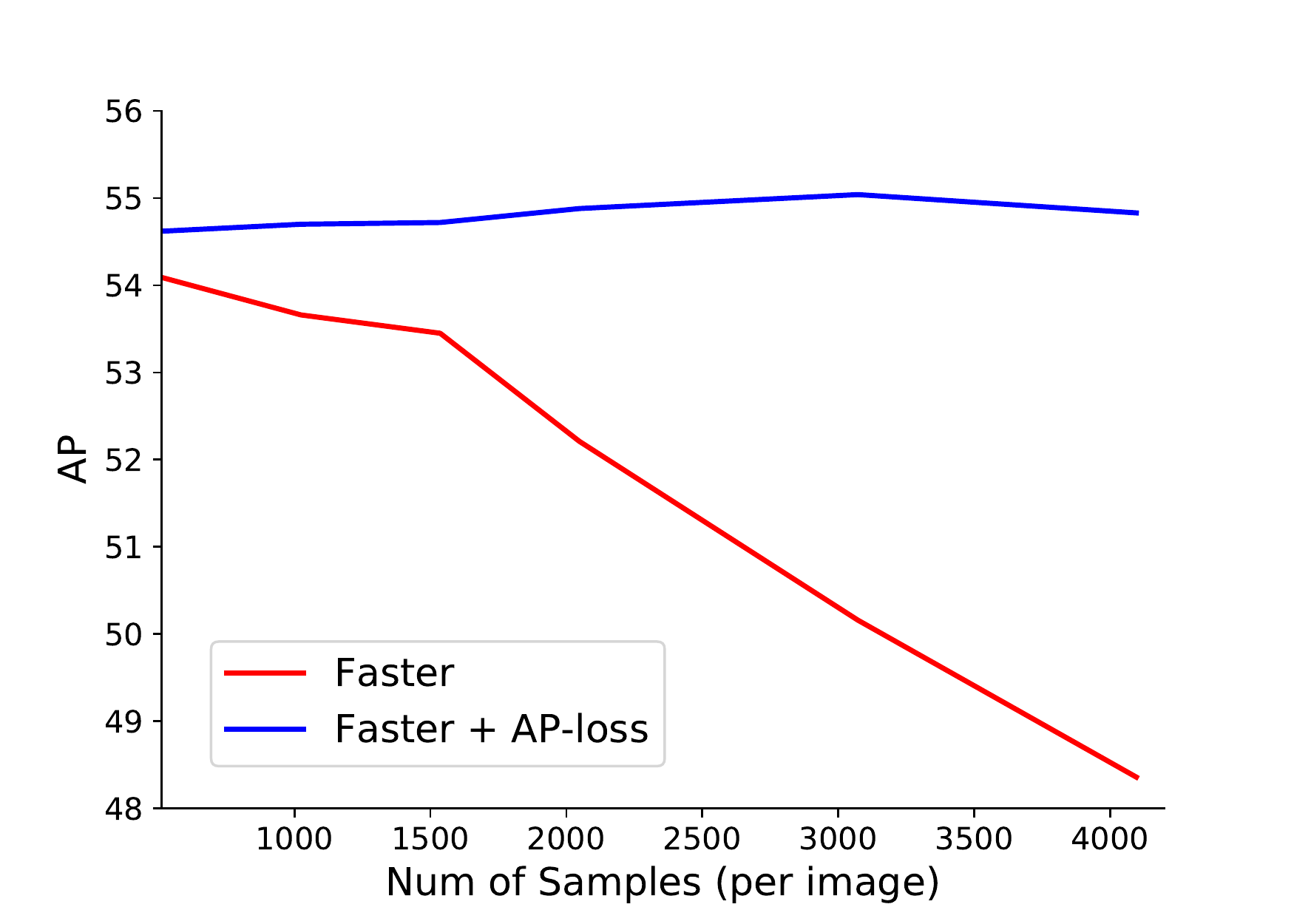}
  }
\subfloat[AP$_{50}$]{
  \includegraphics[width=0.33\linewidth]{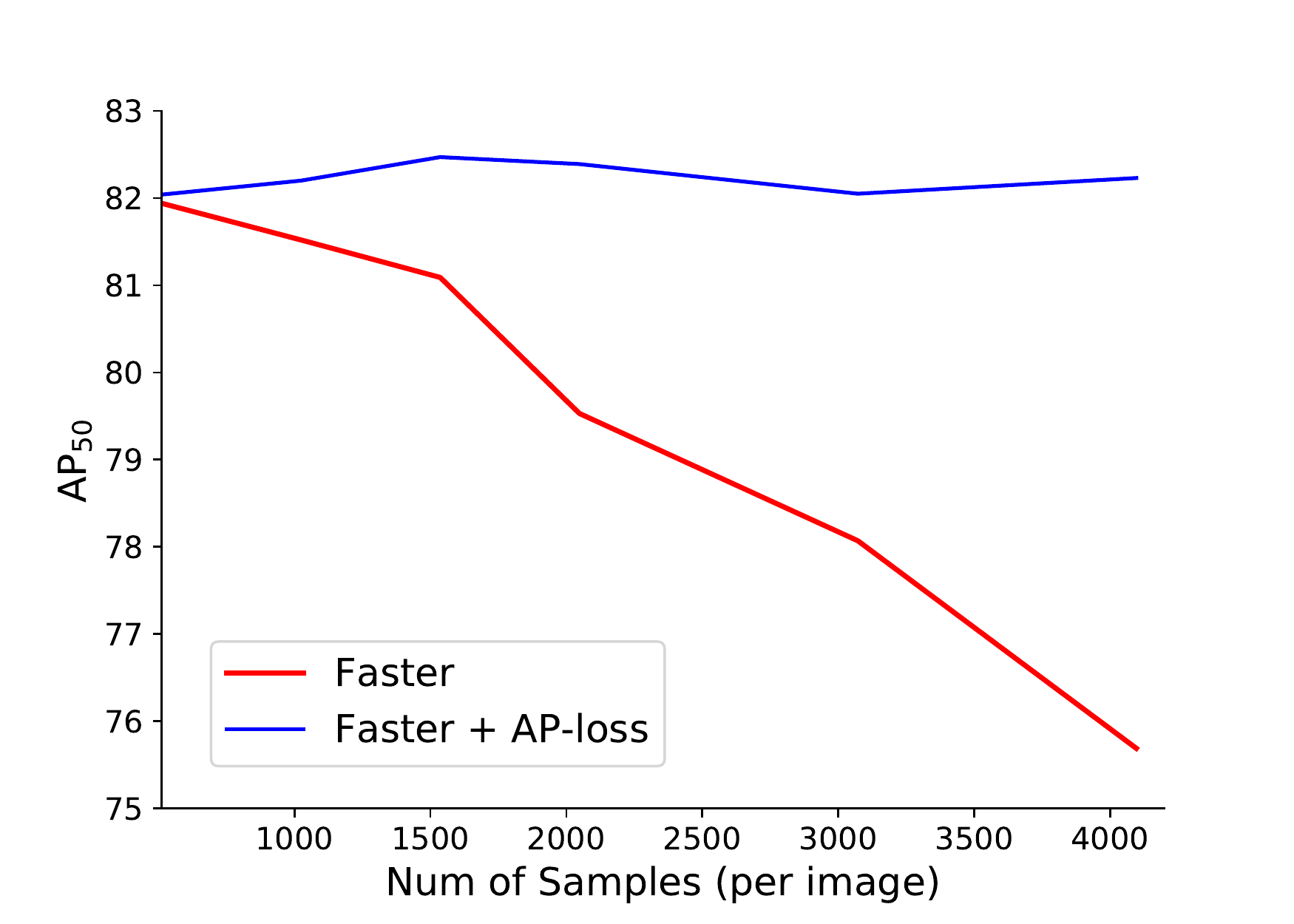}
  }
\subfloat[AP$_{75}$]{
  \includegraphics[width=0.33\linewidth]{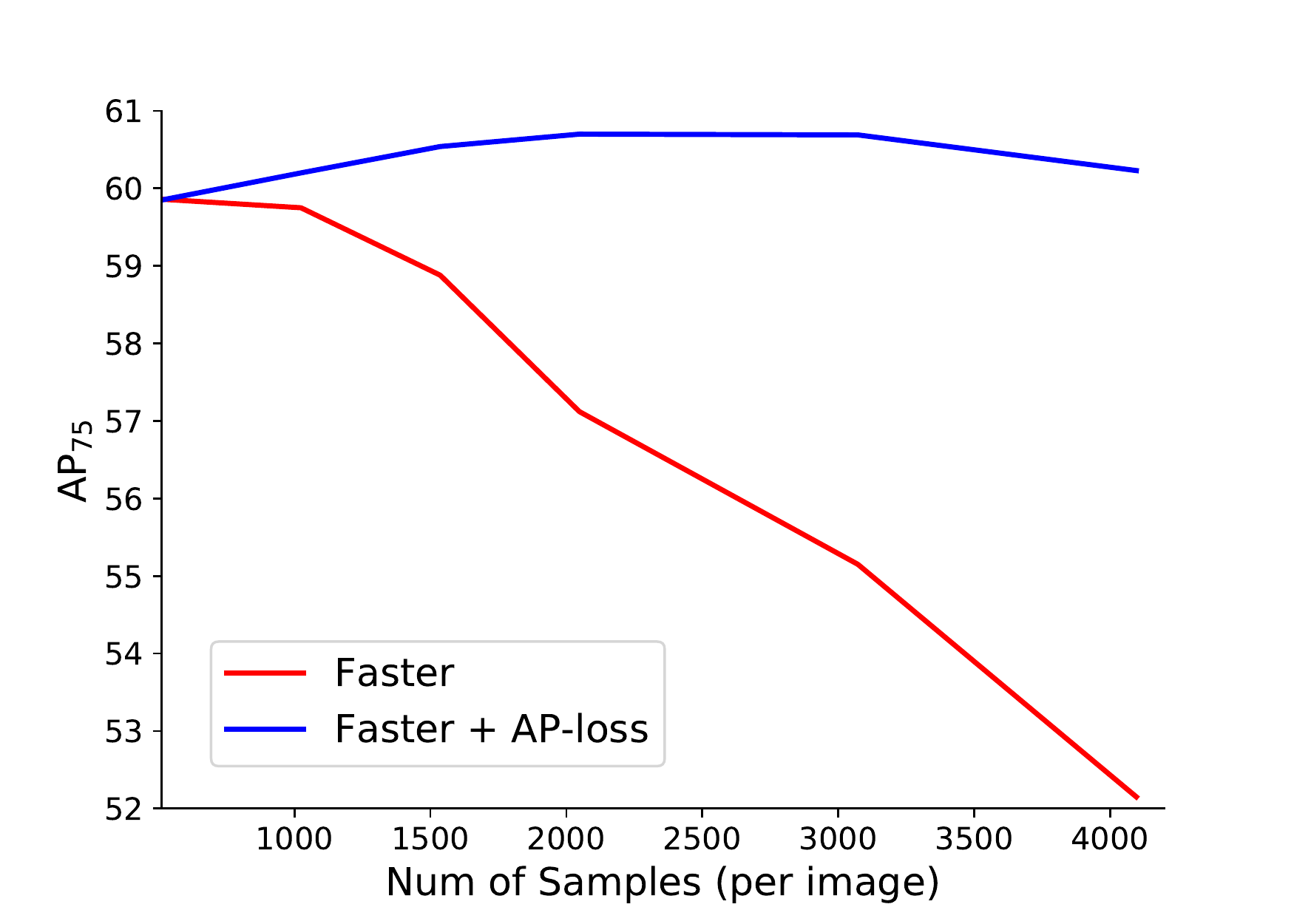}
  }
\vspace{-2mm}
  \caption{Faster R-CNN trained with different number of RoI samples in the second stage. (Best viewed in color)}
\vspace{-3mm}
\label{fig-two-stage}
\end{figure}

We also evaluate the AP-loss on two-stage detector. To simulate different imbalance conditions, the models are trained with different numbers of RoI samples in the second stage. Due to the computational complexity, it is difficult to completely simulate the extreme imbalance in the second stage. So instead, we simulated a relatively moderate imbalance condition, where the sample number ranges from 512 to 4096 (note that the sample (anchor) number is larger than $3\times10^4$ in RetinaNet). We replace the classification loss in the second stage with AP-loss. No positive-negative ratio constraint is used in the second stage. The evaluated detector is Faster R-CNN with FPN~\cite{lin2017feature} and ROIAlign~\cite{he2017mask}. We use ResNet-50 as the backbone model. We also applied both flipping and scale jittering (from 480 to 800) for training data augmentation, in the same way as the official implementation in Detectron2~\cite{wu2019detectron2}. During testing, the input image is resized to ensure the shorter side is 800 pixels. The experimental results are shown in \autoref{two-stage} and \autoref{fig-two-stage}.

Based on the experimental results, we report some positive observations: \textbf{1) AP-loss can effectively handle the imbalance issue.} With increasing class imbalance, the performance of the original Faster R-CNN model worsens when trained with more RoI samples in the second stage. Conversely, the AP-loss based model is not affected by this imbalance issue, and maintains the performance very well. If comparing on the same number of samples per image, the advantages of AP-loss becomes increasingly significant as the imbalance increases. Here, \autoref{fig-two-stage} aptly captures this difference in performance. \textbf{2) AP-loss is also helpful in the balanced condition.} When using the default setting (512 samples per image), the AP-loss is still performed slightly better than the original Cross Entropy loss on Faster R-CNN. \textbf{3) AP-loss utilizes more meaningful RoI samples.} Note that the performance of AP-loss based model slightly increases when more RoI samples are used.
%which is unexpected.
We believe the reason for this is that more valuable or meaningful RoIs are involved in training. Thus, under such conditions, the superiority of the AP-loss shows through, as it allows the two-stage detector to pick up more valuable RoIs, while preventing the detector side (second stage) from being affected by the imbalance issue. As a result, the best performance of AP-loss achieves a \(\sim\) 1\% AP improvement over the best result of the original Faster R-CNN (55.04 \textit{v.s.} 54.09).

\end{document}